\newcommand{\CLASSINPUTinnersidemargin}{10mm}
\documentclass[10pt,journal,compsoc]{IEEEtran}

\ifCLASSOPTIONcompsoc
  \usepackage[nocompress]{cite}
\else
  \usepackage{cite}
\fi

\usepackage{longtable}
\usepackage{rotating}
\usepackage{multicol}
\usepackage{supertabular}
\usepackage{setspace}
\usepackage{graphicx}
\graphicspath{{./figs/}}
\DeclareGraphicsExtensions{.pdf,.jpg,.png}
\usepackage{cite}
\usepackage[cmex10]{amsmath}
\usepackage{amssymb}
\usepackage{graphicx}
\usepackage{cases}
\usepackage{mdwmath}
\usepackage{mdwtab}
\usepackage{array}
\usepackage{url}
\usepackage[ruled,vlined]{algorithm2e}
\usepackage{booktabs}
\usepackage{threeparttable}
\usepackage{color}
\usepackage{caption}
\usepackage{pifont}
\usepackage{wasysym}
\usepackage{amssymb}
\usepackage{multirow}
\usepackage{makecell}
\usepackage{graphicx}
\usepackage[colorlinks,linkcolor=black]{hyperref}    % submitting
\usepackage{marvosym}
\usepackage{lipsum}
\usepackage[caption=false,font=footnotesize]{subfig}
\usepackage{amsmath}
\usepackage{xspace}
\usepackage{bm}
\usepackage{amsthm,amsmath,amssymb}
\usepackage{mathrsfs}
\usepackage{mathrsfs}
\usepackage{multicol} 
\usepackage{hyperref}
\usepackage{tikz}
\usetikzlibrary{mindmap, shadows}
\hypersetup{
    colorlinks=true,             %link color
    linkcolor=black,             %internal link color
    filecolor=black,             %file color
    urlcolor=black,              %url color
    citecolor=black,             %reference color
    pdftitle={Overleaf Example},
    pdfpagemode=FullScreen,
}

\tikzset{
    every annotation/.style = {draw=black, line width=1.5pt, fill=green!0, font=\Large},
    every node/.style = {concept, circular drop shadow},
    root/.style = {concept color=black!20, font=\Large\bfseries, text width=10em},
    level 1 concept/.append style = {font=\large\bfseries, sibling angle=50, text width=7.7em, level distance=15em, inner sep=0pt},
    level 2 concept/.append style = { text width=7.7em, font=\bfseries, level distance=12em,sibling angle=60}
}

\newcommand*{\info}[4][16.3]{
    \node[annotation, #3, scale=0.65, text width=#1 em,inner sep=2mm] at (#2){
        \list{$\bullet$}{
            \topsep=0pt\itemsep=0pt\parsep=0pt\parskip=0pt
            \labelwidth=8pt\leftmargin=8pt\itemindent=0pt
            \labelsep=2pt
        }
        #4
        \endlist
    };
}

\makeatletter

\newcommand{\Rmnum}[1]{\expandafter\@slowromancap\romannumeral #1@}
\makeatother
\usepackage{graphicx}
\usepackage{float}
\usepackage{diagbox}

\newcommand{\ie}{\textit{i}.\textit{e}., }
\newcommand{\eg}{\textit{e}.\textit{g}., }

\ifCLASSINFOpdf
\else
\fi

\begin{document}

% \title{A Survey on Deep Multi-modal Learning for Visual Speech: Advance and the Road Ahead}
% \title{A Survey on Deep Multi-modal Learning for Visual Gesture Language}
% \title{A Survey on Deep Multi-modal Learning for Body Language}
\title{A Survey on Deep Multi-modal Learning for Body Language Recognition and Generation}
%recent advance and the road ahead
\author{Li~Liu,\IEEEmembership{~Member,~IEEE},
        Lufei~Gao,
        Wentao~Lei,
        Fengji~Ma,
        Xiaotian~Lin,
        Jinting~Wang
        % <-this % stops a space
\IEEEcompsocitemizethanks{
\IEEEcompsocthanksitem Li~Liu, Wentao~Lei, Fengji~Ma, Xiaotian~Lin, and Jinting~Wang are with the Hong Kong University of Science and Technology (Guangzhou), Guangzhou 511458, China. E-mail: avrillliu@hkust-gz.edu.cn.\protect

\IEEEcompsocthanksitem Lufei Gao is with the Shenzhen Research Institute of Big Data, Shenzhen, China.\protect

}
\thanks{All authors are equal contribution.}
}

\markboth{IEEE Transactions on Pattern Analysis and Machine Intelligence,~Vol.~X, No.~X, December~2022}% 
{Chunhui Zhang \MakeLowercase{\textit{et al.}}: Bare Advanced Demo of IEEEtran.cls for IEEE Computer Society Journals}

\IEEEtitleabstractindextext{
\begin{abstract} 
Body language (BL) refers to the non-verbal communication expressed through physical movements, gestures, facial expressions, and postures. It is a form of communication that conveys information, emotions, attitudes, and intentions without the use of spoken or written words. It plays a crucial role in interpersonal interactions and can complement or even override verbal communication. Deep multi-modal learning techniques have shown promise in understanding and analyzing these diverse aspects of BL, which often incorporate multiple modalities.
The survey explores recent advances in deep multi-modal learning, emphasizing their applications to BL generation and recognition. Several common BLs are considered \ie Sign Language (SL), Cued Speech (CS), Co-speech (CoS), and Talking Head (TH), and we have conducted an analysis and established the connections among these four BL for the first time. Their generation and recognition often involve multi-modal approaches, for example, multi-modal feature representation, multi-modal fusion, and multi-modal joint learning will be introduced.
% Fusion methods such as early fusion, late fusion, and cross-modal attention mechanisms are discussed, highlighting their effectiveness in combining visual and audio modalities.
% A wide range of deep learning architectures, including CNNs, RNNs, and transformer-based models, are covered, showcasing their ability to capture spatio-temporal dependencies and learn representations from visual and audio inputs. The advantages and limitations of these architectures are analyzed in the context of tasks like lip-reading, Cued Speech (CS) recognition, Talking Head generation and audio-visual emotion recognition.
Benchmark datasets for BL research are well collected and organized, along with the evaluation of state-of-the-art (SOTA) methods on these datasets. The survey highlights challenges such as limited labeled data, multi-modal learning, and the need for domain adaptation to generalize models to unseen speakers or languages.
Future research directions are presented, including exploring self-supervised learning techniques, integrating contextual information from other modalities, and exploiting large-scale pre-trained multi-modal models. Real-world applications and user-centric evaluations are emphasized to drive practical adoption. 
In summary, this survey paper provides a comprehensive understanding of deep multi-modal learning for various BL generations and recognitions for the first time. By analyzing advancements, challenges, and future directions, it serves as a valuable resource for researchers and practitioners in advancing this field. \textcolor{black}{In addition, we maintain a continuously updated paper list for deep multi-modal learning for BL recognition and generation: https://github.com/wentaoL86/awesome-body-language}.

\end{abstract}

}

\maketitle
\IEEEdisplaynontitleabstractindextext

\IEEEpeerreviewmaketitle

\ifCLASSOPTIONcompsoc
\IEEEraisesectionheading{\section{Introduction}\label{sec:introduction}}
\else
\section{Introduction}
\label{sec:introduction}
\fi

\IEEEPARstart{B}{ody} language (BL), as a vital component of non-verbal communication, holds great significance in facilitating effective communication and enhancing social interactions. The ability to analyze and understand BL has various applications, ranging from BL recognition and generation to digital human interaction and assistive technologies. Understanding BL often necessitates the incorporation of multiple modalities. Deep multi-modal learning, which combines visual, audio 
and text modalities have emerged as a promising approach to enhancing the accuracy and robustness of intelligent BL multi-modal conversion systems.

\begin{figure}[htb]
\begin{minipage}[b]{1.0\linewidth}
 \centering
 \centerline{\includegraphics[width=6.0cm]{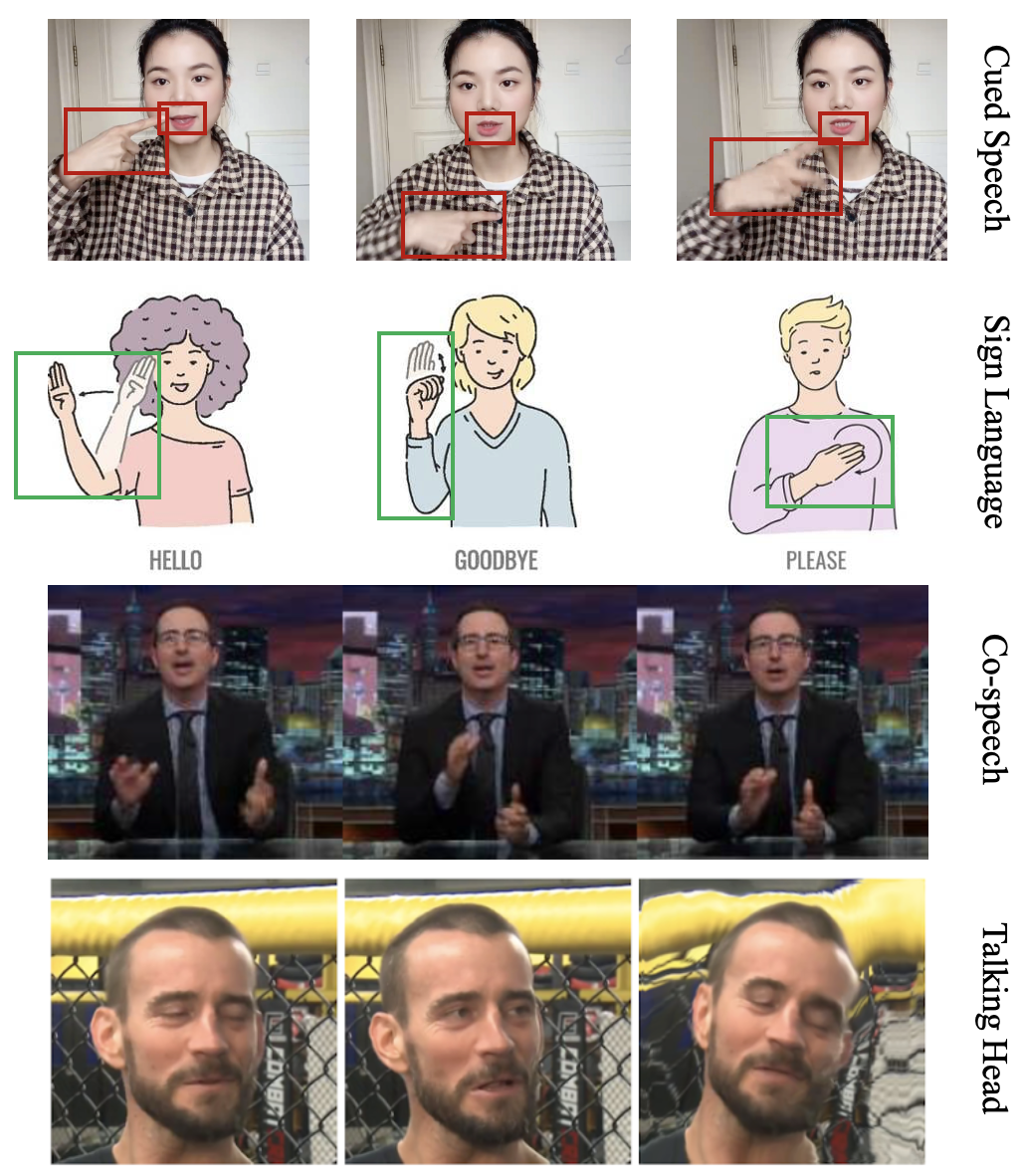}}
\end{minipage}
\caption{Examples of Cued Speech, Sign Language, Co-speech and Talking Head, respectively.}
\label{fig:CS-SL-TH-Co}
\end{figure}

%这里我想写这个领域的难点，要特别强调多模态在此领域的难点

In this survey, we primarily focus on four typical BLs and use them as examples to review and analyze the multi-modal BL recognition and generation. Figure \ref{fig:CS-SL-TH-Co} presents a simple diagram for the four types of BLs, \ie Cued Speech (CS) \cite{cornett1967cued}, Sign Language (SL) \cite{Joksimoski2022SLreview}, Co-speech (CoS) \cite{liu2022audio} and Talking Head (TH) \cite{zhang2023metaportrait}. In this field, there have been numerous previous works, which have made significant progress. 
However, despite the progress made in deep multi-modal learning for BL generation and recognition, several challenges and open research questions remain, such as the multi-modal learning of different types of data modalities, the scarcity of labeled datasets, representing fine-grained cues, modeling temporal dynamics, and limited computational resources. These challenges need to be addressed in multi-modal BL recognition and generation to further advance the field and make applications in human-computer interaction (HCI), social robotics, and affective computing more effective, etc.

% Limited labeled data, domain adaptation, generalization to unseen speakers and languages, and the need for user-centric evaluations are among the key challenges. The paper highlights these challenges and emphasizes the importance of addressing them to advance the field.
% Lastly, the survey paper outlines future research directions to inspire further advancements in deep multi-modal learning for BL. Areas such as self-supervised learning, integration of 3D spatial information, and incorporation of contextual information from other modalities are identified as promising avenues for exploration. Real-world applications and user-centric evaluations are stressed to ensure the practical adoption of BL systems.

\begin{figure*}[htb]
\begin{minipage}[b]{1.0\linewidth}
 \centering
 \centerline{\includegraphics[width=15.0cm]{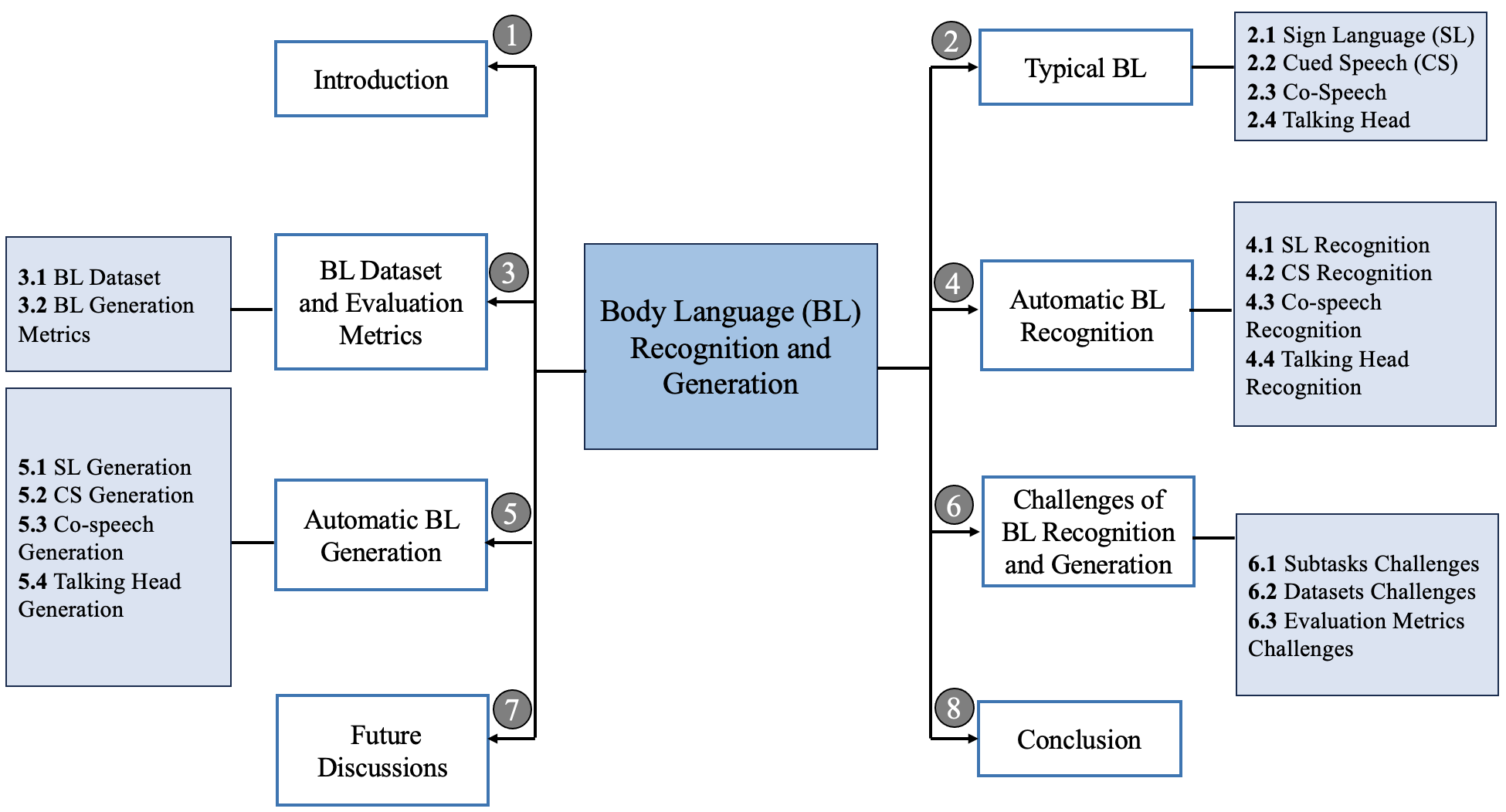}}
\end{minipage}
\caption{The architecture of this survey.}
\label{fig:Outline-paper}
\end{figure*}

\begin{figure*}
\centering
  \scalebox{0.6}{%
\begin{tikzpicture}
    \path[mindmap, concept color=black!20, text=black]            
        node[root] {Body Language}
        [clockwise from=0]
        child[concept color=green!40!blue!50, text=black, grow=-30]{
            node{{Body Language Generation}}
            [clockwise from=30]
            child { node[concept color=green!40!blue!50,text=black] (SLP) {{Sign Language production}}}
            child { node[concept color=green!40!blue!50,text=black] (CSG){{Cued Speech Generation }}}
            child { node[concept color=green!40!blue!50,text=black] (THG){{Talking Head Generation }}}
            child { node[concept color=green!40!blue!50,text=black] (COG){{Co-speech Generation}}}
        }
        child[concept color=green!60!blue!60,grow=90, text=black]{
            node[concept] {{Typical Body Language}}
            [clockwise from=180]
            child { node[concept color=green!60!blue!60,text=black] (SL) {{Sign Language}}}
            child { node[concept color=green!60!blue!60,text=black] (CS) {{Cued Speech}}}
            child { node[concept color=green!60!blue!60,text=black] (CO) {{Co-speech}}}
            child { node[concept color=green!60!blue!60,text=black] (TH) {{Talking Head}}}
        }
        child[concept color=orange!60!yellow!60,grow=210]{
            node[concept] (GLR) {{Body Language Recognition}}
            [clockwise from=270]
            child { node[concept](LRR) {{Lip reading recognition}}}
            child { node[concept](CSR) {{Cued Speech recognition}}}
            child { node[concept](SLR) {{Sign Language recognition}}}            
        }
        ;
        \info {SLP.north east}{above,anchor=west,shift={(1.25em, -2.5em)}}{%
            \item[] Text2sign\cite{Stephanie2020text2sign}
            \item[] HLSTM\cite{guo2018hierarchical}
            \item[] ESN\cite{saunders2020everybody}
        }
        \info [15]{COG.south}{below,anchor=north,shift={(0em, -0.25em)}}{%
            \item[] DiffGAN\cite{Ahuja2022low}
            \item[] RG\cite{ao2022rhythmic}
            \item[] SEEG\cite{Liang2022seeg}
            \item[] HA2G\cite{liu2022learning}
        }
                \info [15]{CSG.east}{below,anchor=west,shift={(0.25em, -0.25em)}}{%
            \item[] Paul Duchnowski\cite{duchnowski1998automatic}
            \item[] G{\'e}rard Bailly\cite{bailly2008retargeting}
        }
        \info [15]{THG.east}{below,anchor=west,shift={(0.25em, -1.25em)}}{%
            \item[] Wav2lip \cite{prajwal2020lip}
            \item[] Audio2head\cite{wang2021audio2head}
            \item[] AD-NERF\cite{guo2021ad}
            \item[] DiffTalk\cite{shen2023difftalk}
        }
        \info [15]{LRR.west}{below,anchor=east,shift={(-0.25em, -1.25em)}}{%
            \item[] Rule-based\cite{lucey2008patch}
            \item[] LBP\cite{zhou2011towards}
            \item[] SDF\cite{wu2016novel}
            \item[] PTSLP\cite{ma2021end}
        }
        \info [15]{CSR.west}{below,anchor=east,shift={(-0.25em, -0.25em)}}{%
            \item[] Syn\cite{liu2019novel}
            \item[] MMFSL\cite{papadimitriou2021multimodal}
            \item[] Re-Syn\cite{liu2020re}
            \item[] CMML\cite{liu2023cross}
        }
        \info [15]{SLR.west}{below,anchor=east,shift={(-0.25em, -0.25em)}}{%
            \item[] DTW\cite{zhang2014threshold}
            \item[] HMMs\cite{yang2016continuous}
            \item[] FCN\cite{cheng2020fully}
            \item[] RL\cite{wei2020semantic}
        }
            \info [15]{CS.west}{below,anchor=east,shift={(-0.25em, -0.25em)}}{%
            \item[] CS\cite{cornett1967cued}
            \item[] MCCS\cite{liu2019pilot}
        }
            \info [15]{SL.west}{below,anchor=east,shift={(-0.25em, -0.25em)}}{%
            \item[] SLreview\cite{Joksimoski2022SLreview}
            \item[] UnorgSign\cite{unorgsign2022}
        }
            \info [15]{CO.east}{below,anchor=west,shift={(0.25em, -0.25em)}}{%
            \item[] CoSreview\cite{nyatsanga2023comprehensive}
            \item[] CoSreview\cite{liu2022audio}
            \item[] SE\cite{pan2022speaker}
        }
            \info [15]{TH.east}{above,anchor=west,shift={(0.25em, -0.25em)}}{%
            \item[] THreview\cite{zhang2023metaportrait}
            \item[] THE\cite{sondermann2023like}
            \item[] VHTHG\cite{song2023virtual}
        }
        %                 \info [15]{SL.west}{below,anchor=east,shift={(-0.25em, -0.25em)}}{%
        %     \item[]since 1998\\
        %     \item \cite{duchnowski1998automatic}Paul Duchnowski
        %     \item \cite{duchnowski1998automatic}G{\'e}rard Bailly
        % }
        
    \end{tikzpicture}
    }
    \caption{Structured taxonomy of the existing BL research which includes three genres. Only several representative methods of each category are demonstrated.}
    \label{fig:taxonomy}
\end{figure*}
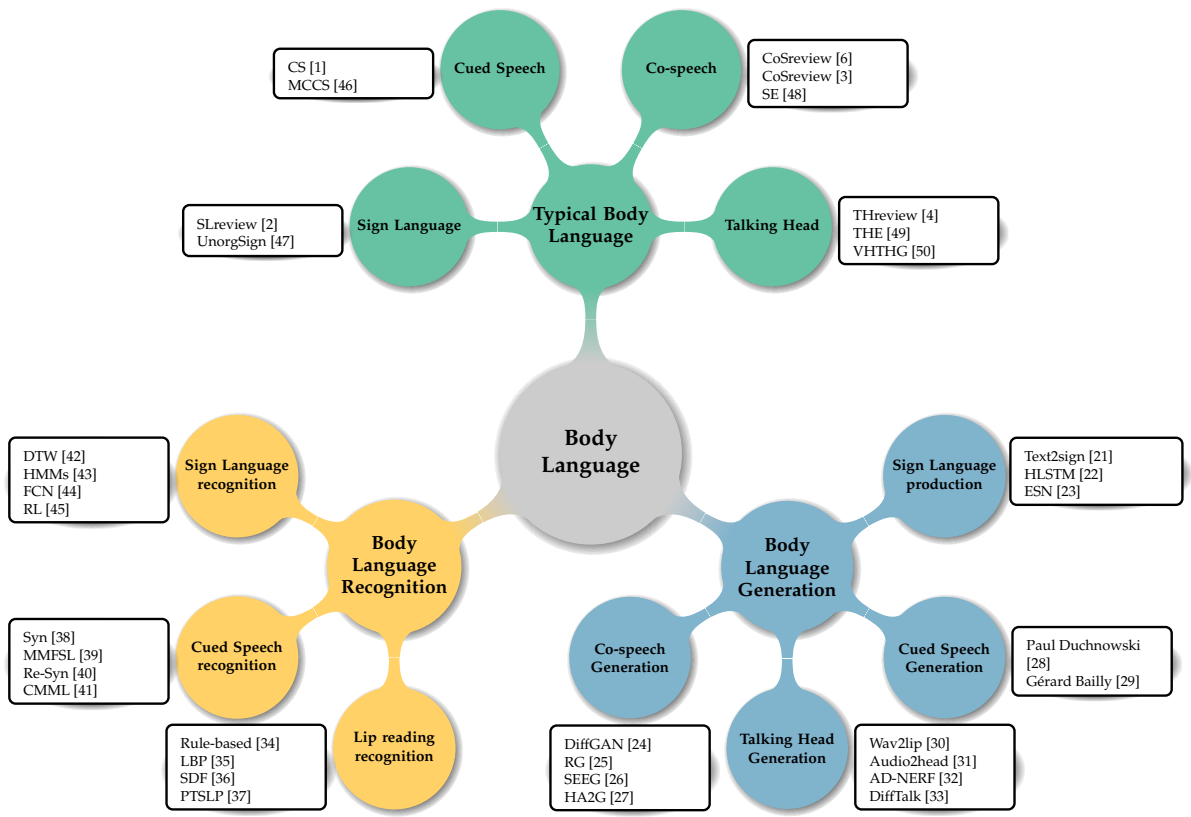

\textbf{Organization of This Survey.}
In this survey, we first introduce four typical variants of BL and establish the connections between these four types in Section 2. Then, We organize and present various types of datasets for BL recognition and generation, along with evaluation metrics in Section 3. In Sections 4 and 5, we provide detailed reviews of the BL recognition and generation of CS, SL, CoS and TH, respectively.
Furthermore, in Section 6, we give a detailed analysis of the challenges for these types of BL. Finally, we discuss and conclude this survey by proposing multiple research
directions that need to be studied. The architecture of this survey is visualized in Figure \ref{fig:Outline-paper}. The structured taxonomy of the existing BL research and some representative works are shown in Figure \ref{fig:taxonomy}.

\begin{table}[htbp]
  \small
  \begin{center}
  \caption{The number of existing reviews.}
  \renewcommand{\arraystretch}{1.5} % 调整行间距
    \label{tab:related-work}
    \begin{tabular}{ccccccccc}
      \hline 
      \textbf{Type} & \textbf{SL} & \textbf{CS} & \textbf{CoS} & \textbf{TH} & \textbf{LR} & \textbf{SL+CS} & \textbf{LR+TH} & \textbf{Total}\\
      \hline
      \textbf{R} & 5 & 1 & 0 & 0 & 5 & 1 & 0 & 12  \\
      \hline
      \textbf{G} & 4 & 0 & 1 & 3 & 0 & 0 & 0 & 8 \\
      \hline
      \textbf{R\&G} & 1 & 0 & 0 & 0 & 0 & 0 & 1 & 2\\
      \hline
      \textbf{Total} & 10 & 1 & 1 & 3 & 5 & 1 & 1 & 22 \\
      \hline 
      \end{tabular}
    \end{center}
     \footnotesize{\textit{The corresponding terms for the abbreviations are as follows: R -- Recognition; G -- Generation; SL -- Sign Language; CS -- Cued Speech; CoS -- Co-speech; TH -- Talking Head; LR -- Lip Reading.}}
\end{table}

\textbf{Differences from Existing Reviews.}
Table \ref{tab:related-work} presents the number of review articles related to BL recognition and generation in the relevant field. While there are already 22 existing surveys, the differences between our survey and these prior works can be summarized as follows:
\begin{itemize}
    \item \textbf{Scope}. Existing reviews on BL~\cite{rastgoo2021sign,nyatsanga2023comprehensive,Rastgoo2021recognition} have only focused on specific subtasks within the field. For BL recognition, the reviews \cite{fernandez2018survey, fenghour2021deep,chand2023survey,bhaskar2018survey, radha2021survey, rastgoo2021sign,koller2020quantitative, adeyanju2021machine, papastratis2021artificial, madhiarasan2022comprehensive} concentrate on SL recognition. Regarding BL generation, \cite{nyatsanga2023comprehensive} only explores CoS generation and \cite{Chen2020lrsd, sha2023deep, zhen2023human} delves into TH generation. \cite{Sheng2022vsa} integrates subtasks: LR recognition and TH generation. Unlike the reviews mentioned earlier, this paper focuses on two primary tasks: \textbf{recognition} and \textbf{generation}. Each task is expanded to incorporate four different types of BL: \textbf{SL}, \textbf{CS}, \textbf{CoS}, and \textbf{TH}. As far as we know, this is the first to encompass all four types of BL along with their corresponding recognition and generation tasks.  
    \item \textbf{Timeline.} This survey highlights the latest advances, major challenges and deep learning (DL)-based multi-modal approaches in the aforementioned research areas from 2017 to the present. Please note that we will consistently update the repository we maintain with the latest developments. It is expected that this study will facilitate knowledge accumulation and the creation of deep multi-modal BL methods, providing readers, researchers, and practitioners with a roadmap to guide future direction.
\end{itemize}

To summarize, this survey provides a thorough examination of the progress made in deep multi-modal learning techniques for automatic BL recognition and generation. It also outlines the road ahead for future research in this area. The objective is to offer researchers and practitioners a consolidated understanding of the field, covering the foundational principles, multi-modal fusion methods, DL architectures, benchmark datasets, challenges, and potential directions. 

\begin{figure}[htb]
\begin{minipage}[b]{1.0\linewidth}
 \centering
 \centerline{\includegraphics[width=9.5cm]{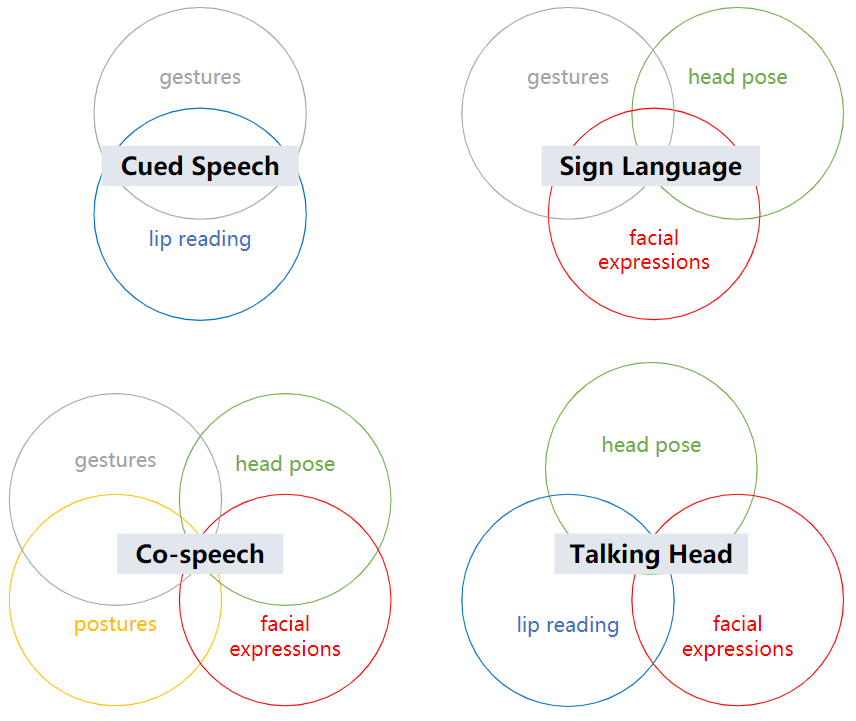}}
\end{minipage}
\caption{Element compositions of four typical body language cases.}
\label{fig:elem_comp}
\end{figure}

% \begin{figure}[htb]
% \begin{minipage}[b]{1.0\linewidth}
%  \centering
%  \centerline{\includegraphics[width=9.0cm]{diffusion_process.png}}
% \end{minipage}
% \caption{Illustration of Diffusion Process in Co-Speech Gesture Generation. The diffusion process $q$ gradually introduces Gaussian noise into the gesture sequence ($x_0$ sampled from the real data distribution). The generation process $p_\theta$ learns to denoise the white noise ($x_T$ sampled from the normal distribution) conditioned on context information c. Here, $x_t$ represents the corrupted gesture sequence at the t-th diffusion step.}
% \label{fig:diffusion}
% \end{figure}

\section{Typical Body Language}
BL through which humans convey information usually involves five aspects, \ie gestures, facial expressions, lip reading (LR), head pose and postures. In this survey, we refer to these five aspects as the basic elements of BL. 

Gestures refer to the use of hand movements to convey meaning. People communicate through actions such as waving, pointing, or gesturing with their hands. Additionally, facial expressions play a crucial role as a basic element of BL. Humans express emotions and intentions by altering the facial muscles around the eyes, eyebrows, mouth, etc. Another fundamental element is LR, which involves interpreting speech by observing the movements of the lips and mouth. Furthermore, head pose, including tilting or turning the head, can also convey information related to attention, interest, or specific desires. Lastly, postures, such as standing, sitting, or body leaning, contribute to conveying emotional states and social intentions within BL.

It is common for BL cases to consist of two or more of these modalities. As shown in Figure \ref{fig:elem_comp}, we listed four typical BL cases that are discussed in this survey, and each of them can be regarded as a composition of the basic BL elements.
In this section, we will provide a comprehensive overview of these four BL cases, including their concepts, significance, and the challenges that exist in their corresponding recognition or generation tasks.

\subsection{Sign Language}
SL is categorized as a natural language commonly used in deaf communities \cite{Joksimoski2022SLreview}. 
Based on data from the World Federation of the Deaf, the worldwide population of the deaf is estimated to be around 72 million, with over 80\% living in developing nations\cite{unorgsign2022}. Over 300 different SLs are used by these individuals, each having its own distinct vocabulary and grammar. SL is also known as a visual language which is generally composed of several visual partials, such as gestures, facial expressions, head pose and body postures. Specifically, six basic parameters are listed as the basic components of SL in \cite{kothadiya2022deepsign}, \ie hand shape, orientation, movement, location, mouth shape and eyebrow movements. Taking an overall perspective into account, we regard gestures, facial expressions, and head poses as the primary visual modalities in SL.

SL is the major communication tool for the deaf, yet it is difficult to be mastered. In order to eliminate communication barriers, it is of great significance to develop technologies for automatic SL processing, including SL recognition (SLR) that extracts words or utterances by capturing and analyzing image or video sequences of the SL data, SL generation (SLG) that generates visualizable SL animations from input with semantic meaning and SL translation that translates the extracted information to another signed or spoken language \cite{de2023machine}\cite{kahlon2023machine}. This survey mainly focuses on the literature review of SLR and SLG in order to deeply understand the important issues and difficulties in the field of SL processing.

As a highly dynamic and multi-modal visual language, SL involves a combination of multiple visual elements that have complementary semantics. Therefore, extracting and fusing high-dimensional features from different modalities effectively is an important task.
Deep multi-modal learning techniques play a pivotal role in addressing these challenges and advancing the field of SL processing.
By combining visual and spatial information from video or depth sensors with linguistic cues, these approaches have shown promising results in improving the accuracy and naturalness of SLR and SLG systems.

In Section \ref{ssec:slr} and \ref{ssec:slg}, we investigate the recent research advancements and techniques in deep multi-modal learning specifically for SLR and SLG. Besides, we delve into the challenges that are associated with these tasks and emphasize the potential applications and future directions in this field.

\subsection{Cued Speech}
\begin{figure}[htb]
\begin{minipage}[b]{1.0\linewidth}
 \centering
 \centerline{\includegraphics[width=9.0cm]{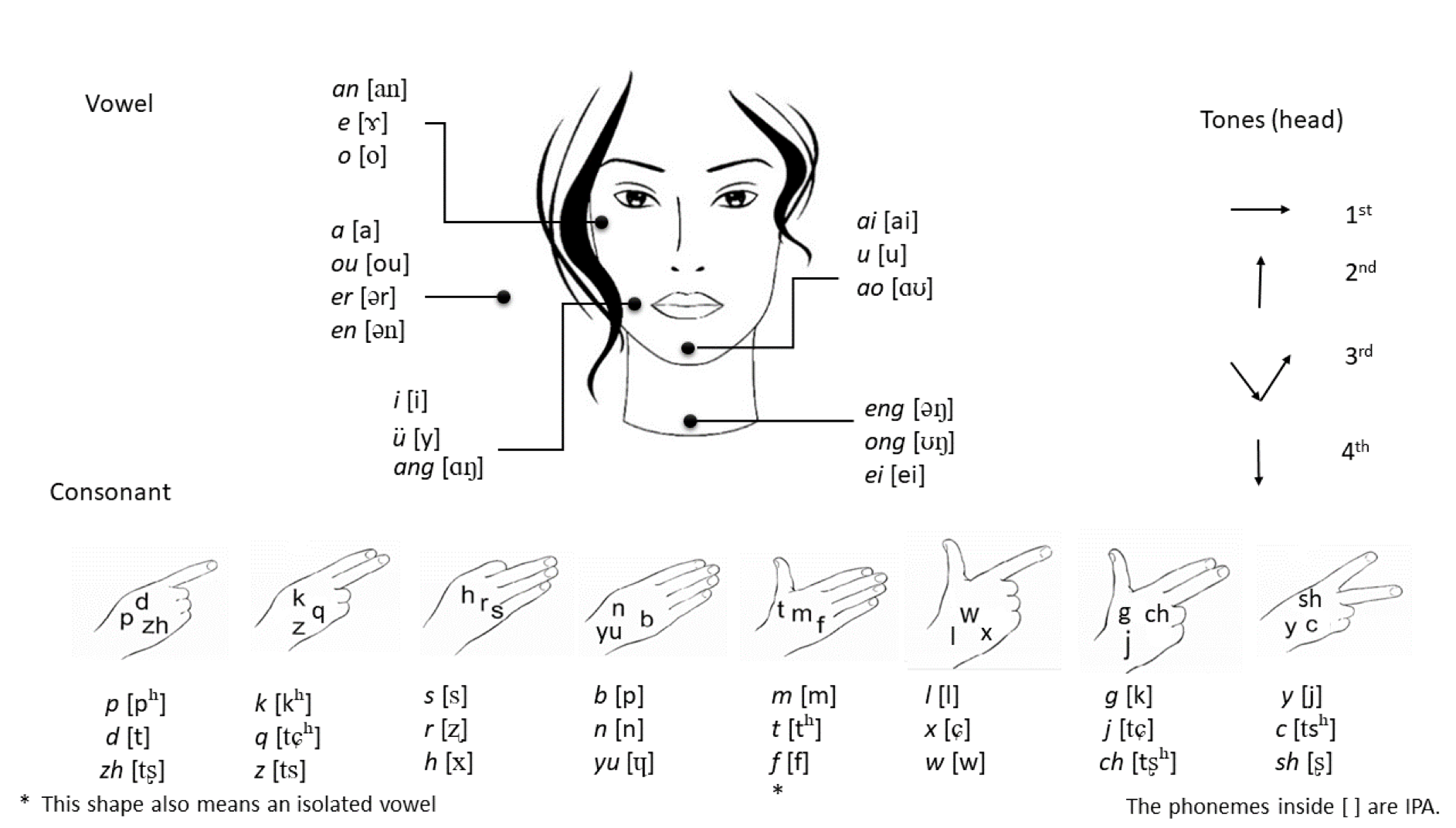}}
\end{minipage}
\caption{
The corresponding table between hand shapes and hand positions in Mandarin Chinese CS for vowels and consonants, respectively.}
\label{fig:CS-Chart}
\end{figure}

CS is a visual communication system proposed by Cornett \cite{cornett1967cued} to enhance speech perception for individuals with hearing loss. CS uses a set of hand shapes and positions, named cues, to code the phonemes such as consonants and vowels. Figure \ref{fig:CS-Chart} presents the chart for Mandarin Chinese CS (MCCS) \cite{liu2019pilot,liu2022objective}. Gestures in CS functions as a complement to lip-reading, visualizing the phonetic details that can be observed from the mouth movements to remove ambiguities caused by lip-reading alone. As a clear and unambiguous visual counterpart to the auditory information in spoken language, CS enables individuals with hearing loss to better understand and distinguish speech sounds, facilitating their language acquisition, spoken capabilities, reading skills, and overall communication abilities.

CS has currently been adapted to approximately 65 languages and dialects globally, including English, French, Chinese, etc. \cite{csorg}. Recently, there has been growing interest in developing technologies for automatic recognition and generation in the CS research field \cite{liu2018modeling,liu2018automatic}. These technologies aim to enhance accessibility for people who primarily use CS for communication.  
For instance, utilizing automatic CS recognition (ACSR), people can effortlessly transcribe gestures and lip-reading into corresponding spoken language at a phonemic level \cite{wang2021cross,liu2019automatic}.
In the opposite direction, a digital agent equipped with automatic CS generation (ACSG) can convert spoken input into authentic CS expressions.

To process CS data efficiently, it is crucial to effectively extract information from two modalities: hand and lip movement. However, this task poses challenges in several aspects. Firstly, there exists an inherent asynchrony phenomenon when the human brain processes speech with gestures \cite{liu2020re}; secondly, recognizing or generating appropriate hand shapes and lip movements entails tackling fine-grained image processing problems \cite{park2022synctalkface,liu2017inneravsp,liu2017automatic}. Consequently, deep multi-modal learning techniques have emerged as a prominent research trend to uncover the interplay between gestures and LR, aiming to achieve high-performance ACSR or ACSG systems.

In Section \ref{ssec:csr} and \ref{ssec:csg}, we will discuss some SOTA methods to solve the problems lying in CS processing. We explore the challenges and opportunities in leveraging deep multi-modal learning techniques in this new research area, aiming to enhance the accessibility and inclusivity of communication for individuals who rely on CS.

\subsection{Co-speech}
CoS refers to the non-verbal behaviors and signals that accompany and complement spoken language during communication \cite{nyatsanga2023comprehensive}. It encompasses various visual cues, such as gestures, body postures, and facial expressions such as eye gaze and blinking, which can be used in conjunction with speech to convey additional information and meaning \cite{alibali1999illuminating}\cite{kang2016hands}.   
CoS gestures contribute substantially to the overall comprehension and interpretation of spoken language \cite{kendon1994gestures}. They serve as contextual cues, accentuate salient points, convey emotional states, and facilitate social interactions \cite{kendon2004gesture}. 

With the development of AI agents technologies, there has been extensive research exploration in CoS generation or synthesis to give AI agents such as digital humans more expressive and realistic BL \cite{ferstl2018investigating, yoon2019robots, yoon2022genea, ginosar2019learning, liu2022learning}. The primary objective of this task is to generate a sequence of human BL by utilizing speech audio and transcripts as input, enhancing the performance of human-machine interaction systems. On the other hand, most existing gesture recognition methods primarily focus on recognizing specific types of gestures \cite{poppe2010survey,wu2016deep,wan2016chalearn,materzynska2019jester}, overlooking their connections with other modalities such as speech.

CoS signals not only play a crucial role in enhancing the clarity, expressiveness and emotional content of verbal communication, but also capture the rich communicative context, and reveal the speaker's social identity and cultural affiliation \cite{pan2022speaker}. Therefore, it is a growing trend towards exploring multi-modal approaches that take into account both the visual information from gestures and the accompanying speech signals, which allows for more comprehensive and accurate analysis in the areas such as emotion recognition and dialogue understanding \cite{zeng2022gesturelens}.

In Section \ref{ssec:cosp_r}, we provide a brief overview of the work on automatic CoS recognition. Due to its limited application scenarios, research in this field is relatively scarce. In Section \ref{ssec:cosp_g}, we review the SOTA techniques and advancements in deep multi-modal learning for CoS generation, highlighting the potential applications and future research directions in this field.

\begin{table*}[htbp]
  \caption{Multi-Modal Body Language Datasets.}
  \begin{center}
  \label{tab:dataset}
  \setlength{\tabcolsep}{1mm}{
  \begin{tabular}{clccccc}
    \toprule
    \textbf{Type} & \textbf{Name} & \textbf{Year} & \textbf{Scale} & \textbf{Modal} & \textbf{Language} & \textbf{Link} \\ 

    \bottomrule

    \multirow{15}*{Sign Language} 
         &Dicta-Sign \cite{efthimiou2010dicta}&2008&$\sim$1k&Video-Text& English & \href{https://www.sign-lang.uni-hamburg.de/dicta-sign/portal/}{Link}\\
     &PHOENIX-Weather\cite{koller2015continuous}&2012&$\sim$3k&Video-Text& Germany & \href{https://www-i6.informatik.rwth-aachen.de/~koller/RWTH-PHOENIX/}{Link}\\
    &ASLLVD\cite{neidle2012challenges} &2012&$\sim$3K&Video-Text & English &  \href{https://www.bu.edu/asllrp/av/dai-asllvd.html}{Link} \\   
    &SIGNUM \cite{agris2010signum} &2013&$\sim$33K&Video-Text & Germany &  \href{https://www.phonetik.uni-muenchen.de/forschung/Bas/SIGNUM/}{Link} \\ 
    &DEVISIGN\cite{lin2015curve} &2014&$\sim$24k&Video-Text & Chinese &  \href{http://vipl.ict.ac.cn/homepage/ksl/data_ch.html}{Link} \\
    &ASL-LEX 1.0\cite{caselli2017asl} &2017&$\sim$1K &Video-Text & English &  \href{https://asl-lex.org/download.html}{Link} \\    
    &PHOENIX14T\cite{camgoz2018neural} &2018&$\sim$68K&Video-Text & Germany &  \href{https://www-i6.informatik.rwth-aachen.de/~koller/RWTH-PHOENIX-2014-T/}{Link} \\    
    &CMLR\cite{mavi2022new} &2019&$\sim$102K &Image-Text & Chinese &  \href{https://www.vipazoo.cn/CMLR.html}{Link} \\
    &KETI\cite{ko2019neural} &2019&$\sim$15K&Video-Text & Korean &  Not Available \\ 
    &GSL\cite{adaloglou2021comprehensive} &2020&$\sim$3K&Video-Text & Greek &  \href{https://zenodo.org/record/3941811#.ZHb2LXZBxD8}{Link} \\
    &ASL-LEX 2.0\cite{sehyr2021asl} &2021&$\sim$ 10K&Video-Text-Depth & English &  \href{https://asl-lex.org/download.html}{Link} \\ 
        &How2sign\cite{duarte2021how2sign} &2021&$\sim$35K&Video-Text-Skelton(2D)-Depth & English &  \href{https://how2sign.github.io/}{Link} \\ 
    &Slovo\cite{kapitanov2023slovo} &2023&$\sim$20K &Video-Text & Russian  &  \href{https://github.com/hukenovs/slovo}{Link} \\
    &AASL\cite{al2023rgb} &2023&$\sim$8K  &Image-Text & Arabic &  \href{https://www.kaggle.com/datasets/muhammadalbrham/rgb-arabic-alphabets-sign-language-dataset}{Link} \\
    &ASL-27C\cite{mavi2022new} &2023&$\sim$23K &Image-Text & English &  \href{https://www.kaggle.com/datasets/ardamavi/27-class-sign-language-dataset}{Link} \\ 
    \bottomrule
            \multirow{5}*{Cued Speech} 
              ~  & FCS\cite{liu2018visual} &2018&$\sim$13k&Video-Text-Audio&French&\href{https://zenodo.org/record/5554849}{Link}\\
        & BEC\cite{liu2019automatic} &2019&$\sim$3k &Video-Text-Audio&English&\href{https://zenodo.org/record/3464212}{Link}\\
        & PCSC \cite{Trochymiuk2007VOT} &2020&20 (P) &Video-Text-Audio&Polish &\href{https://phonbank.talkbank.org/access/Clinical/PCSC.html}{Link}\\   
        & CLeLfPC\cite{bigi2022clelfpc} &2022&350 &Video-Text-Audio&French&\href{https://www.ortolang.fr/market/corpora/clelfpc}{Link}\\
        & MCCS-2023\cite{liu2023cross} &2023&$\sim$132k &Video-Text-Audio-Skelton(2,3D)&Chinese&\href{https://mccs-2023.github.io/}{Link}\\ 
    \bottomrule
    
    \multirow{5}*{Co-speech}
                & Trinity\cite{ferstl2018investigating} &2018&224(Min)&Videos-Text-Audio-Skelton(2,3D)&English&\href{https://trinityspeechgesture.scss.tcd.ie/}{Link}\\
                & TED-Gesture\cite{yoon2019robots} &2019&$\sim$252k&Videos-Text-Audio-Skelton(2D)&English&\href{https://github.com/youngwoo-yoon/youtube-gesture-dataset}{Link}\\ 
            
                & Talking With Hands \cite{yoon2022genea} &2019&200&Videos-Text-Audio-Skelton(2,3D)&English&\href{https://github.com/facebookresearch/TalkingWithHands32M}{Link}\\
                & Speech2Gesture\cite{ginosar2019learning} &2019&$\sim$60k&Videos-Text-Audio-Skelton(2D)&English&\href{http://people.eecs.berkeley.edu/~shiry/speech2gesture/}{Link}\\
                & TED-Expressive\cite{liu2022learning} &2022&$\sim$252k&Videos-Text-Audio-Skelton(2,3D)&English&\href{https://github.com/alvinliu0/HA2G}{Link}\\
        \bottomrule
                \multirow{24}*{Talking Head}
                & GRID  \cite{cooke2006Grid}&2006&$\sim$34k  &Video-Text&English&\href{https://spandh.dcs.shef.ac.uk/gridcorpus/}{Link}\\  
                & eNTERFACE \cite{martin2006enterface}& 2006& $\sim$1k& Video-Text-Audio& Multiple& \href{http://www.enterface.net/enterface05}{Link}\\
               & MIRACL-VC1 \cite{rekik2016adaptive}&2014&$\sim$3k &Video-Text-Depth&English&\href{https://sites.google.com/site/achrafbenhamadou/-datasets/miracl-vc1}{Link}\\      
                & CREMA-D\cite{cao2014crema} &2015&$\sim$7k &Video-Text-Audio&English&\href{https://github.com/CheyneyComputerScience/CREMA-D}{Link}\\
                & TCD-TIMIT \cite{cao2014crema} &2015&$\sim$7k &Video-Text-Audio&English&\href{https://sigmedia.tcd.ie/TCDTIMIT/}{Link}\\
        & MODALITY \cite{czyzewski2017audio} &2015&$\sim$6k &Video-Text-Audio&English&\href{http://www.modality-corpus.org/}{Link}\\ 
        & LRW \cite{chung2017lip} &2016&$\sim$539k&Video-Text&English&\href{https://www.robots.ox.ac.uk/~vgg/data/lip_reading/lrw1.html}{Link}\\   
        & MSP-IMPROV \cite{busso2016msp} &2016& $\sim$ 1K &Video-Text-Audio&English&\href{https://ecs.utdallas.edu/research/researchlabs/msp-lab/MSP-Improv.html}{Link}\\ 
        & ObamaSet\cite{suwajanakorn2017synthesizing} &2017&$\sim$1k &Video-Text-Audio&English&\href{https://github.com/supasorn/synthesizing_obama_network_training}{Link}\\
              ~  & VoxCeleb1\cite{nagrani2017voxceleb} &2017&$\sim$22k&Video-Text-Audio&English&\href{https://www.robots.ox.ac.uk/~vgg/data/voxceleb/vox1.html}{Link}\\                
              ~  & VoxCeleb2\cite{chung2018voxceleb2} &2018&$\sim$146k&Video-Text-Audio&English&\href{https://www.robots.ox.ac.uk/~vgg/data/voxceleb/vox2.html}{Link}\\
                        & LRS2 \cite{afouras2018deepAVSR}&2018&$\sim$96k &Video-Text&English&\href{https://www.robots.ox.ac.uk/~vgg/data/lip_reading/}{Link}\\
                & LRS3-TED \cite{afouras2018lrs3}&2018&$\sim$119k &Video-Text&English&\href{https://www.robots.ox.ac.uk/~vgg/data/lip_reading/}{Link}\\
             & RAVDESS \cite{livingstone2018ryerson} &2018&$\sim$1k  &Video-Text-Audio&English&\href{https://www.kaggle.com/datasets/uwrfkaggler/ravdess-emotional-speech-audio}{Link}\\  
             & MELD \cite{poria2018meld} &2018&$\sim$13k &Video-Text-Audio&English&\href{https://affective-meld.github.io/}{Link}\\ 
        & AVSpeech \cite{ephrat2018looking}& 2018&$\sim$150k& Video-Audio& Multiple& \href{http://looking-to-listen.github.io/}{Link}\\
        & VOCASET\cite{cudeiro2019capture} &2019&480 &Video-Text-Audio-3DFace&English&\href{https://voca.is.tue.mpg.de/}{Link}\\
                        & LRW-1000 \cite{yang2019lrw}&2019&$\sim$718K &Video-Text&Chinese&\href{https://vipl.ict.ac.cn/resources/databases/201810/t20181017_32714.html}{Link}\\
                &  FaceForensics++\cite{rossler2019faceforensics++} &2019&$\sim$1k &Video-Text-Audio&English&\href{https://github.com/ondyari/FaceForensics}{Link}\\
        & MEAD\cite{wang2020mead} &2020& $\sim$281k &Video-Text-Audio&English&\href{https://wywu.github.io/projects/MEAD/MEAD.html}{Link}\\
        & HDTF\cite{zhang2021flow} &2021&$\sim$10k &Video-Text-Audio&English&\href{https://github.com/MRzzm/HDTF}{Link}\\
         & AnimeCeleb \cite{kim2022animeceleb} &2022&$\sim$2.4M &Video-Text-Audio-3DFace&English&\href{https://github.com/kangyeolk/AnimeCeleb}{Link}\\
        & VLRDT \cite{berkol2023visual}&2022&$\sim$2k &Video-Text&Turkish&\href{https://data.mendeley.com/datasets/4t8vs4dr4v/1}{Link}\\
         &  KoEBA \cite{hwang2023discohead}& 2023& 104(P)& Video-Text-Audio& Korea & \href{https://github.com/deepbrainai-research/koeba}{Link}\\
         
    \bottomrule
    \multirow{10}*{Others}
                & AV Letters  \cite{matthews2002extraction}&2002&$\sim$19k  &Video-Text&English&\href{http://www.ee.surrey.ac.uk/Projects/LILiR/datasets/avletters1/index.html}{Link}\\ 
                & AV Digits  \cite{petridis2018visual}&2002&$\sim$5k  &Video-Text&English&\href{https://ibug-avs.eu/}{Link}\\  
                % & GLips \cite{GLips}&2018&250K  &Video-Text&German&\href{https://www.fdr.uni-hamburg.de/record/10048/export/hx}{Link}\\         
    % \bottomrule
    %     \multirow{6}*{Gesture Generation} 
              % ~  &  KIT-ML\cite{kit} &2016&3,911&Video-Text-Skelton&English&\href{https://motion-annotation.humanoids.kit.edu/dataset/}{Link}\\
             & Aoyama Gakuin \cite{takeuchi2017creating} &2017&$\sim$1k &Videos-Text-Audio-Skelton(2D)&Japanese& Not Available\\
                     ~  & P2PSTORY\cite{singh2018p2pstory} &2018&$\sim$13k&Video-Text-Audio&Multiple&\href{https://www.media.mit.edu/projects/p2pstory/overview/}{Link}\\
                        & AMASS\cite{mahmood2019amass} &2019&$\sim$18k &video-text-Skelton(3D)&English&\href{https://amass.is.tue.mpg.de/download.php}{Link}\\ 
            % \multirow{3}*{Body Language} 
        & BoLD\cite{luo2020arbee} &2020&$\sim$10k &Video-Text-Audio-Skelton(3D)&English&\href{https://cydar.ist.psu.edu/emotionchallenge/index.php}{Link}\\
         & PATS \cite{ahuja2020no} &2020&$\sim$84k&Videos-Text-Audio-Skelton(2D)&English&\href{https://chahuja.com/pats/}{Link}\\  
                & BABEL\cite{punnakkal2021babel} &2021&$\sim$28k &video-text-Skelton(3D)&English&\href{https://babel.is.tue.mpg.de/data.html}{Link}\\ 
         & HumanML3D\cite{guo2022generating} &2022&$\sim$15k &video-text-Skelton&English&\href{https://github.com/EricGuo5513/HumanML3D}{Link}\\  
        & BEAT\cite{liu2022beat} &2023&$\sim$3k &Video-Text-Audio-Skelton(3D)&Multiple&\href{https://pantomatrix.github.io/BEAT-Dataset/index.html}{Link}\\
    \bottomrule
  \end{tabular}}
  \end{center}
\end{table*}

\subsection{Talking Head}
TH refers to a virtual or digital representation of the human face or head, typically used in multimedia applications, computer graphics, and HCI. It is usually an animated character that appears on a screen and can simulate various facial expressions, head actions, and speech with synchronized lip movements \cite{Chen2020lrsd,wang2023emotional,wang2023memory}. TH aims to enhance user experiences in various applications, from virtual assistants to entertainment platforms, by providing interactive and immersive communication interfaces.

In 2003, visual text-to-speech (VTTS) that generates talking faces driven by a speech synthesizer has been proposed for HCI systems \cite{cosatto2003lifelike}. Speech synthesis techniques are used to convert text input into synthesized speech, allowing the virtual character to speak. Facial animation algorithms are employed to animate the virtual character's facial movements, including lip synchronization with the generated speech. These algorithms analyze the phonetic information in the speech and map it onto corresponding facial movements. Additionally, sophisticated computer graphics techniques are utilized to generate realistic textures, lighting, and shading for the virtual character, enhancing its visual appearance\cite{gambino2008virtual}. Previous approaches to TH generation faced many limitations and were unable to achieve high-quality and realistic results due to constraints like limited data availability and computing power. 

TH generation needs to fuse and synchronize information from different modalities to ensure consistency and coherence between the animation, sound, and text of the character. This involves the alignment, fusion, and synchronization of each modal to produce a more uniform response. In recent years, DL and multi-modal neural networks advance the performance of TH generation from multiple perspectives \cite{zhen2023human,song2023virtual}. 
By using multi-modal or cross-modal techniques based on a large amount of data, TH generation can integrate user input from different sources and interact in a more natural and realistic way. This enables multi-modal HCI systems to better understand user intentions, generate responses accordingly, and provide a more immersive and personalized interactive experience.

In Section \ref{ssec:thr}, we give a brief review of TH recognition. In Section \ref{ssec:thg}, we explore the applications and advancements in deep multi-modal learning for TH generation. We discuss the challenges associated with creating realistic and expressive virtual characters, including the synthesis of natural-sounding speech and the accurate representation of facial expressions. We review the SOTA techniques and highlight the potential future developments in this field, aiming to improve the realism and interactivity of THs in various applications.

\section{Body Language Dataset and Evaluation Metrics}

\subsection{Body Language Datasets}
Datasets have played a crucial role in the entire history of BL research, serving as a common foundation not only for measuring and comparing the performance of competing algorithms but also for driving the field toward increasingly complex and challenging problems. Particularly in recent years, DL techniques have brought significant success to BL research, with a substantial amount of annotated data being key to this success. The availability of large-scale image collections through the internet has made it possible to construct comprehensive datasets. Additionally, the availability of multi-modal data has provided richer information for related tasks, opening up new possibilities for future BL recognition and generation research.

\begin{figure*}[htb]
\begin{minipage}[b]{1.0\linewidth}
 \centering
 \centerline{\includegraphics[width=19.0cm]{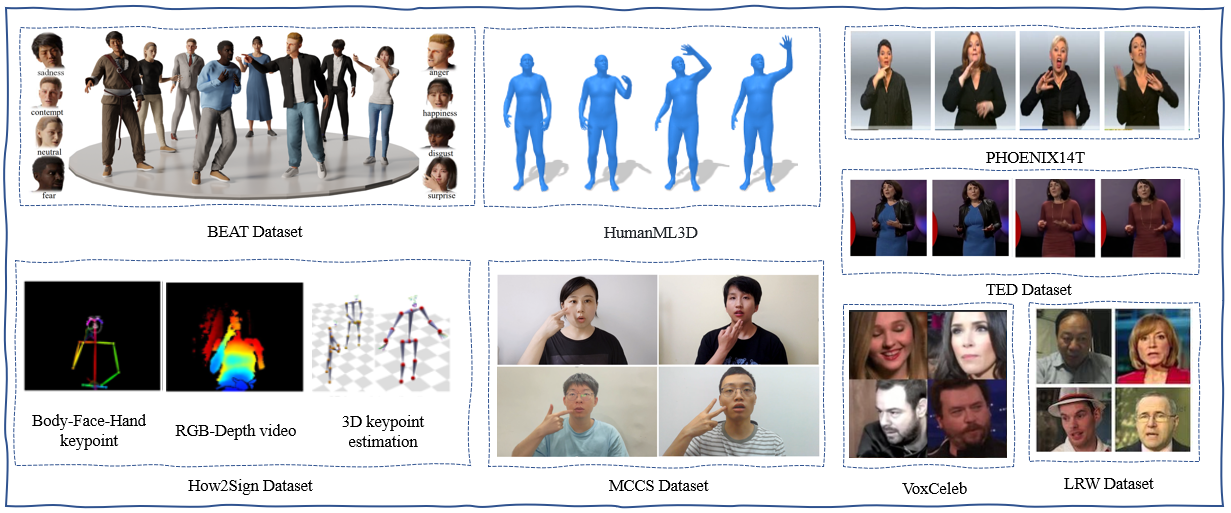}}
\end{minipage}
\caption{Some examples of BL datasets.}
\label{fig:datasets}
\end{figure*}

In this section, we have collected and presented relevant datasets pertaining to BL tasks. As shown in Table \ref{tab:dataset}, we have categorized them into five types based on data format and task purposes: CS, SL, CoS, TH, and others (Here ``others" means these datasets are multi-modal BL datasets but are not designed for these four tasks). We have introduced the relevant information about these datasets, including publication year, dataset scale, available modalities, and the languages used in the datasets. Moreover, we have provided official links to these datasets to facilitate easier access for researchers. Please note that we measure the dataset scale based on the number of video clips/sequences. For datasets that do not provide these numbers, we provide the duration of the videos in minutes (represented as "Min") or the number of performers (represented as "P"). Some examples of BL datasets are shown in Figure \ref{fig:datasets} for more details.

\begin{figure}[t]
    \centering
    \includegraphics[width=0.4\textwidth]{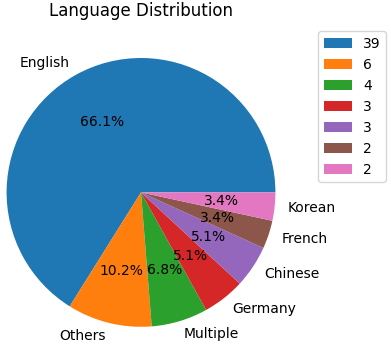}
    \caption{The Distribution of language used in BL datasets.}
    \label{fig:distribution}
\end{figure}

We present the distribution of dataset languages in Figure \ref{fig:distribution}. The chart shows the related datasets are primarily English datasets, but it also includes datasets in other languages like English datasets, Chinese datasets, and German datasets. This illustrates that current BL research is predominantly focused on English, but there is also growing importance placed on cross-cultural and multilingual datasets.
Another problem is the difference in the format and standards of BL datasets. Different datasets may have varying storage formats, recording requirements, and model standards.

\subsection{BL Generation Metrics}
In order to evaluate the performance of BL gesture generation methods, we summarize the main gesture generation metrics and show these generation metrics, and corresponding calculation formulas in Table \ref{tab:gesturemetrics}.  A total of seven metrics are introduced for evaluation, namely PCK\cite{yang2012articulated}, FGD\cite{yoon2020speech}, MAE\cite{asakawa2022evaluation}, STD\cite{asakawa2022evaluation}, PMB\cite{ao2022rhythmic}, MAJE\cite{yoon2020speech}, and MAD\cite{yoon2020speech}. Percentage of Correct Keypoints (PCK) assesses the accuracy of generated motion by comparing keypoints with actual motion. A predicted keypoint is considered correct if it falls within a specified threshold of the actual keypoints. Mean Absolute Error (MAE) quantifies the average difference between standardized coordinate values of generated and actual keypoints. Standard Deviation (STD) represents the variability or distribution of keypoints from their mean position after standardization. Fréchet Gesture Distance (FGD) measures dissimilarity between the distributions of latent features in generated and ground truth gestures, incorporating both location and spread. Percentage of Matched Beats (PMB) considers a motion beat matched if its temporal distance to an audio beat is below a threshold. Mean Absolute Joint Error (MAJE) calculates average errors between generated and ground truth joint positions across all time steps and joints. Mean Absolute Difference (MAD) computes average differences in joint accelerations, considering magnitude and direction. These criteria provide comprehensive insights into the accuracy, similarity, and alignment between generated and ground truth motion data.

The TH generation results can be evaluated quantitatively from multiple perspectives. Evaluation metrics include identity-preserving metrics, audio-visual synchronization metrics, image quality-preserving metrics, expression metrics, and eye-blinking metrics.

\begin{figure*}[htbp]
    \centering
    \includegraphics[width=0.9\textwidth]{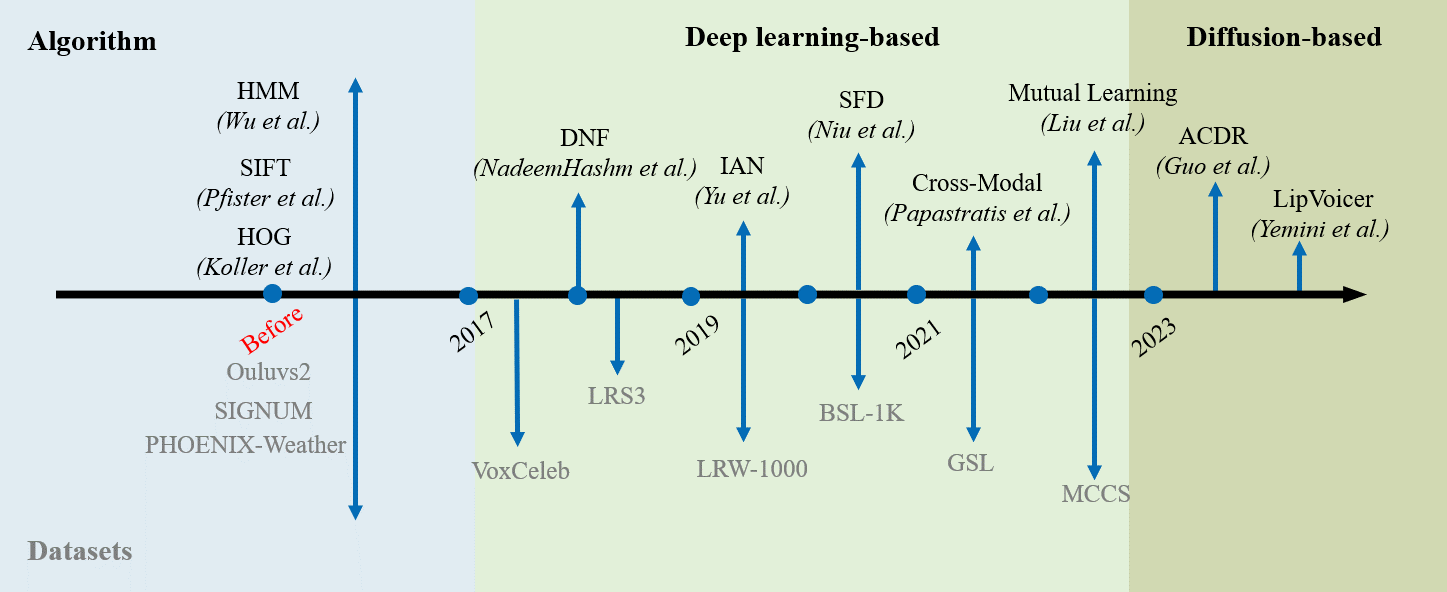}
    \caption{The milestones of Datasets and Methods for BL recognition.}
    \label{fig:timeline-BL-recognition}
\end{figure*}

\section{Automatic Body Language Recognition}
%\textbf{Xiaotian}
Here, we will introduce the recognition for the four BL variants, with a particular focus on the application expansion and innovation of multi-modal learning. In Figure \ref{fig:timeline-BL-recognition}, we present a summary of some representative works for BL recognition.
\begin{figure*}[t]
    \centering
    \includegraphics[width=0.9\textwidth]{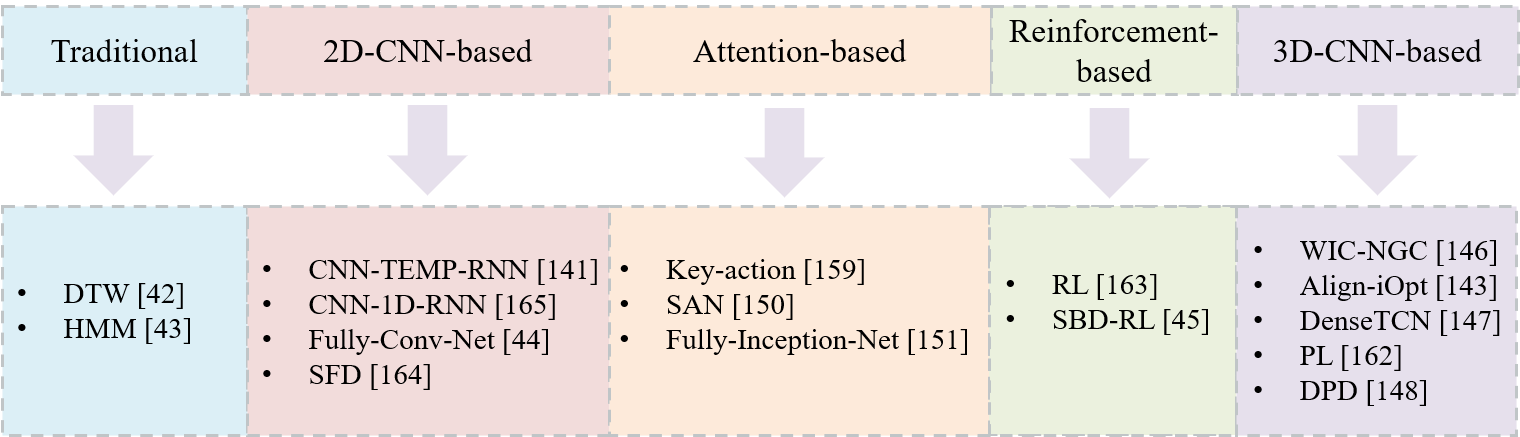}
    \caption{The comprehensive methods of SL recognition.}
    \label{fig:timeline-SLR}
\end{figure*}

\subsection{Sign Language Recognition}
\label{ssec:slr}
SLR aims to utilize a classifier to recognize SL glosses from video streams. It can be classified into two types in general according to the content of the SL: continuous SL recognition and isolated SL recognition. In this paper, we focus on continuous SL recognition (CSLR), in which the feature encoder module first extracts semantic representations from the sign video, and then the sequential module performs the mapping from the extracted semantics to the text sequence. In addition, some training strategies have been investigated for sufficient training. The comprehensive approaches are presented in Figure \ref{fig:timeline-SLR}. 

\textbf{Feature Encoder.} Since the hand acts a dominant role in the expression of SL, it has evolved over the past three decades, and we can divide these methods into the two following types. 
\begin{itemize}
    \item \textbf{Handcraft-based Method.} In the early research, handcrafted features are used to extract the hand motion, shape and orientation, such as HOG\cite{koller2015continuous, buehler2009learning}, Grassmann covariance matrix (GCM)\cite{wang2019novel} and SIFT\cite{pfister2013large}. However, these methods require manual feature extraction and cannot directly be applied to different gestures, which means that different gestures necessitate distinct feature extraction approaches, resulting in a substantial amount of work. 
    \item \textbf{CNN-based Method.} With the development of DL, CNNs\cite{he2016deep,carreira2017quo,qiu2017learning,qiu2019learning} generally replace the handcraft-based methods, becoming the most powerful feature extractor for SLR. Many researchers try to explore the reasonable CNN-based architecture to directly extract discriminative visual features from the video sequence. Specifically, exist works used 2D-CNN-TCN\cite{hu2022collaborative, cui2019deep,pu2020boosting, cheng2020fully} and 3D-CNN\cite{hu2022collaborative,pu2019iterative,huang2018video,hu2021global, wei2019deep,guo2019dense,zhou2019dynamic} as the backbone to extract spatial-temporal discriminative cues. For instance, IAN\cite{pu2019iterative} utilizes 3D-ResNet\cite{qiu2017learning} for visual representation. DNF\cite{nadeemhashmi2018lip} subtly designs 2D-CNN with the 1D temporal convolution, which has become one of the mainstream baseline methods. Although CNN-based methods can effectively capture spatial features in gesture images, they are limited in handling the temporal dynamics of gestures directly, and 3D-CNN-based methods involve significant computational overhead.

\end{itemize}

\textbf{Sequential Module.} There exist three representative approaches\cite{slimane2021context,zhou2020self} for CSLR. In the early research, HMM\cite{koller2019weakly,koller2017re,wu2016deep,koller2018deep} is used to learn the correspondence between the visual representation and sign gloss sequence. However, gesture actions of SLR often have long-term dependencies, and HMM struggles to capture such complex sequential patterns. Additionally, HMM does not consider the alignment between input and output modalities. To this end, the RNN-based methods with CTC loss\cite{cui2017recurrent,min2021visual,cui2019deep,pu2019iterative,cihan2017subunets} are developed for CSLR to replace the HMM model, which improves the model's ability to handle data with incomplete alignments, but the ability to model global information is still limited. Therefore, to better understand the semantic relationship of the entire sign language sequence, encoder-encoder\cite{guo2018hierarchical,guo2019hierarchical,li2020key} has become a commonly used sequential framework. For instance, Guo et al.\cite{guo2018hierarchical} utilizes the encoder-decoder framework with hierarchical deep recurrent fusion to merge cues from RGB and skeleton modalities.

\textbf{Training Strategy.} For sufficient training, some optimization strategies are widely used, with the most prominent being CTC\cite{hao2021self,min2021visual,zhou2021spatial} and Iterative Training\cite{cui2019deep,pu2020boosting,pu2019iterative,zhou2021spatial} strategies. On top of these two strategies, Pu et al. \cite{pu2020boosting} 
 introduce a cross-modality constraint called CMA to aid the training. Hao et al. \cite{hao2021self}  propose a three-stage optimization approach, which improves the recognition performance but it is time-consuming. Recently, Min et al. \cite{min2021visual} further present two auxiliary constraints over the frame-level probability distributions, making the entire model end-to-end trainable.

\begin{table*}[htbp]
  \begin{center}
  \caption{The timeline of some representative works for BL recognition.}
    \renewcommand{\arraystretch}{1.5} % 调整行间距
  \setlength{\tabcolsep}{10pt} % 调整列间距
  \small % 调整文本大小
    \label{tab:recognitionwork}
    \begin{tabular}{m{1.5cm}<{\centering}|m{1cm}<{\centering} |m{2.3cm}<{\centering} |m{1.6cm}<{\centering} |m{1.8cm}<{\centering} |m{3.2cm}<{\centering} |m{3.2cm}<{\centering}}
      \hline \textbf{Type} & \textbf{Year} & \textbf{Ref} & \textbf{Feature Extraction} & \textbf{Sequence Model} & \textbf{Learning Paradigm} & \textbf{Dataset} \\
      \hline
      \multirow{12}{*}{SL} 
       & 2019 & Pei et al. \cite{pei2019continuous} & 3D-ResNet
       & BGRU & CTC & Phoenix-2014\\ \cline{2-7}
       & 2019 & Pei et al. \cite{zhang2019continuous} & 3D-ResNet
       & Transformer & Reinforcement Learning & Phoenix-2014\\ \cline{2-7}
       & 2019 & Cui et al. \cite{cui2019deep} & CNN 
       & RNN & Iterative Training & Phoenix-2014 and SIGNUM \\ \cline{2-7}
       & 2020 & Niu et al. \cite{niu2020stochastic} & CNN 
       & Transformer & CTC & Phoenix-2014 \\ \cline{2-7}
       & 2020 & SAFI \cite{zhou2020self} & 2D-CNN \emph{plus} 1D-CNN & SAN & ACE \emph{plus} CTC & Phoenix-weather and SIGNUM \\ \cline{2-7}
       & 2021 & Koishybay et al. \cite{koishybay2021continuous} & 2D-CNN \emph{plus} 1D-CNN & RNN & Iterative GR \emph{plus} CTC & Phoenix-weather and SIGNUM \\ \cline{2-7}
       & 2021 & SLRGAN \cite{Papastratis2021Continuous} & CNN & BiLSTM & GAN  & Phoenix-weather, CSL and GSL \\ \cline{2-7}
       & 2022 & Chen et al. \cite{chen2022two} & S3D & BLC & CTC \emph{plus} Self-distillation  & Phoenix-2014 \\ \cline{2-7}
       & 2022 & Zhou et al. \cite{zhou2021spatial} & SMC 
       & BiLSTM \emph{plus} SA-LSTM & CTC \emph{plus} Keypoint Regression & PHOENIX-2014, CSL and PHOENIX-2014-T   \\ \cline{2-7}
       & 2023 & Hu et al. \cite{hu2023self} & 2D-CNN 
       & 1D-CNN \emph{plus} BiLSTM & SSTM \emph{plus} TSEM & PHOENIX-2014, PHOENIX-2014-T, CSL and CSL-Daily   \\ \cline{2-7}
       & 2023 & Zheng et al. \cite{zheng2023cvt} & CNN 
       & VAE & CTC \emph{plus} Contrastive Alignment Loss & PHOENIX-2014 and PHOENIX-2014-T   \\ 
       \hline
      \multirow{9}{*}{CS} 
       & 2018 & Liu et al. \cite{liu2018visual} & CNN 
       & HMM & - & French CS \\ \cline{2-7}
      & 2021 & Papadimitriou et al. \cite{papadimitriou2021fully} & 2D-CNN \emph{plus} 3D-CNN & Attention-based CNN & - & French and British English CS \\ \cline{2-7}
     & 2021 & Liu et al. \cite{liu2020re} & CNN 
     & MSHMM & HPM & French and British English CS \\ \cline{2-7}
      & 2021 & Wang et al. \cite{wang2021cross} &  CNN \emph{plus} ANN & BiLSTM \emph{plus} FC & Cross-Modal Knowledge Distillation & French and British English CS \\ \cline{2-7}
      & 2022 & Sankar et al. \cite{sankar2022multistream} &  Bi-GRU & Bi-GRU & CTC & CSF18 \\ \cline{2-7}
      & 2023 & Liu et al. \cite{liu2023cross} &  ResNet-18 & Transformer & Cross-Modal Mutual Learning & Chinese, French, and British English CS\\ 
      \hline
   
    \end{tabular}
  \end{center}
\end{table*}

\begin{table*}[htbp]
  \begin{center}
  \caption{The timeline of SL generation works.}
    \renewcommand{\arraystretch}{1.5} % 调整行间距
  \setlength{\tabcolsep}{10pt} % 调整列间距
  \small % 调整文本大小
    \label{tab:generationwork-SL}
    \begin{tabular}{m{1.5cm}<{\centering}|m{1cm}<{\centering} |m{1.8cm}<{\centering} |m{1.6cm}<{\centering}|m{1.8cm}<{\centering} |m{2cm}<{\centering} |m{4.5cm}<{\centering}}
      \hline \textbf{Type} & \textbf{Year} & \textbf{Ref} & \textbf{Input Modality} & \textbf{Framework} & \textbf{Dataset} & \textbf{Description} \\
      \hline
      \multirow{21}{*}{SL} 
      & 2011 & kippet et al. \cite{kipp2011sign} & RCB video& EMBR & ViSiCAST & A gloss-centric tool is proposed to enable the comparison of avatars with human signers. But it is necessary to incorporate non-manual features. \\ \cline{2-7}
      & 2016 & John et al. \cite{mcdonald2016automated} &  RGB video & Segmental framework & Own dataset & This approach achieves automatic realism in generated images with low complexity, but it requires positioning the shoulder and torso. \\ \cline{2-7}
      & 2016 & Sign3D\cite{gibet2016interactive} &  RGB video&Heterogeneous Database & Own dataset & 
This approach guarantees sign avatars that are easily understood and widely accepted by viewers, but it is restricted to a limited set of sign phrases. \\ \cline{2-7}
      % & 2018 & NSLT\cite{camgoz2018neural} &  RGB video & PHOENIX14 T & This approach demonstrates robustness in jointly aligning, recognizing, and translating sign videos. However, a drawback is the requirement to align the signs in the spatial domain. \\ \cline{2-6}
      & 2018 & HLSTM\cite{guo2018hierarchical} &  RCB video &LSTM & Own dataset & This approach shows robustness in effectively aligning the word order with visual content in sentences. Nevertheless, a limitation arises when generalizing it to new datasets. \\ \cline{2-7}
      & 2020 & Text2Sign\cite{Stephanie2020text2sign} & Text &Transformer & PHOENIX14T & It demonstrates robustness in handling the dynamic length of the output sequence. However, It did not incorporate nonmanual information. \\ \cline{2-7}
      % & 2020 & Saunders et al. \cite{saunders2020progressive} &  Text & PHOENIX14 & This approach shows robustness when trained with minimal gloss and skeletal level annotations. However, The model suffers from a high complexity. \\ \cline{2-6}
      & 2020 & Zelinka et al. \cite{Zelinka2020Neural} &  Text&CNN & Crech news & This method is robust to missing part, but face expression is not included. \\ \cline{2-7}
      & 2020 &  ESN\cite{saunders2020everybody} &  Text&GAN & PHOENIX14T & It shows Robustness to non-manual feature generation. But the genrated signs are not realistic . \\ \cline{2-7}
      & 2020 & Necati et al. \cite{camgoz2020multichannel} &  Text&Transformers & PHOENIX14T & It does not need the gloss information, but the model is complex \\ \cline{2-7}
      & 2020 & Saunders et al. \cite{Saunders2020AdversarialTF} &  Text& GAN & PHOENIX14T & Robust to manual feature generation. The generated signs are not realistic.\\ \cline{2-7}
       & 2022 & DSM. \cite{i̇nan2022modeling} &  Gloss& Transformer & PHOENIX14T & This work improves the prosody in generated Sign Languages by modeling intensification in a data-driven manner.\\ \cline{2-7}
      & 2022 & SignGAN. \cite{Saunders2022signgan} &  Text& FS-Net & meineDGS & It tackles large-scale SLP by learning to co-articulate between dictionary signs and improves the temporal alignment of interpolated dictionary signs to continuous signing sequences\\ \cline{2-7}
      & 2023 & PoseVQ-Diffusion. \cite{xie2022vector} &  Gloss& CodeUnet & PHOENIX14T & It proposes a vector quantized diffusion method for conditional pose sequences generation and develops a novel sequential k-nearest-neighbors method to predict the variable lengths of pose sequences for corresponding gloss sequences\\ \hline
    \end{tabular}
  \end{center}
\end{table*}

\begin{table*}[htbp]
  \begin{center}
  \caption{The timeline of Co-speech  and Cued Speech generation works.}
    \renewcommand{\arraystretch}{1.5} % 调整行间距
  \setlength{\tabcolsep}{10pt} % 调整列间距
    \label{tab:generationwork-CoS}
  \small % 调整文本大小
    \begin{tabular}{m{1.5cm}<{\centering}|m{1cm}<{\centering} |m{2cm}<{\centering}|m{1.8cm}<{\centering} |m{2cm}<{\centering} |m{1.8cm}<{\centering} |m{4.3cm}<{\centering}}
      \hline \textbf{Type} & \textbf{Year} & \textbf{Ref} & \textbf{Input Modality}& \textbf{Framework} & \textbf{Dataset} & \textbf{Description} \\
      \hline
 \multirow{20}{*}{ CoS }
      & 2015 & DCNF\cite{chiu2015predicting} &  Text&  FC network & DIAC & This work integrated speech text, prosody, and part-of-speech tags to generate co-verbal gestures using a combination of FC networks and a Conditional Random Field (CRF). \\ \cline{2-7}
      & 2019 & S2G\cite{ginosar2019learning} &   RGB video& CNN & S2G & This work presents a method for generating gestures with audio speech, utilizing cross-modal translation and training on unlabeled videos. But it relies on noisy pseudo ground truth for training \\ \cline{2-7}
      % & 2019 & DRAM\cite{ahuja2019react} &Pose & Own dataset &  It generates natural gesticulation for avatars in a telepresence setting, taking into account the history of both the avatar and the human interlocutor. \\ \cline{2-6}
      & 2020 &StyleGestures \cite{alexanderson2020style} &  RCB video& LSTM & Trinity & It achieves natural variations without manual annotation and allows control over gesture style while maintaining perceived naturalness. \\ \cline{2-7}
      & 2021 & A2G\cite{li2021audio2gestures} &  Text& CVAE & Trinity & This work employed a CVAE to generate diverse gestures from speech input and involved a one-to-many mapping of speech-to-gesture. \\ \cline{2-7}
      & 2021 & Text2Gestures\cite{bhattacharya2021text2gestures} &  Text& Transformer & MPI-EBEDB & Their approach employed Transformer-based encoders and decoders to generate sequential joint positions based on the text and previous pose. \\ \cline{2-7}
      & 2022 & ZeroEGGS\cite{ghorbani2022exemplar} &  Text& Variational Framework & Own dataset &  A VAE-based framework is utilized to generate style-controllable CoS gestures and allowed for the generation of stylized gestures by conditioning on a zero-shot motion example \\ \cline{2-7}
      % & 2022 &  SEEG\cite{Liang2022seeg} &  Text & TED & Pros and cons \\ \cline{2-6}
      & 2022 & DiffGAN\cite{Ahuja2022low} &  Text& Diffusion Model & PATS &   An adversarial domain-adaptation approach is proposed to personalize the gestures of a speaker \\ \cline{2-7}
      & 2022 & RG\cite{ao2022rhythmic} &  Trinity and TED& QVAE & PHOENIX14T & This work introduces a novel CoS gesture synthesis method that effectively captures both rhythm and semantics. \\ \hline      
 \multirow{6}{*}{ CS }
      & 1998 & Paul et al. \cite{duchnowski1998automatic} &  Text& Template & Own dataset &  Relying on manually selected keywords, low-context sentences, and pre-defined gesture templates. Its limitations included constrained expressiveness and increased manual effort. \\ \cline{2-7}
      & 2008 &G{\'e}rard et al. \cite{bailly2008retargeting} &   RGB video& Template & Own dataset &  A post-processing algorithm was introduced to fine-tune synthesized hand gestures by addressing rotation, translation, and adaptation to new images. However, it relies on prior knowledge for adapting to new images. \\ \hline
    \end{tabular}
  \end{center}
\end{table*}

\subsection{Cued Speech Recognition}
\label{ssec:csr}
% Cued Speech (CS) was first proposed by Dr. Cornett\cite{cornett1967cued}, aiming to help hearing-impaired people communicate better. It uses unique hand coding to assist lip reading, reducing errors in the process of lip reading.
Automatic lip reading is a crucial component of ACSR. Therefore, we will first introduce the research progress in automatic lip-reading and then review ACSR.
\subsubsection{Automatic Lip Reading}
Advances in DL have led to a promising performance in lip-reading methods. Generally, DL-based lip-reading methods consist of two main parts, one is the extraction of visual feature information, and the other is the classification of sequence features.

\textbf{Feature Extraction.} 
Traditional studies use pixel-based\cite{li2008novel}, shape-based\cite{alizadeh2008lip, ma2016lip}, and hybrid-based\cite{lan2012view, watanabe2017lip,liu2017inner,liu2016extraction,wang2021three} approaches to extract the visual feature. However, these methods are not only sensitive to image illumination change, lip deformation, and rotation but also cannot extract automatically.

Recently, DL has gradually become the mainstream research method in lip visual feature extraction, which can be divided into four categories. First, 2D-CNN-based methods are used\cite{garg2016lip, lee2016multi}, which solves the problem of automatic feature extraction, but it can only process single-frame images and has a weak ability to process continuous frames, ignoring the spatio-temporal correlation between continuous frames. Then, 3D-CNN-based methods have received extensive attention\cite{fung2018end, xu2018lcanet, wiriyathammabhum2020spotfast, weng2019learning}. Although this method can solve the problem of spatio-temporal correlation of continuous frames, it loses the extraction of fine-grained feature information by 2D convolution to a certain extent. According to the aforementioned issues, the hybrid methods \cite{feng2021efficient, xu2020discriminative, luo2020pseudo} of 2D-CNN and 3D-CNN are also introduced to solve the problem of spatio-temporal feature extraction and local fine-grained feature extraction simultaneously. This method utilizes 3D-CNN to extract spatio-temporal information and then directly accesses 2D-CNN to extract fine-grained local information. However, it still affects the time information of feature coding to some extent. For that purpose, some other neural networks have gradually become a popular choice for lip visual feature extraction, such as Autoencoder model\cite{gehring2013extracting,noda2015audio,petridis2016deep,wang2021attention}. 

\textbf{Recognition Modeling.} So far, there have been many works viewing lip reading as a sequence-to-sequence task and using sequence-based methods to deal with it, such as RNN, LSTM, and Transformer. It divides the feature representations extracted from the feature extractor into equal time steps, feeding each of them sequentially to the classification layer. For instance, \cite{wand2018investigations,weng2019learning,zhang2020can,xiao2020deformation,luo2020pseudo,zhao2020mutual} utilize Long-Short Term Memory (LSTM) networks and Gated Recurrent Unit (GRU) to capture both global and local temporal information. Considering that Temporal Convolutional Network (TCN) has the advantage of faster converging speed with longer temporal memory than LSTM or RNN models, it is also widely used in this task.  For example, Bai et al.\cite{bai2018empirical} first propose a simple yet effective TCN architecture, indicating that TCN can become a reasonable alternative to RNN as a sequential model. Following this work, Martinez et al.\cite{martinez2020lipreading} further demonstrate that multi-scale TCN can outperform RNN in lip reading isolated words. However, these methods are relatively weak in modeling long-term dependencies and cannot directly capture long-term dependencies in sequences. Therefore,  a new trend in the use of Transformer\cite{vaswani2017attention} for lip-reading tasks has emerged\cite{afouras2018deepLR, ma2021end}. 

Although the aforementioned methods achieve promising performance, they cannot solve the problem of inconsistency between the input and the output modality for lip reading. For that purpose, many advanced works are further developed in recent years, such as attention mechanisms\cite{son2017lip,chung2017lip,xu2018lcanet,afouras2018deepLR, lu2019automatic, zhou2019modality, zhang2019understanding} and contrastive learning\cite{torfi20173d}.

\subsubsection{Automatic Cued Speech Recognition}
The literature on ACSR can be classified into three main categories: Multimodal Feature Extraction, Multimodal Fusion, and ACSR Modeling. We discuss them separately in this section and
review the representative works of CS in Table \ref{tab:recognitionwork}

\textbf{Multi-modal Feature Extraction.} In the literature, there are several popular methods for CS feature extraction (\ie lips, hand position and hand shape).
\begin{itemize}
\item \textbf{Traditional Method.} It uses artificial markings to record lips and hands from video images\cite{heracleous2010cued,heracleous2012continuous}. For example, Burger et al.\cite{burger2005cued} let the speaker wear black gloves to obtain accurate hand segmentation, while Noureddine et al.\cite{liu2020re} placed blue marks on the speaker's fingers to obtain the coordinates of the fingers. However, both the speaker’s clothing color and the background color can affect the accuracy of the hand segmentation.
\item \textbf{CNN-based Method.} Recently, some CNN-based methods are utilized to get rid of artificial markings. For example, the CNN model is used in \cite{liu2018visual, liu2020re, papadimitriou2021fully} to extract visual features from the regions of the lip and hand. On the basis of using the CNN model for the feature extraction of lips and hand shape, Liu et al. \cite{liu2020re} further adopt the artificial neural network (ANN) to process the hand position feature. However, although CNN-based methods do not require artificial marks, their performances are limited by data scarcity.
\item \textbf{Other DL-based Method.} Considering the data-hungry problem for multi-modal, some researchers try to introduce some advanced methods to solve this issue. For instance, Wang et al. \cite{wang2021cross} use lips, hand shape, and hand position to pre-train multi-modal feature extractor, using it for feature extraction of ACSR task. In addition, in another of their work \cite{wang2021attention}, the three-stage multi-modal feature extraction model based on self-supervised contrastive learning and self-attention mechanism is proposed to model spatial and temporal features of CS hand shape, lips, and hand position.

\end{itemize}

\textbf{Multi-modal Fusion.} Most existing works in ACSR tend to direct concatenate the multi-modal feature flows, letting the model learn such features implicitly \cite{heracleous2012continuous,wang2021attention,liu2018visual,papadimitriou2021fully}. For instance, \cite{heracleous2012continuous, wang2021attention} utilize artificial marks to obtain regions of interest (ROIs) and directly concatenated features of lip and hand. MSHMM\cite{liu2018visual} merges different features by giving weights for different CS modalities. However, to the best of our knowledge, a critical issue in ACSR is the asynchrony between hand and lip articulations \cite{liu2020re, liu2018automatic,gao2023novel}, while these researches mainly assume lip-hand movements are synchronous by default, ignoring the asynchronous issue.

Therefore, to tackle asynchronous modalities in the ACSR task, Liu et al. \cite{liu2020re} propose to utilize the re-synchronization method to align the hand and lips features, which is realized by introducing the prior knowledge of the hand position and hand shape. Nevertheless, since the acquisition of prior knowledge depends on speakers and specific
datasets, it is difficult to directly apply it to other languages. For that purpose, Liu et al. \cite{liu2023cross} further propose a Transformer-based cross-modal mutual learning framework for multi-modal feature fusion. The framework captures linguistic information by constructing a modality-invariant shared representation and uses this linguistic information to guide cross-modal information alignment. Recently, \cite{zhang2023cuing} proposes a novel Federated CS recognition (FedCSR) framework to train an model of CS recognition in the decentralized data scenario. Particularly, they design a mutual knowledge distillation fusion mechanism to maintain cross-modal semantic consistency of the CS multi-modalities, which learning a unified feature space for both speech and visual feature. 

\textbf{ACSR Modeling.} ACSR aims to transcript 
visual cues of speech to text.  In the early research, traditional statistical methods are used, which map sequences of hand-crafted features to phonemes using statistical models, such as HMM \cite{heracleous2012continuous,heracleous2010cued} and HMM-GMM\cite{liu2018visual,liu2020re}. However, such methods only consider the relationships between the current state and the previous one, which means that longer contextual information cannot be captured.

More recently, traditional DL-based methods (\ie CNN-based, LSTM-based) have been developed to alleviate the aforementioned problem. For instance, Sankar et al.\cite{papadimitriou2021multimodal} propose a novel RNN model trained with a Connectionist Temporal Classification (CTC) loss \cite{graves2006connectionist}. Papatimitriou et al.\cite{papadimitriou2021fully} propose a fully convolutional model with a time-depth separable block and attention-based decoder. However, such traditional DL-based methods still cannot capture long-time dependencies well, while it would be desirable to capture global dependency \cite{vaswani2017attention} over dynamic longer because of the context relationships of phonemes in long-time CS videos. For that purpose, Transformer-based methods \cite{papadimitriou2021fully} receive a lot of attention on the ACSR task in recent years. This kind of method achieves promising performance on the ACSR task, but it still requires powerful computing resources and a large dataset for training and parameter tuning. 

Therefore, considering the existing corpus for ACSR is limited, some advanced methods such as cross-modal knowledge distillation method \cite{wang2021cross} and contrastive learning method \cite{wang2021attention}, are also introduced to this task.

%\begin{figure*}[t]
%    \centering
%    \includegraphics[width=0.9\textwidth]{CSR.png}
%    \caption{ The evolution of Cued Speech recognition.}
%    \label{fig:timeline-CSR}
%\end{figure*}

\subsection{Co-speech Recognition}
\label{ssec:cosp_r}
Although the existing research on CoS mainly focuses on the generation of CoS gestures, some scholars have shown that recognizing emotional expressions in CoS are crucial to this generation task. 
For example, Bock et al. \cite{boeck2014disposition} is the first to use the EmoGes corpus for emotion recognition in CoS gesture generation. Bhattacharya et al. \cite{bhattacharya2021speech2affectivegestures} proposed to leverage the Mel-frequency cepstral coefficients and the text transcript computed from the input speech in separate encoders in our generator to learn the desired sentiments and the associated affective cues.

\subsection{Talking Head Recognition}
\label{ssec:thr}
Since the development of the TH generation still has a long way to go, the focus of recent studies is not on TH recognition. TH recognition is primarily treated as an evaluation metric for TH generation algorithms. However, humans have the ability to detect and identify a person from their face, even when there are changes in gender or facial expressions. However, it is difficult to build an automatic face recognition system. Therefore, the focus of TH recognition is primarily on capturing the essential facial attributes of the target speaker rather than full-fledged recognition of the speaker’s identity.
In the work proposed by Wen et al. \cite{wen2019face}, they classified the face identity to assess the performance of voice-based face reconstruction for known subjects. For unknown subjects, they used a gender classifier to evaluate the gender of the generated faces. Additionally, the feature distance, such as Cosine, $L_1$, and $L_2$ distances, between the target face and the generated face can be calculated to measure the accuracy of the generated face. To achieve this, a pre-trained face recognition model like FaceNet \cite{schroff2015facenet} or ArcFace \cite{deng2019arcface} is employed as a feature extractor. The landmark distance (LMD) can also be measured as the disparity between the generated face and the real-world target face images.

\begin{figure*}[htbp]
    \centering
    \includegraphics[width=0.9\textwidth]{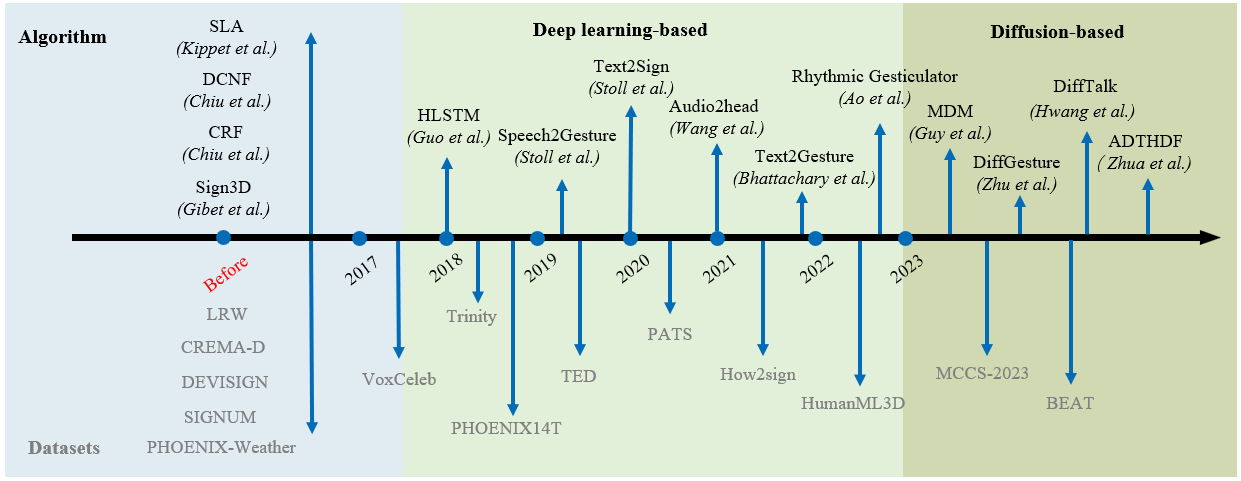}
    \caption{The milestones of Datasets and Methods for BL generation.}
    \label{fig:timeline-cospeech}
\end{figure*}

\begin{table}[ht]\small
    \begin{center}
    \setlength{\tabcolsep}{10pt} % 设置列间距为10pt
    \setstretch{2.1}
    \caption{Metrics for gesture generation.}
    \label{tab:gesturemetrics}
    \begin{tabular}{|c|c|} 
     \hline 
        \textbf{Metrics}  & \textbf{Calculation Formula} \\
        \hline 
        PCK \cite{yang2012articulated}  & $\text{PCK} = \frac{1}{N} \sum_{i=1}^{N}\textbf{1}(d_i \leq \tau){N} $\\
        FGD \cite{yoon2020speech} & $\text{FGD} = \max_{\pi} \left( \frac{1}{T} \sum_{t=1}^{T} d(g_t^*, g_{\pi(t)}) \right)$ \\
         MAE \cite{asakawa2022evaluation}  & $\text{MAE} = \frac{1}{T} \sum_{t=1}^{T} \left| g_t - g_t^{*} \right|$ \\
         STD \cite{asakawa2022evaluation} & $\text{STD} = \sqrt{\frac{1}{N} \sum_{i=1}^{N} (g_i - \bar{g})^2}$ \\
        PMB \cite{ao2022rhythmic} & $\operatorname{PMB}=\frac{1}{N_m} \sum_{i=1}^{N_m} \sum_{j=1}^{N_a} \textbf{1}\left[\left\|\boldsymbol{b}_i^m-\boldsymbol{b}_j^a\right\|_1<\delta\right]$ \\
        MAJE \cite{yoon2020speech} & $\text{MAJE} = \frac{1}{N \cdot T} \sum_{t=1}^{T} \sum_{n=1}^{N} \left| g_t^n - g_t^{*n} \right|$ \\
         MAD \cite{yoon2020speech} & $          \text{MAD} = \frac{1}{N \cdot T} \sum_{t=1}^{T} \sum_{n=1}^{N} \left| a_t^n - a_t^{*n} \right|_2$ \\
        \hline
        
    \end{tabular}
    \end{center}
            \footnotesize{\textit{The corresponding meanings for the alphabet are as follow: $N$ -- The number of samples; $d_i$ -- The distance of generated points and ground truth; $\tau$ -- The thresohold of PCK; $T$ -- the number of generated frames; $g_t$ -- the generated gestures; ${g_t}^{\*}$ -- the ground truth; $b_i$ -- The key frame corresponding to the beat. $a_i$ -- The movement aceleration of generated gestures. }}
\end{table}

\section{Automatic Body Language Generation}

% Sign Language, Cued Speech, Co-speech, \textbf{Wentao}

The gesture generation task aims to generate a continuous sequence of gestures (\ie face, head, and hand) using multi-modal inputs (\eg gloss, speech, and text). In this section, we present the related works on gesture language generation and review the development timeline of gesture language generation applications, such as CS, SL, CoS gesture generations, and TH Generation, respectively.

\subsection{Sign Language Generation}
\label{ssec:slg}
At the very beginning, we first present the difference between SL, CoS, and CS in Figure \ref{fig:elem_comp}. SL generation has been studied for a long time. In this part, we mainly discuss the DL-based research on SL generation. For other SL generation methods, please refer to \cite{rastgoo2021sign}. In Table \ref{tab:generationwork-SL}, we present a summary of the details of the related SL generation works.

\textbf{Multi-modal Feature Extraction.}
 As a special visual language, the inputs of the SL gesture generation task are not only text and speech but also SL Gloss. It is a marking system for recording SL words and phrases, usually using written symbols and short descriptions to represent gestures, mouth movements as well as other non-gesture features. SL Gloss is suitable for recording the content of SL in written form to facilitate learners to learn and understand SL expressions. Previous work \cite{Saunders2020AdversarialTF,saunders2020progressive} first converts spoken language to gloss and then uses gloss as input to extract features to generate SL gestures. Some work \cite{Zelinka2020Neural} use spoken language words and their characters as input to extract the word embedding of text, then the text features were used for gesture generation.

\textbf{Generative Methods.} For the SL generation task, there are several popular DL-based methods:
% 1) Avatar method. some Researchers explored the use of sign avatars to address communication barriers for the hearing-impaired. Sign avatars utilize 3D animated models to display signed conversations, accommodating different sign languages. Motion data from deaf individuals is captured using specialized cameras and sensors, and then transformed into sign avatars through computational methods. 
% : \cite{kipp2011sign,mcdonald2016automated,gibet2016interactive} is using 3D animation avators to display sign language, which can be more usable and acceptable for viewers. But it needs to solve problems such as unnatural gestures and lack of non-gesture information such as eye gaze and facial expressions. 
% \\
1) Neural Machine Translation (NMT) method, which \cite{camgoz2018neural,guo2018hierarchical,Stephanie2020text2sign} views the SL generation as a translation task. It uses the neural machine translation model to process SL text input, which can handle the output SL sequence of dynamic length but needs to solve problems such as domain adaptation. 2) Motion Graph method \cite{Stephanie2020text2sign} uses motion graphics technology to construct a directed graph from motion capture data and generate SL. This method can handle the continuity of SL, but it requires large scales of data and another challenge is the scalability and computational complexity of the graph to select the best transitions. 
3)
Conditional generation methods such as Generative Adversarial Networks (GAN) and Variational Auto-Encoders (VAEs) are also employed to generate SL videos. A hybrid model, including a VAE and GAN combination, has been proposed for the generation of people performing SL \cite{stoll2018sign,vasani2020generation}. However, the problems such as model complexity and video quality need to be solved. 
4) Other methods. In addition to the previous work, some research tries to introduce novel transformer-based model architectures for SLP. For example, \cite{saunders2020progressive} proposes a Progressive Transformers to generate continuous sign sequences from spoken language sentences. \cite{ventura2020can} combines a transformer with a Mixture Density Network (MDN) to manage the translation from text to skeletal pose. Although these works have brought performance improvements, the cost of the model complexity cannot be ignored.
% \textcolor{red}{In recent years, DL-based methods applied to generate SL video \cite{ventura2020can} were tained in supervised manner, thus they requires a large amount of annotated video data.
% In addition to creating skeleton visualizations, \cite{ventura2020can} have taken a step further by utilizing the cutting-edge human motion transfer technique to generate realistic videos.
% Another common approach is to use a gesture dictionary to generate SL video \cite{signdic}. In this approach, a dictionary of gestures is created for the SL gesture mapping. However, it lacks of flexibility and fluency in the generated video, and requires high cost of manual construction of the gesture dictionary.\\
% Current research on SL generation mainly focuses on Indian and German SL. \cite{isign} have developed a novel approach to utilize animation data by employing an intermediate 2D pose representation. This approach enables us to train a SL animation model that can effectively analyze real-world SL performances captured in video format. \cite{i̇nan2022modeling} introduces several strategies, based on the linguistic characteristics of SL, which inform how intensity modifiers can be accurately represented in gloss annotations. To the best of our knowledge, the studies on Chinese SL (CSL) are still limited. \cite{sign2014} used a rule-based manual and non-manual method to synthesize a context-dependent CSL animation. \cite{xiao2020skeleton} proposed a skeleton-based CSL recognition and generation framework with a recurrent neural network (RNN). } 

Even though the SL generation has made some progress, some challenges in the CSL generation are still unsolved, \ie 1) The SL relies on facial expression to identify the specific meaning and avoid ambiguity. But few works consider facial expressions. 2) The scale of the SL gestures library is very large. According to the official Chinese SL dictionary, there are about 5600 kinds of frequently used SL gestures. Most of the dataset only covers a small portion of all gestures, for example, \cite{xiao2020skeleton} builds a CSL dataset with 500 categories. The huge number of gestures brings a huge cost for the DL-based models to construct the mapping relationship.

\begin{figure*}[htb]
\begin{minipage}[b]{1.0\linewidth}
 \centering
 \centerline{\includegraphics[width=12.0cm]{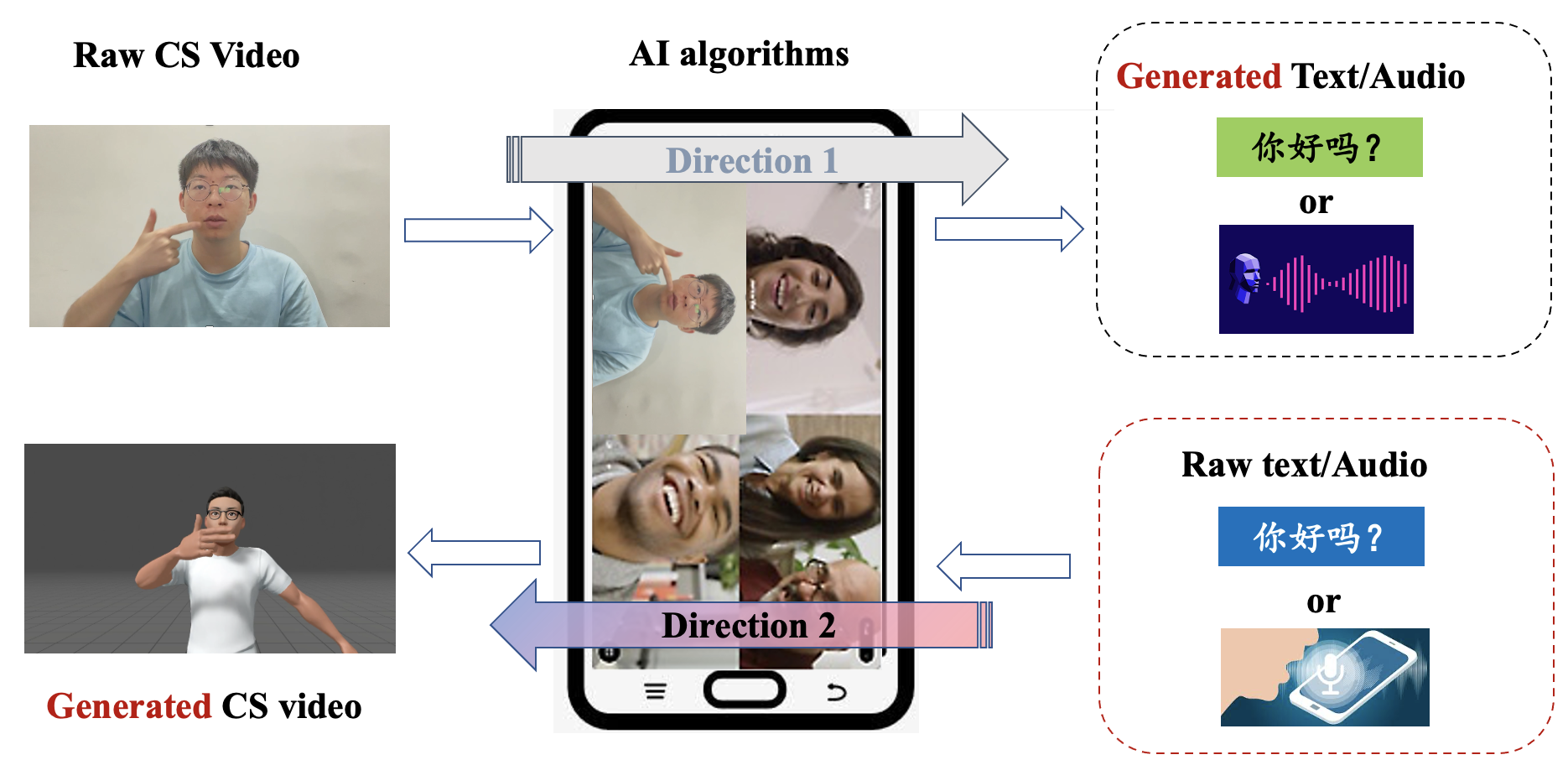}}
\end{minipage}
\caption{The overall framework of the conversion between CS and text/audio. Direction 1 means CS to text/audio recognition, and direction 2 means text/audio to CS gesture generation. The first direction aims to recognize text or audio to make normal hearing better understand the hearing-impaired people, and the second direction can help the hearing-impaired to visually understand normal-hearing people.}
\label{fig:Conversion_CS}
\end{figure*}

\subsection{Cued Speech Generation}
\label{ssec:csg}
As a lip-hand aided system, CS requires generating both lip and hand gestures simultaneously. Therefore, it is very important to extract multi-modal features such as speech features and text features. Among them, speech features have a strong correlation with lip movement. At the same time, text features play an important role in determining hand shape and position according to the coding system.
As depicted in Figure \ref{fig:Conversion_CS}, the generation of multi-modal CS hand gestures from audio-text is a crucial component of the CS conversion system. Previous studies in the literature have made limited initial attempts at CS gesture generation, which is mainly from two perspectives of multi-modal feature extraction and generation methods. Since the related work is relatively small, we incorporate the summary of the related CS generation works with the CoS In Table \ref{tab:generationwork-CoS}. 

\textbf{Multi-modal Feature Extraction.} For CS generation, the feature includes continuous lip shape and hand shape movements. \cite{duchnowski1998automatic} used specific manually selected keywords, along with low-context sentences \cite{rothauser1969ieee} as a feature, and pre-defined corresponding manual templates for hand gestures. CS recognition was performed, followed by the mapping of recognized text to the hand templates. However, this approach heavily relied on manual designs, which not only constrained the expressiveness of CS gestures but also increased the amount of manual effort required. 

\textbf{Generative Method.} To the best of our knowledge, there is still a lack of research on end-to-end deep learning-based CS gesture generation. Only \cite{duchnowski1998automatic} proposed a post-processing algorithm to adjust synthesized hand gestures, involving correction of hand rotation and translation, as well as adaptation of the algorithm to new images. Nevertheless, this method requires prior human knowledge to adapt the algorithm to new images, leading to limited robustness.

% \textcolor{red}{Previous about the rhythm works?}

\subsection{Co-speech Generation}
\label{ssec:cosp_g}
The milestones of CoS generation in recent years are
presented in Figure \ref{fig:timeline-cospeech}. The upper part is related datasets and the lower part is the algorithm. 
The target of CoS gesture generation is to generate a sequence of body movements based on the corresponding audio input. It has been widely used in virtual character animation, especially in virtual speech and advertising. We divided it into three stages based on performance and popularity, Which are rule/statistical-based methods, DL-based methods, and Diffusion-based methods. In Table \ref{tab:generationwork-CoS}, we present a summary of the details of the related SL generation works. 

\textbf{Multi-modal Feature Extraction.} In the CoS gesture generation task, the data of different modalities such as text and speech contain semantic and rhythmic information. How to extract and fuse these features to get a better representation is an important topic. \cite{yoon2020speech} uses a tri-modal encoder to encode text, speech, and person IDs separately, and then perform feature fusion, sampling from the fused feature space to complete the generation task. \cite{kim2023flame} separately models speech and text information. Instead of directly fusing at the feature level, it establishes two pipelines to model the dynamic and semantic information of the gesture motion, so as to generate accurate and rhythmic gesture sequences.

\textbf{Generative Model.} Numerous endeavors have been made in the process of choosing the generative model for CoS gesture generation task. In the early research, rule-based approaches \cite{cassell2001beat,cassell1994animated,wagner2014gesture} were used, which required the manual construction of a gesture library and the development of rules mapping from spoken language to gestures in the library. These methods had limited flexibility and required expert knowledge, but it is easier to be interpreted and were effective at handling semantic gestures. Then, statistical-based methods\cite{levine2010gesture} replaced the manually written rules with traditional statistical models (\eg HMMs) trained on a dataset but still required the high-cost manual construction of a gesture library. In recent years, DL-based end-to-end approaches \cite{ao2022rhythmic, qian2021speech} have been developed, which use raw ``speech-gesture" datasets such as Trinity and TED \cite{ferstl2018investigating,yoon2019robots} to train deep neural networks for end-to-end gesture generation. These methods have reduced system complexity and produced more natural and fluid gestures, but they cannot guarantee the accuracy of generated rhythmic and semantic gestures. Meanwhile, most CoS research works do not consider the generation of the whole body, which also limits its expressiveness. Recently, diffusion models \cite{Dhariwal2021diffusion} have emerged as powerful deep generative models. Zhu et al. \cite{zhu2023taming} introduced a novel diffusion-based framework called DiffGesture, which effectively captures the associations between audio and gestures and maintains temporal coherence to generate high-quality CoS gestures. However, the diffusion-based method has limitations in terms of training cost and the need for multiple steps to achieve satisfactory results, which hinders its real-time application in CoS gesture generation.

\subsection{Talking Head Generation}
\label{ssec:thg}

\begin{figure}[t]
    \centering
    \includegraphics[width=\linewidth]{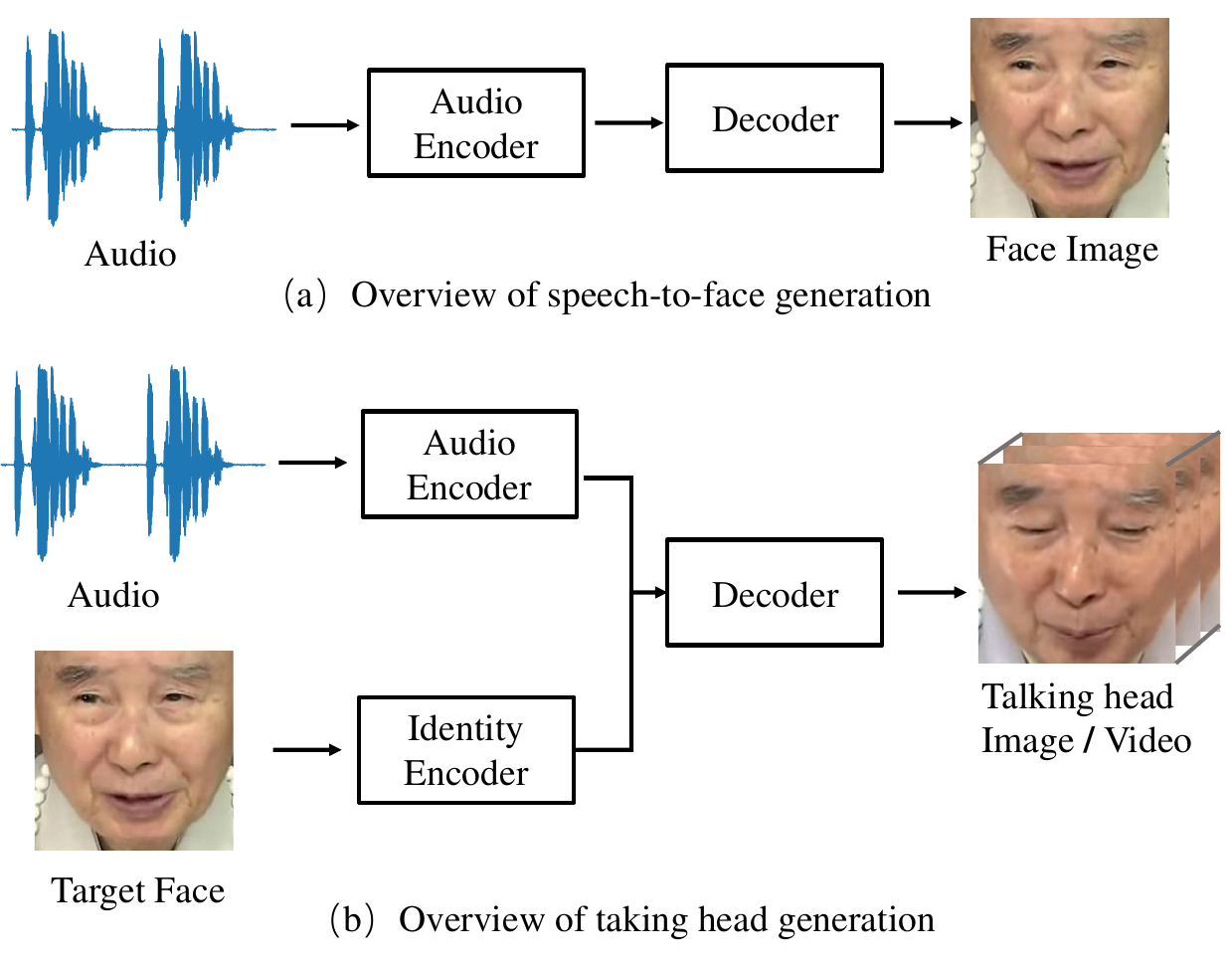}
    \caption{The two basic problems of speech-to-face generation.}
    \label{fig:overview}
\end{figure}

TH generation has become an emerging research topic in recent years. As shown in Figure \ref{fig:overview}, talking face generation from an audio clip or dynamic TH generation from a target image and an audio clip are two fundamental research problems. The problems' solutions are essential to enabling a wide range of practical applications: (a) Entertainment: Generating virtual characters with realistic expressions and voice output can be applied to virtual reality games, special effects in movies, and other fields, to enhance user experience; (b) Virtual assistants: Generating virtual assistants with natural language voice and facial expressions can be used in customer service, robot assistants, and other scenarios to improve natural language interaction experience; (c) Human-machine interaction: Generating virtual characters with realistic expressions and voice output can be used for virtual meetings, remote education, and other scenarios to improve human-machine interaction effectiveness.
(d) Healthcare: Generating virtual doctors with natural speak voices and facial expressions can be used in telemedicine, psychotherapy, and other scenarios to improve service quality and user experience.

\subsubsection{Speech-to-face Generation}
There is a strong connection between speech and face attributes, such as age, gender, and the shape of the mouth, which directly affect the mechanics of speech generation \cite{teager1990evidence}. Additionally, properties of speech such as language, accent, speed, and pronunciation are frequently shared among various nationalities and cultures. These properties can consequently manifest as standard physical facial features.

\textbf{Feature Extraction.}
For speech and face feature extraction, Oh et al. \cite{oh2019speech2face} employs a trained face recognition network \cite{parkhi2015deep} to obtain face embedding, and a voice encoder that takes a complex spectrogram of speech as input and output speech features. Duarte et al. \cite{duarte2019wav2pix} design a speech encoder modified from SEGAN discriminator \cite{pascual2017segan} to learn audio embedding. Similarly, Wen et al. \cite{wen2019face} develops a voice embedding network consisting of six convolution layers to learn speech features. A voice encoder included voice activity detection and V-net is used in Fang et al. \cite{fang2022facial} to output audio embedding. 

\textbf{Generation Model.} Oh et al. \cite{oh2019speech2face} employ a pre-trained face decoder \cite{cole2017synthesizing} to reconstruct the face image.
Motivated by the success of GAN \cite{goodfellow2020generative} in generation images with high quality. Duarte et al. \cite{duarte2019wav2pix} developed a conditional GAN called WavPix that is able to generate face images directly from the speech. For better identity matching, Wen et al. \cite{wen2019face} introduced the second discriminator to verify the identity of face image output. Considering that emotional expression is a key face attribute of a realistic face image, Fang et al. \cite{fang2022facial} applied two classifiers to measure identity and emotion semantic relevance in generating. In \cite{wang2022residual}, a Face-based Residual Personalized Speech Synthesis Model (FR-PSS) containing a speech encoder, a speech synthesizer and a face encoder is designed for PSS.

These aforementioned methods can generate face images from speech, however, the authenticity and accuracy of the reconstructed face image still need to be improved: (a) Explicit cross-modal correlation learning is vital for identity information preservation, which is not explored in the previous methods. (b) The face images synthesized by the GAN-based or CNN-based generator lack details and authenticity.

\textbf{Evaluations Metrics.} For speech-to-face generation methods, identity information preservation is the key factor, therefore, quantitative metrics related to identity consistency are used to evaluate the performance, which includes landmark distance, feature distance, and face attributes evaluation. Landmark distance is to calculate the distance of landmark (LMD) of generated face image and true face, where the landmark is achieved by Dlib \cite{king2009dlib} pre-trained DL methods such as FaceNet \cite{schroff2015facenet}. 
Feature distances are Cosine, $L_{2}$, and $L_{1}$  distances calculated between the feature of the true face and generated face. The face attributes are generally evaluated by attribution recognition accuracy like gender recognition, identity recognition, and face retrieval.
The quality of generated face images is also important for speech-to-face generation, Fréchet Inception Distance (FID), and Inception score (IS) are two common metrics to evaluate performance.
Those abovementioned metrics are highlighted in Table \ref{tab:Speech-to-face quantitative metrics}.
\begin{table}[ht]
\centering
    \caption{Summary of quantitative metrics of Speech-to-face generation}
    \renewcommand{\arraystretch}{1.3}
    \begin{tabular}{c|c}
    \hline
    \textbf{Metrics' degree} & \textbf{Metrics} 
    \\ \hline
    Identity preservation    & \renewcommand{\arraystretch}{1}
                                 \begin{tabular}[c]{@{}c@{}}LDM \cite{oh2019speech2face},
                                  \\ Cosine, $L_{2}$, $L_{1}$  \cite{oh2019speech2face},
                                  \\ Face retrieval  \cite{oh2019speech2face},
                                  \\ Identity recognition \cite{wen2019face },
                                  \\ Gender classification \cite{wen2019face}\end{tabular} 
    \\ \hline
    Image quality            & IS \cite{fang2022facial}, FID \cite{fang2022facial}                                                             
    \\ \hline
    \end{tabular}
    \label{tab:Speech-to-face quantitative metrics}
\end{table}
\begin{table*}
    \centering
    \caption{Summary of recent studies related to Talking Head generation. The following aspects are concluded: the network architecture for image synthesis and driving source; the methods work for a specific target or arbitrary identity; the audio feature is synchronized with lip motions or not; the ability to generate personalized attributes, and if any intermediate face models are used.}
    \renewcommand{\arraystretch}{1.2}
    \begin{tabular}{c|c|c|c|c|c|c|c}
         \hline
        \textbf{Framework}   & \textbf{Methods} & \textbf{Year} & \textbf{Driving source} & \textbf{Target} & \textbf{Audio features} & \textbf{Personalized} & \textbf{Face model} \\
         \hline
          \multirow{30}{*}{GAN}
         &Chen et al. \cite{chen2018lip}& 2018& \thead{Audio}& Arbitrary& Sync& No& No\\ \cline{2-8}
      
         &Song et al. \cite{song2019talking}& 2019&  Audio& Arbitrary& Sync& No& No\\ \cline{2-8}

         &Zhou et al. \cite{zhou2019talking}& 2019&  Audio& Arbitrary& Sync& No& No\\ \cline{2-8}
         
         &ATVG \cite{chen2019hierarchical}& 2019& Audio& Arbitrary& not sync& No& 2D landmarks\\ \cline{2-8}

          &Vougioukas et al. \cite{vougioukas2019end}& 2019& Audio& Arbitrary& Sync&\renewcommand{\arraystretch}{1}\begin{tabular}[c]{@{}c@{}}Eye blinks,\\eyebrow\end{tabular}& No\\ \cline{2-8}
       
          &Kefalas et al . \cite{kefalas2020speech}& 2020& Audio& Arbitrary& No sync& No & No\\ \cline{2-8}
        
          &Sinha et al. \cite{sinha2020identity}& 2020&  Audio& Arbitrary& No Sync& Eye blinking& No\\ \cline{2-8}
         
          &Wang et al. \cite{wang2020speech}& 2020&  Audio& Arbitrary& Sync& Head pose& 2D landmark\\ \cline{2-8}
         
          &Wav2lip \cite{prajwal2020lip}& 2020&  Audio&Arbitrary& Sync& No& No\\  \cline{2-8}
        
          &Eskimez et al. \cite{eskimez2020end}& 2020& Audio& Arbitrary& Sync& No& No\\ \cline{2-8}
        
          &Yi et al. \cite{yi2020audio}& 2020&  Video&Specific& Not sync& Head pose& 3DMM\\ \cline{2-8}
       
          &Chen et al. \cite{chen2020talking}& 2020& Video& Arbitrary& Not sync& Head pose& 3DMM\\ \cline{2-8}

          & Mittal et al. \cite{mittal2020animating}& 2021& Audio&Arbitrary& Not sync& No& No\\ \cline{2-8}
     
          &MEAD \cite{wang2020mead}& 2020& Audio& Arbitrary& Not sync& Emotion& No\\ \cline{2-8}
         &Zhu et al. \cite{zhu2021arbitrary}& 2021& Audio& Arbitrary& No sync& No& No\\ \cline{2-8}
            
          &FACIAL \cite{zhang2021facial}& 2021&  Video& Arbitrary& Not sync&\renewcommand{\arraystretch}{1}\begin{tabular}[c]{@{}c@{}}Head pose,\\eye blinking\end{tabular}& 3DMM\\  \cline{2-8}

          &Zhang et al. \cite{zhang2021flow}& 2021& Audio&Arbitrary&Sync&\renewcommand{\arraystretch}{1}\begin{tabular}[c]{@{}c@{}}Head pose,\\eyebrow\end{tabular}& 3DMM\\  \cline{2-8}
         
          &Si et al. \cite{si2021speech2video}& 2021&  Audio& Arbitrary& No sync& Emotion& No\\ \cline{2-8}
         
         & Chen et al. \cite{chen2021talking}& 2021& Audio&Arbitrary&  Sync& No& No\\ \cline{2-8}
        
          &PC-AVS \cite{zhou2021pose}& 2021&  Video& Arbitrary& Sync& Head pose& No\\ \cline{2-8}
         
         &GC-VAT \cite{liang2022expressive}& 2022&Video& Arbitrary& Sync&\renewcommand{\arraystretch}{1}\begin{tabular}[c]{@{}c@{}}Head pose,\\expression\end{tabular}& No\\ \cline{2-8}
       
          &Wang et al. \cite{wang2022one}& 2022& Audio& Arbitrary& Sync& Head pose& No\\ \cline{2-8}
       
          &EAMM \cite{ji2022eamm}& 2022& Video& Arbitrary& No sync& Emotion& No\\ \cline{2-8}
         
          &SPACE \cite{gururani2022spacex}& 2022& Audio& Arbitrary& No sync& \renewcommand{\arraystretch}{1}\begin{tabular}[c]{@{}c@{}}Head pose,\\emotion\end{tabular}& 2D landmark\\ \cline{2-8}
    
           &DIRFA \cite{wu2023audio}& 2023&  Audio& Arbitrary& Sync& No& No\\ \cline{2-8}
         
           &DisCoHead \cite{hwang2023discohead}& 2023& Video& Arbitrary& Sync&\renewcommand{\arraystretch}{1}\begin{tabular}[c]{@{}c@{}}Head pose,\\eye blinking,\\eyebrow\end{tabular} &No\\ \cline{2-8}
           
           &OPT \cite{liu2023opt}& 2023&  Audio& Arbitrary& No sync& \renewcommand{\arraystretch}{1}\begin{tabular}[c]{@{}c@{}}Head pose,\\expression\end{tabular}& 3DMM\\ \cline{2-8}
          
           &Wang et al. \cite{wang2023progressive}& 2023&  Audio& Abitrary&  Sync& \renewcommand{\arraystretch}{1} \begin{tabular}[c]{@{}c@{}}Head pose,\\expression,gaze,\\eye blinking\end{tabular}& No\\ \cline{2-8}
          
           &Zhang et al. \cite{zhang2023talking}& 2023&  Audio& Abitrary&  No sync& No& No\\ \cline{2-8}
          \hline
    \end{tabular}
    \label{tab: Summary of recent studies based on GAN }
\end{table*}

\begin{table*}
    \centering
    \caption{Summary of recent studies related to Talking Head generation. The following aspects are concluded: The network architecture for image synthesis; Driving source; The methods work for a specific target or arbitrary identity; The audio feature is synchronized with lip motions or not; The ability to generate personalized attributes, and if any intermediate face models are used.}
 \renewcommand{\arraystretch}{1.2}
    \begin{tabular}{c|c|c|c|c|c|c|c}
         \hline
         \textbf{Framework}   & \textbf{Methods} & \textbf{Year} & \textbf{Driving source} & \textbf{Target} & \textbf{Audio features} & \textbf{Personalized} & \textbf{Face model} \\
         \hline
           \multirow{6}{*}{CNN}
                &X2Face \cite{wiles2018x2face}& 2018&Audio, video& Arbitrary&Sync&\renewcommand{\arraystretch}{1}\begin{tabular}[c]{@{}c@{}}Head pose,\\expression\end{tabular}& No\\ \cline{2-8}
                &Jamaludin et al. \cite{jamaludin2019you} & 2019& Audio& Arbitrary& Sync& No& No\\ \cline{2-8}
                &Wen et al. \cite{wen2020photorealistic}& 2020& Video, audio& Arbitrary& No sync&\renewcommand{\arraystretch}{1}\begin{tabular}[c]{@{}c@{}}Head pose,\\expression\end{tabular}& 3DMM\\ \cline{2-8}
                &LipSync3D \cite{lahiri2021lipsync3d}& 2021&  Video& Specific& No sync& No& 3DMM\\ \cline{2-8}
                &Audio2head \cite{wang2021audio2head}& 2021& Audio& Arbitrary& Sync& Head pose& 2D landmark\\ \cline{2-8}
                &Lu et al. \cite{lu2021live}& 2021&  Audio& Specific& No sync&\renewcommand{\arraystretch}{1}\begin{tabular}[c]{@{}c@{}}Head pose,\\eyebrow\end{tabular}\\ 
           \hline
            RNN
                 &Bigioi et al. \cite{bigioi2022pose}& 2022& Video, audio& Arbitrary& No sync& Head pose& 2D landmark.\\
            \hline
            VAE
                 &SadTalker \cite{zhang2023sadtalker}& 2023&  Audio& Abitrary& No sync&\renewcommand{\arraystretch}{1}\begin{tabular}[c]{@{}c@{}}Head pose,\\eye blinking\end{tabular}& 3DMM\\
            \hline
            \multirow{4}{*}{NeRF}
                 &AD-NeRF \cite{guo2021ad}& 2021& Audio& Specific& No sync& No& No\\ \cline{2-8}
                 &DFA-NERF \cite{yao2022dfa}& 2022&  Video&specific&  Sync&\renewcommand{\arraystretch}{1}\begin{tabular}[c]{@{}c@{}}Eye blinking,\\head pose\end{tabular}& No\\ \cline{2-8}
                 &DFRF \cite{shen2022learning}& 2022&  Audio& Arbitrary& No sync& No& 3DMM\\  \cline{2-8}
                 &SSP-NeRF \cite{liu2022semantic}& 2022&  Video& Arbitrary& No sync& No& 3DMM\\ 
            \hline
            \multirow{4}{*}{DM} 
                &Yu et al. \cite{yu2022talking}& 2022& Audio&Arbitrary&  Sync&  Facial motion& No\\ \cline{2-8}
                &Zhua et al. \cite{zhua2023audio}& 2023&  Video& Arbitrary& Sync&\renewcommand{\arraystretch}{1}\begin{tabular}[c]{@{}c@{}}Eye blinking,\\head pose\end{tabular}& 3DMM\\ \cline{2-8}
                &DiffTalk \cite{shen2023difftalk}& 2023& Audio & Arbitrary& Sync& No& No\\ \cline{2-8}
                &Xu et al. \cite{xu2023multimodal}& 2023& Audio, text& Arbitrary& No sync& Emotion& 3DMM\\
            \hline
        \end{tabular}
        \label{tab: summary of other methods}
\end{table*}

\subsubsection{Talking Head Generation}
Given a target face image and a speech clip, TH generation aims at synthesizing a sequence of target face images where the lip motion, head pose, and facial expressions are synchronized with the audio. Significantly different from the speech-to-face generation task, which extracts the identity of the speaker from the given speech, the TH generation task focuses on the content of the speech.

\textbf{Multi-modal Feature Extraction.}
A VGG-M network pre-trained on the VGG Face dataset \cite{chatfield2014return} is employed in \cite{jamaludin2019you} to learn face features and a speech encoder modified from VGG-M is used to learn speech embedding. 
Three temporal encoders are used to extract representations of the speaker's identity, the audio segment, and the facial expressions, and a polynomial fusion layer is designed to generate a joint representation of the three encodings \cite{kefalas2020speech}. 
Differently, Mittal et al. \cite{mittal2020animating} develop a VAE to disentangle the phonetic content, emotional tone, and other factors into different representations solely from the input audio signal. To effectively disentangle each motion factor and achieve fine-grained controllable TH generation, Wang et al. \cite{wang2023progressive} propose a progressive disentangled representation strategy by separating the factors in a coarse-to-fine manner, where we first extract
unified motion feature from the driving signal, and then isolate each fine-grained motion from the unified feature.
A pre-trained audio-to-AU module is employed in \cite{chen2021talking} to extract the speech-related AU information from speech.
%For talking head generation methods based on 2D landmark \cite{chen2019hierarchical,wang2020speech}  , additional networks 
%like CNN and GRU \cite{wang2020speech},  LSTM-based audio transformation network \cite{chen2019hierarchical} are employed to transfer audio signal to landmark representations.

\textbf{Multi-modal Learning.} For TH video generation, speech-synchronized lip movement, facial expressions, and head pose generation are key factors. Therefore, in the training stage, audio-visual cross-modal correlation learning is necessary for the consistency of these facial movements in a sequence. Chen et al. \cite{chen2018lip} propose an audio-visual correlation loss to synchronize lip changes and speech changes in a video regarding that variation along the temporal axis between two modalities are more likely correlated, specifically, the cosine similarity loss is used to maximize the correlation between the derivative of audio feature and visual variations. For joint audio-visual representation learning, Zhou et al. \cite{zhou2019talking} enforces the audio features and visual features to share a classifier so that they can share the same distribution, additionally, a contrastive loss is employed to close the paired audio and visual features. Eskimez et al. \cite{eskimez2020end} designs a pair discriminator to improve the synchronization between the mouth shape and the input speech in the generated video. Zhu et al. \cite{zhu2021arbitrary} introduces the theory of mutual information neural estimation in talking face generation task to learn the cross-modal coherence.

\textbf{Generation Model.} The development of DL-based methods including CNN, RNN, GAN, Variational Autoencoder (VAE), Neural Radiance Fields (NeRF), and diffusion model (DM) have been explored in recent years.  We compare the difference among them in Table \ref{tab: Summary of recent studies based on GAN }
and Table \ref{tab: summary of other methods}.

The GAN-based methods are the mainstream for TH generation, in particular, because of their ability to synthesize data before the stronger generator DM emerged. In Table \ref{tab: Summary of recent studies based on GAN }, we briefly list the recent works related to TH generation based on the GAN framework. Chen et al. \cite{chen2018lip} proposes a three-stream GAN to generate speech-synchronized lip video. Wang et al. \cite{wang2020speech} uses the GAN base network with an attentional mechanism to identify features related to head information. 
Zhang et al. \cite{zhang2021facial} designs a FACIAL-GAN to encoder explicit and implicit attribute information for talking face video generation with audio-synchronized lip motion, personalized and natural head motion, and realistic eye blinks. 

In addition to GAN-based approaches, inspired by the NeRF \cite{martin2021nerf}, Guo et al. \cite{guo2021ad} develops the audio-driven NeRF (AD-NeRF) model for TH synthesis, in which an implicit neural scene representation function is learned to map audio features to dynamic neural radiation fields for speaker face rendering. However, AD-NeRF often suffers from head and torso separation during the rendering stage. Therefore, a semantic-aware speaker portrait NeRF (SSP-NeRF) is proposed by Liu et al. \cite{liu2022semantic}. They employ the semantic awareness of speech to address the problem of incongruity between local dynamics and global torso. The problem of slow rendering speed can not be ignored. To improve the real-time performance, Yao et al. \cite{yao2022dfa} proposes a NeRF method that takes lip movement features and personalized attributes as two disentangled conditions, where lip movements are directly predicted from the audio inputs to achieve lip-synchronized generation.
 
Diffusion Probabilistic Models (DM) have shown strong ability in various generation tasks \cite{huang2022prodiff, saharia2022palette}. Zhua et al. \cite{zhua2023audio} proposes an audio-driven diffusion model for TH video generation, in which the lip motion features are aligned with the TH by contrastive learning. Yu et al. \cite{yu2022talking} proposes audio-to-visual diffusion prior trained on top of the mapping between audio and disentangled non-lip facial representations to semantically match the input audio while still maintaining both the photo-realism of audio-lip synchronization and the overall naturalness.
Shen et al. \cite{shen2023difftalk} employs the emerging powerful diffusion models and model the TH generation as an audio-driven temporally coherent denoising process (DiffTalk). 
Xu et al. \cite{xu2023multimodal} first represents the emotion in the text prompt, which could inherit rich semantics from the CLIP, allowing flexible and generalized emotion control.

\begin{figure}[ht]
    \centering
    \includegraphics[width=\linewidth]{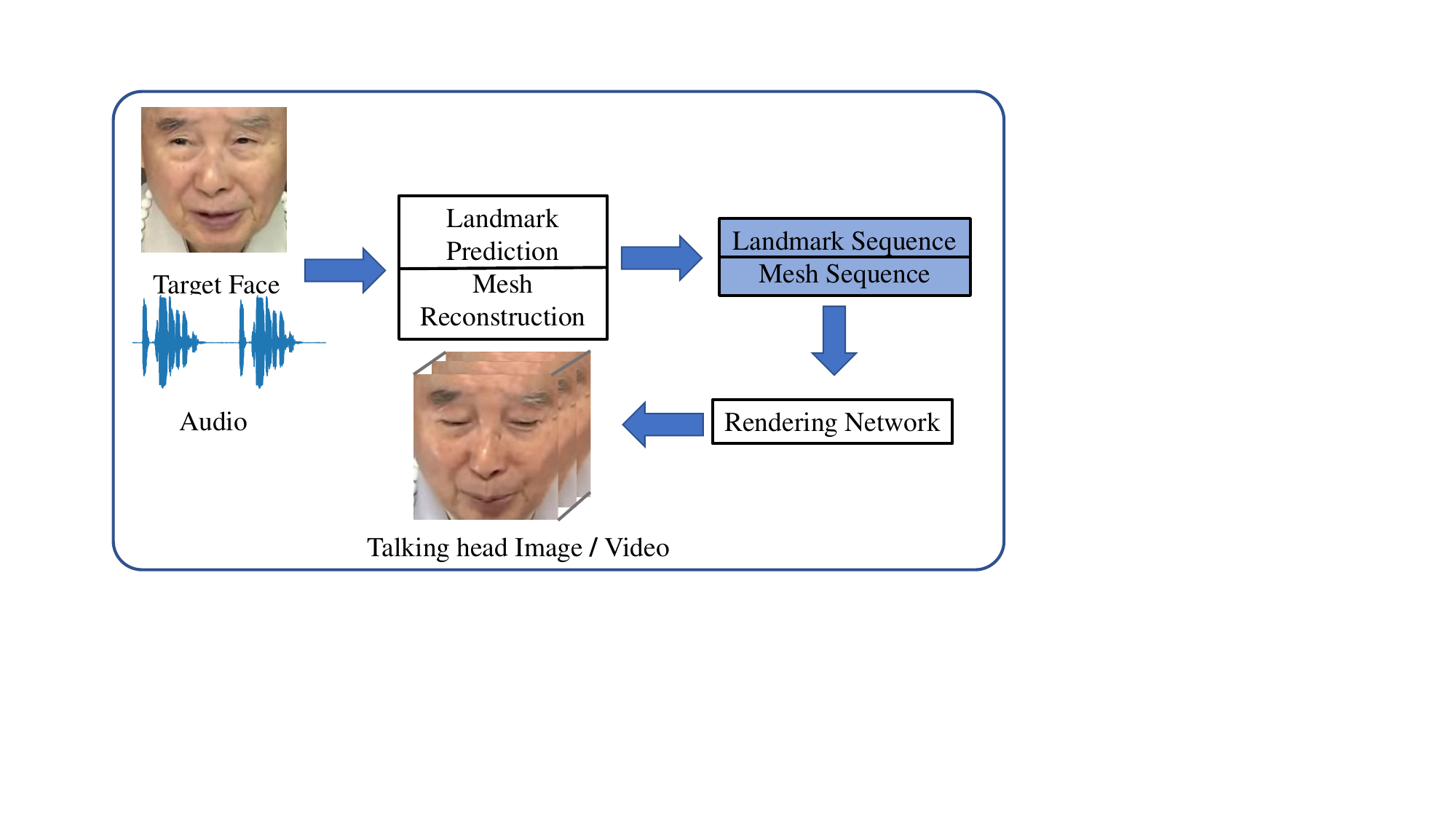}
    \caption{The typical pipeline of TH generation methods based on the intermediate face.}
    \label{fig:pipeline}
\end{figure}

For better facial appearance transfer, intermediate faces such as 2D landmarks or 3DMM are widely used in TH generation. Figure. \ref{fig:pipeline} illustrates a simplified pipeline of the TH generation methods based on intermediate face, which mainly consists of two steps: low-dimensional driving source data are mapped into facial parameters; then rendering network is used to convert the learned facial parameters into high-dimensional video output. 

\textbf{Evaluation Metrics.} 
Various perspectives reveal that the generated text-to-speech (TTS) output lacks the authenticity of human speech: (a) The target individual's face should match that of the synthetic video's speaker, (b) The generated speaker's mouth should synchronize the audio, (c) The produced TH video should be of a good caliber, (d) The expression of the speaker in the generated video should be natural and match the emotion of the audio, and (e) Eye blinking should be expected when talking. Thus, the quantitative metrics of TH generation can be classified from these five views, as shown in Table \ref{tab: talking head generation quantitative metrics}.

\begin{table}[h]
    \centering
    \caption{Summary of quantitative metrics of Talking Head Generation}
    \renewcommand{\arraystretch}{1.3}
    \begin{tabular}{c|c}
    \hline
        \textbf{Metrics' degree} &\textbf{ Metrics}\\
        \hline
        Identity-preserving&  
                            \renewcommand{\arraystretch}{1}
                            \begin{tabular}[c]{@{}c@{}}
                                PSNR, SSIM \cite{wang2004image}, FID, LMD,\\
                                LPIPS \cite{zhang2018unreasonable}, CSIM, IS, ACD \cite{tulyakov2018mocogan}
                            \end{tabular} \\
        \hline
        \begin{tabular}[c]{@{}c@{}}Audio-visual \\ synchronization\end{tabular}&    
                            \renewcommand{\arraystretch}{1}
                            \begin{tabular}[c]{@{}c@{}}
                                AV Conf, AV Off \cite{chung2017out},\\
                                WER \cite{vougioukas2019end}, LMD$_{m}$ \cite{liang2022expressive}, \cite{tulyakov2018mocogan},\\
                                Sync$_{conf}$ \cite{liang2022expressive},\\
                                LRSD, LRA \cite{chung2017lip}
                            \end{tabular} \\
        \hline
        Image quality preserving& CPBD \cite{narvekar2009no}, FDBM \cite{de2013image}\\
        \hline
        Expression& Classification accuracy \cite{zeng2022expression}\\
        \hline
        Eye blinking&  
                            \renewcommand{\arraystretch}{1}
                            \begin{tabular}[c]{@{}c@{}}
                                EAR \cite{vougioukas2020realistic}, Blink rate,\\
                                Blink median duration \cite{zhang2021facial}
                            \end{tabular} \\
        \hline         
    \end{tabular}
    \label{tab: talking head generation quantitative metrics}
\end{table}

\textbf{Audio Input Pre-processing.}
Most of the TH generation works are audio signal driven. Here, we will introduce how previous work has dealt with speech signals in this field. In general, the audio waveform is resampled at 16KHz, and then the audio feature is computed \cite{wang2022acoustic}. Spectrogram, MFCC, and Fbank are the three mostly used audio features. Fang et al. \cite{fang2022facial} performs an ablation experiment on these three audio features, and they found that Fbank achieved the best performance, while the Spectrogram performed the worst FID. The reasons they guessed that Spectrogram contained much redundant information, MFCC discarded some related information, and Fbank kept balance. However, MFCC is used the most in the talking face generation. 

\section{Challenges of BL Recognition and Generation}

% \textbf{Fengji}
The existing BL recognition and generation methods have not been capable of meeting real-world requirements under exposure to various challenges. In order to fully demonstrate the typical challenges of BL recognition and generation in the field of BL, we elaborate in detail on SL, CS, and TH from three aspects: \textbf{subtasks challenges}, \textbf{datasets challenges} and \textbf{evaluation metrics challenges}. 
\subsection{Subtasks Challenges}
To more fully illustrate the challenges of BL recognition and generation tasks, we split each major task into three subtasks, \ie Lip reading,
SL recognition, and CS recognition. From the perspective of task definition, the TH task itself is more focused on the generation process. Moreover, limited by the development of the existing TH generation, it is difficult for researchers to capture the basic facial attributes of the target speaker. The existing studies lack an exploration of TH recognition, so the challenge of TH recognition is not included in the discussion of subtask challenges in this survey. Current research on CoS predominantly concentrates on CoS gesture generation. While some studies have demonstrated a positive impact on the CoS Generation task, the majority of recent works do not prioritize CoS Recognition as a primary focus. So the challenge of CoS Recognition is not included in the discussion of subtask challenges in this survey.

The challenges of BL recognition tasks are mainly due to the efficiency of the cross-modal feature fusion, and the specific challenges of each subtask are as follows. 

% \textcolor{red}{Explain No Co-Speech}
\begin{itemize}
    \item \textbf{Lip Reading.}
    There are two primary challenges in automatic lip reading: intra-class difference and inter-class similarity. The former is hindered by factors such as speech emotion, speed, gender, age, skin color, and speech habits, making it difficult to distinguish variations within the same word category. Additionally, the semantic disparities between words used in different contexts significantly impact lip reading. The latter challenge stems from the abundance of word categories, leading to challenges in visually distinguishing similar-looking words belonging to different classes. Addressing these challenges is crucial for improving the accuracy and effectiveness of lip reading recognition systems, which have valuable implications for aiding communication for individuals with hearing impairments and advancing the field's applications.
    % There are two challenges in lip reading recognition, intra-class difference and inter-class similarity~\cite{Sheng2022vsa}. The former challenge suffers from speech emotion, speed, gender, age, skin color, and speech habits. At the same time, the semantic difference between words in different contexts will also have a serious impact on BL recognition. The latter challenge suffers from the plethora of word categories.
    
    \item \textbf{Sign Language Recognition.}
    SL recognition encounters significant challenges arising from the pronounced variations in gestures, which seriously impede its accuracy. Moreover, factors like hand shape, illumination conditions, and resolution play pivotal roles in limiting SL recognition performance. Additionally, occlusion, including self-occlusion between fingers and occlusion between hands and other body parts, adversely affects feature fusion, becoming a key influencing factor in SL recognition. Another pressing challenge is the development of a real-time multilingual SL recognition system. Addressing these complexities is essential to advance the field and improve the efficiency and inclusivity of SL recognition technologies.
    % The high variations of gestures seriously interfere with SL recognition~\cite{Rastgoo2021recognition}. Meanwhile, hand shape, illumination condition and resolution are all important factors limiting SL recognition. In addition to the above factors, occlusion (including self-occlusion between fingers, occlusion between hands and other parts of the body, etc.) defects feature fusion which is a main affecting factor of SL recognition. Another challenge in SL recognition is to design a multilingual real-time SL recognition system.
    \item \textbf{Cued Speech Recognition.}
    The primary obstacle in CS recognition is the hand preceding phenomenon~\cite{liu2023cross}, where the hand movements often occur faster than the corresponding lip movements, anticipating the next phoneme. This phenomenon hampers the efficiency of lip and hand feature fusion in CS recognition. Besides,
due to variations in individual CS coding habits and styles, adaptability in multi-cuer scenarios is also a challenge.
    % The main challenge of CS recognition is the hand preceding phenomenon~\cite{liu2023cross} which means the hand usually moves faster than the lips to prepare for the next phoneme.  hand preceding phenomenon defects the efficiency of the feature fusion of CS recognition. 
\end{itemize}

The challenge of BL generation mainly stems from the stability and quality of the generated gesture, and the specific challenges of each subtask are as follows.
\begin{itemize}
    \item \textbf{Cued Speech Generation.}
In conclusion, the CS generation faces several challenges that need to be addressed for the development of effective systems. The lack of large-scale annotated datasets, the complexity of modeling CS gestures, and the need for accurate asynchronous alignment between cued signs and spoken words are key challenges. Additionally, integrating audio and visual modalities and achieving generalization to new speakers and languages are important considerations. Overcoming these challenges through advancements in modeling ability, multi-modal fusion, and the availability of diverse datasets will contribute to the improvement of CS generation systems.
    \item \textbf{Sign Language Generation.}
    In the realm of SL production, numerous obstacles warrant attention, chief among them being domain adaptation and model collapse. The former obstacle arises from the inherent variations in word styles and meanings across different languages, necessitating effective adaptation strategies. Furthermore, a noteworthy challenge lies in the limited proficiency of generating uncommon and unseen words, hindering the overall performance of the system. Moreover, the persisting issues of model collapse, non-convergence, and instability within generative models further compound the complexities faced in Sign Language production. Addressing these multifaceted challenges is crucial for advancing the SOTA in this domain and facilitating more reliable and robust SL generation.
    % There are several challenges such as domain adaptation and model collapse in SL prodction~\cite{rastgoo2021sign}. One of the challenges in SL production is related to domain adaptation due to different word styles and meaning in different languages. Another challenge is the poor generation performance of uncommon and unseen words. At the same time, the model collapse, non-convergence, and instability of generative models are also challenges in SL production.
    
    \item \textbf{Co-speech Generation.}
    The generation process of CoS encounters challenges due to the presence of highly idiosyncratic and non-periodic spontaneous gestures.  The accurate capture of finger motion poses difficulties, resulting in the manifestation of idiosyncratic gestures.  Furthermore, the non-periodic nature of gestures arises from the substantial variation in gesture behavior.
    
    % The generation processing of SL suffers from the highly idiosyncratic and non-periodic spontaneous gestures~\cite{nyatsanga2023comprehensive}.   Finger motion is difficult to be captured accurately which leads to highly idiosyncratic gestures. The non-periodic gesture is because of the high variation in gesture behavior.
    \item \textbf{Talking Head Generation.}
    TH generation confronts two primary challenges: information coupling and diversity targets. The former encompasses the synchronization of multiple facial elements, such as head posture, facial expression, lip movement, and background motion, while also addressing the ``uncanny valley effect"~\cite{mori2012valley}, a phenomenon common in face generation where generated faces appear almost human-like but lack true realism, leading to discomfort.  The latter challenge pertains to harmonizing temporal resolution and speech features across diverse data modalities, along with the complexity of defining visual quality as a clear training objective.  Overcoming these challenges is crucial for advancing the field and achieving a more realistic and visually coherent TH generation.
    % The challenges in talking head generation can be divided into two categories, information coupling and diversity targets~\cite{Sheng2022vsa}. The former challenge involves coupling information such as head posture, facial expression, lip movement, and background movement, as well as the "uncanny valley effect"~\cite{uncanny} that often occurs when it comes to face generation. The latter challenge mainly includes differences between the temporal resolution and speech features of different data modalities, as well as unclear training objectives due to the difficulty in defining visual quality.
\end{itemize}

\subsection{Datasets Challenges}

The current datasets for SL recognition and generation encounter significant limitations due to the high costs associated with data collection and manual annotation. This results in datasets with small-scale and weak annotations, hindering the progress of BL-related tasks. To create BL datasets, collaboration between language experts and native speakers is essential, further adding to the complexities and expenses involved. A potential solution to address these challenges is to explore self-supervised learning using unlabeled BL data~\cite{sheng2021cross}, which could alleviate the need for extensive manual annotation.

Moreover, privacy protection poses another hurdle, as some large BL datasets~\cite{Mroueh2015deep,shillingford2019large} are not publicly accessible. In light of the high costs and privacy concerns, a viable approach is to leverage existing wild online videos to collect the necessary BL data. Similar to the training datasets used for Contrastive Language-Image Pre-training (CLIP)~\cite{radford2021learning} and DALL-E~\cite{Ramesh2021DALL-E}, employing very large datasets can enhance the generalization capabilities of BL recognition and generation models.

Apart from the dataset challenges, the high costs associated with collecting and annotating 3D data contribute to the scarcity of large-scale 3D BL datasets. Consequently, the development of 3D BL Generation faces significant obstacles in understanding and processing 3D BL data effectively. Overcoming these challenges is essential to advance the field of BL recognition and generation, allowing for more efficient and accurate communication support for individuals with hearing impairments.
% Due to the high cost of data collection and manual annotation, the existing BL recognition and generation datasets are both under the exposure of small-scale and weak annotations.  For example, the creation of datasets related to SL tasks requires collaboration between language experts and native speakers.  A potential solution to this challenge is self-supervised learning with unlabeled BL data~\cite{sheng2021cross}. Another challenge in datasets is privacy protection, some large BL datasets~\cite{privacy1,shillingford2019large} are not publicly available. Due to the above two considerations (high cost and privacy protection), a feasible solution is to use the existing wild online video to capture the required BL data. Similar to the training datasets of  CLIP~\cite{radford2021learning} and DALL-E~\cite{DALL-E}, very large datasets can make BL recognition and generation models have strong generalization. Besides these dataset challenges, the high collection and annotating cost of 3D data lead to the lack of large-scale 3D BL, and 3D BL understanding faces great challenges, which limits the development of 3D BL Generation.
\subsection{Evaluation Metrics Challenges}
The primary nature of the BL recognition task lies in its classification essence, where simple and efficient classification accuracy serves as the prevalent evaluation metric.  However, this paper shifts its focus to the BL generation task and the challenges it poses in terms of evaluation metrics.  Subjective metrics utilized in the BL generation task prove to be costly, time-consuming and lack scalability.  Metrics like human likeness and gesture appropriateness, although valuable, suffer from non-replicability and instability issues.  On the other hand, objective metrics such as PSNR, SSIM, FID~\cite{Heusel2017FID} and LRSD~\cite{Chen2020lrsd} offer advantages over subjective ones but come with limitations in assessing the similarity between gesture and speech, as well as the semantic appropriateness of gestures.  Notably, unlike subjective metrics that evaluate human likeness, the existing literature rarely quantifies objective metrics measuring gesture diversity or various motion appropriateness aspects.  These challenges highlight the need for robust and comprehensive evaluation metrics in the BL generation domain to ensure an accurate and meaningful assessment of generated Sign Language outputs.
% BL recognition task is a classification task in essence, and simple and efficient classification accuracy is the most common evaluation metric. Therefore, this paper focuses on the challenges of evaluation metrics in the BL generation task. Subjective metrics in the BL generation task are expensive, time-consuming, and lack scalability. Subjective metrics such as human likeness and gesture appropriateness are not replicable and unstable.  Objective metrics such as PSNR, SSIM, FID~\cite{Heusel2017FID} and LRSD~\cite{LRSD} do not have the defects of the above subjective metrics, but there are limitations in evaluating the similarity between gesture and speech as well as the semantic appropriateness of gesture. Unlike subjective metrics such as human likeness, existing studies have rarely quantified objective metrics that measure gesture diversity or different kinds of motion appropriateness.

% \subsection{Audio-visual Speech Separation}

%\textcolor{red}{Body Language Datasets?}
% \section{Body Language Datasets} 
% \label{sec:Datasets}

\section{Future Discussions} 
\label{sec:Future Discussions}

Through an extensive summary and analysis of the existing literature, this survey offers the following \textbf{discussions} and \textbf{new insights}:
\begin{enumerate} 

\item The integration of large-scale multi-modal BL datasets and the establishment of a unified low-loss data format are key factors in advancing BL recognition and generation tasks. By collecting extensive datasets from diverse online videos, we can enhance the generalization and robustness of BL recognition and generation models for real-world scenarios. Additionally, the adoption of a unified data standard and adaptable conversion method allows for the seamless integration of different datasets and facilitates collaboration among researchers. This promotes interoperability between models, enabling efficient sharing and utilization of resources within the research community.
    % (2) robots and Embodied agents(Embodied AI).\\
% \item Self-supervised learning with unlabeled data. One possible future direction is multimodal self-supervised BL learning using unlabeled audiovisual data to solve the problem of small-scale and weak annotations datasets.

\item Recently, large-scale pre-training models such as ChatGPT have achieved outstanding performance in various visual-linguistic cross-modal tasks. For instance, CLIP and various variations of the multi-modal CLIP model have emerged. However, they have the following drawbacks: a) they might not deeply connect different types of data as effectively as specialized models; b) they demand in terms of computing power due to their size. c) This model might not allow fine-tuning for specific tasks and could struggle with specialized knowledge; d) it needs a lot of diverse data to work well and could be hard to interpret. To this end, how to build a large-scale multi-modal model for BL recognition and generation is a promising topic. 
\item Besides, it was found that the ability of existing large-scale pre-training models to learn fine-grain features still needs to be improved \cite{mu2023generating}. In BL, fine-grained feature learning is essential, For example, hand positions and lip movements in CS and CoS needed to be accurately recognized and generated to ensure clarity and avoid ambiguity. Therefore, fine-grained BL recognition and generation is a feasible direction to improve their performance.

\item The multi-modal models in the task of BL recognition and generation are very susceptible to the perturbations (attacks) of different modalities, resulting in serious performance degradation. How to pre-train a robust and secure multimodal large-scale model for BL recognition and generation is an urgent problem to be solved. 
\item An essential requirement for BL recognition and generation systems is real-time capability, especially for multilingual and multiple-speakers scenarios. Creating a real-time system is vital to cater to the needs of both the deaf and speaking communities. However, existing audio-visual datasets are predominantly monolingual, with English being the most commonly represented language. In practical applications, multilingual communication is often necessary, highlighting the need for diverse datasets. Additionally, current methods for BL recognition and generation are often limited to specific target identities, as different speakers exhibit significant variations in appearance and habits.  Overcoming these challenges is crucial to develop adaptable and effective real-time BL systems that accommodate various languages and diverse speakers.

\end{enumerate}

\section{Conclusion} 
\label{sec:Conclusion}
This survey has delved into the realm of deep multi-modal learning for automatic BL recognition and generation, shedding light on its potential and challenges. This survey focuses on four classical BL variants, \ie Sign Language, Cued Speech, Co-speech, and Talking Head. Through a meticulous examination of various modalities, including visual, auditory, and textual data, and their integration, we have explored the intricacies of capturing and interpreting these four BL. By reviewing SOTA methodologies, such as feature fusion, representation learning, recognition, and generation methods, we have uncovered the strengths and limitations of current approaches. The significance of datasets and benchmarks in facilitating research progress was also emphasized, with a focus on annotation methodologies and evaluation metrics. Despite the progress, challenges persist, demanding the creation of diverse datasets, addressing limited labeled data, enhancing model interpretability, and ensuring robustness across environments and cultural contexts. Looking ahead, the future holds promises of more sophisticated architectures and training strategies, harnessing the complementary nature of multi-modal data and leveraging advancements in multi-modal learning, large-scale pre-trained model, self-supervised learning, and reinforcement learning. As this research area evolves, it is poised to revolutionize human-human and human-machine interactions, fostering natural and effective communication across domains.

\bibliographystyle{IEEEtran}
\bibliography{references}

\begin{thebibliography}{100}
\providecommand{\url}[1]{#1}
\csname url@samestyle\endcsname
\providecommand{\newblock}{\relax}
\providecommand{\bibinfo}[2]{#2}
\providecommand{\BIBentrySTDinterwordspacing}{\spaceskip=0pt\relax}
\providecommand{\BIBentryALTinterwordstretchfactor}{4}
\providecommand{\BIBentryALTinterwordspacing}{\spaceskip=\fontdimen2\font plus
\BIBentryALTinterwordstretchfactor\fontdimen3\font minus
  \fontdimen4\font\relax}
\providecommand{\BIBforeignlanguage}[2]{{%
\expandafter\ifx\csname l@#1\endcsname\relax
\typeout{** WARNING: IEEEtran.bst: No hyphenation pattern has been}%
\typeout{** loaded for the language `#1'. Using the pattern for}%
\typeout{** the default language instead.}%
\else
\language=\csname l@#1\endcsname
\fi
#2}}
\providecommand{\BIBdecl}{\relax}
\BIBdecl

\bibitem{cornett1967cued}
R.~O. Cornett, ``Cued speech,'' \emph{American annals of the deaf}, vol. 112,
  no.~1, pp. 3--13, 1967.

\bibitem{Joksimoski2022SLreview}
B.~Joksimoski, E.~Zdravevski, P.~Lameski, I.~M. Pires, F.~J. Melero, T.~P.
  Martinez, N.~M. Garcia, M.~Mihajlov, I.~Chorbev, and V.~Trajkovik,
  ``Technological solutions for sign language recognition: A scoping review of
  research trends, challenges, and opportunities,'' \emph{IEEE Access},
  vol.~10, pp. 40\,979--40\,998, 2022.

\bibitem{liu2022audio}
X.~Liu, Q.~Wu, H.~Zhou, Y.~Du, W.~Wu, D.~Lin, and Z.~Liu, ``Audio-driven
  co-speech gesture video generation,'' \emph{Advances in Neural Information
  Processing Systems (NIPS)}, vol.~35, pp. 21\,386--21\,399, 2022.

\bibitem{zhang2023metaportrait}
B.~Zhang, C.~Qi, P.~Zhang, B.~Zhang, H.~Wu, D.~Chen, Q.~Chen, Y.~Wang, and
  F.~Wen, ``Metaportrait: Identity-preserving talking head generation with fast
  personalized adaptation,'' in \emph{Proc. IEEE/CVF-CVRP}, 2023, p.
  22096–22105.

\bibitem{rastgoo2021sign}
R.~Rastgoo, K.~Kiani, S.~Escalera, and M.~Sabokrou, ``Sign language production:
  A review,'' in \emph{Proc. IEEE/CVF-CVRP}, 2021, pp. 3451--3461.

\bibitem{nyatsanga2023comprehensive}
S.~Nyatsanga, T.~Kucherenko, C.~Ahuja, G.~E. Henter, and M.~Neff, ``A
  comprehensive review of data-driven co-speech gesture generation,'' in
  \emph{Computer Graphics Forum}, vol.~42, no.~2.\hskip 1em plus 0.5em minus
  0.4em\relax Wiley Online Library, 2023, pp. 569--596.

\bibitem{Rastgoo2021recognition}
R.~Rastgoo, K.~Kiani, and S.~Escalera, ``Sign language recognition: A deep
  survey,'' \emph{Expert Systems with Applications}, vol. 164, p. 113794, 2021.

\bibitem{fernandez2018survey}
A.~Fernandez-Lopez and F.~M. Sukno, ``Survey on automatic lip-reading in the
  era of deep learning,'' \emph{Image and Vision Computing}, vol.~78, pp.
  53--72, 2018.

\bibitem{fenghour2021deep}
S.~Fenghour, D.~Chen, K.~Guo, B.~Li, and P.~Xiao, ``Deep learning-based
  automated lip-reading: A survey,'' \emph{IEEE Access}, vol.~9, pp.
  121\,184--121\,205, 2021.

\bibitem{chand2023survey}
R.~Chand, P.~Jain, A.~Mathur, S.~Raj, and P.~Kanikar, ``Survey on visual speech
  recognition using deep learning techniques,'' in \emph{Proc. IEEE-CSCITA},
  2023, pp. 72--77.

\bibitem{bhaskar2018survey}
S.~Bhaskar, T.~Thasleema, and R.~Rajesh, ``A survey on different visual speech
  recognition techniques,'' in \emph{Data Analytics and Learning (DAL)}, 2018,
  pp. 307--316.

\bibitem{radha2021survey}
N.~Radha, A.~Shahina \emph{et~al.}, ``A survey on visual speech recognition
  approaches,'' in \emph{Proc. IEEE-ICAIS}, 2021, pp. 934--939.

\bibitem{koller2020quantitative}
O.~Koller, ``Quantitative survey of the state of the art in sign language
  recognition,'' \emph{arXiv preprint arXiv:2008.09918}, 2020.

\bibitem{adeyanju2021machine}
I.~Adeyanju, O.~Bello, and M.~Adegboye, ``Machine learning methods for sign
  language recognition: A critical review and analysis,'' \emph{Intelligent
  Systems with Applications}, vol.~12, p. 200056, 2021.

\bibitem{papastratis2021artificial}
I.~Papastratis, C.~Chatzikonstantinou, D.~Konstantinidis, K.~Dimitropoulos, and
  P.~Daras, ``Artificial intelligence technologies for sign language,''
  \emph{Sensors}, vol.~21, no.~17, p. 5843, 2021.

\bibitem{madhiarasan2022comprehensive}
D.~M. Madhiarasan, P.~Roy, and P.~Pratim, ``A comprehensive review of sign
  language recognition: Different types, modalities, and datasets,''
  \emph{arXiv preprint arXiv:2204.03328}, 2022.

\bibitem{Chen2020lrsd}
L.~Chen, G.~Cui, Z.~Kou, H.~Zheng, and C.~Xu, ``What comprises a good
  talking-head video generation?: A survey and benchmark,'' \emph{arXiv
  preprint arXiv:2005.03201}, 2020.

\bibitem{sha2023deep}
T.~Sha, W.~Zhang, T.~Shen, Z.~Li, and T.~Mei, ``Deep person generation: A
  survey from the perspective of face, pose, and cloth synthesis,'' \emph{ACM
  Computing Surveys}, vol.~55, no.~12, pp. 1--37, 2023.

\bibitem{zhen2023human}
R.~Zhen, W.~Song, Q.~He, J.~Cao, L.~Shi, and J.~Luo, ``Human-computer
  interaction system: A survey of talking-head generation,''
  \emph{Electronics}, vol.~12, no.~1, p. 218, 2023.

\bibitem{Sheng2022vsa}
C.~Sheng, G.~Kuang, L.~Bai, C.~Hou, Y.~Guo, X.~Xu, M.~Pietik{\"a}inen, and
  L.~Liu, ``Deep learning for visual speech analysis: A survey,'' \emph{arXiv
  preprint arXiv:2205.10839}, 2022.

\bibitem{Stephanie2020text2sign}
S.~Stoll, N.~C. Camgoz, S.~Hadfield, and R.~Bowden, ``Text2sign: Towards sign
  language production using neural machine translation and generative
  adversarial networks,'' \emph{International Journal of Computer Vision}, vol.
  128, pp. 891--908, 2020.

\bibitem{guo2018hierarchical}
D.~Guo, W.~Zhou, H.~Li, and M.~Wang, ``Hierarchical lstm for sign language
  translation,'' in \emph{Proc. Conf AAAI Artif. Intell.}, vol.~32, no.~1,
  2018.

\bibitem{saunders2020everybody}
B.~Saunders, N.~C. Camgoz, and R.~Bowden, ``Everybody sign now: Translating
  spoken language to photo realistic sign language video,'' \emph{arXiv
  preprint arXiv:2011.09846}, 2020.

\bibitem{Ahuja2022low}
C.~Ahuja, D.~W. Lee, and L.-P. Morency, ``Low-resource adaptation for
  personalized co-speech gesture generation,'' in \emph{Proc. IEEE/CVF-CVPR},
  June 2022, pp. 20\,566--20\,576.

\bibitem{ao2022rhythmic}
T.~Ao, Q.~Gao, Y.~Lou, B.~Chen, and L.~Liu, ``Rhythmic gesticulator:
  Rhythm-aware co-speech gesture synthesis with hierarchical neural
  embeddings,'' \emph{ACM Transactions on Graphics (TOG)}, vol.~41, no.~6, p.
  1–19, 2022.

\bibitem{Liang2022seeg}
Y.~Liang, Q.~Feng, L.~Zhu, L.~Hu, P.~Pan, and Y.~Yang, ``Seeg: Semantic
  energized co-speech gesture generation,'' in \emph{Proc. IEEE/CVF-CVPR}, June
  2022, pp. 10\,473--10\,482.

\bibitem{liu2022learning}
X.~Liu, Q.~Wu, H.~Zhou, Y.~Xu, R.~Qian, X.~Lin, X.~Zhou, W.~Wu, B.~Dai, and
  B.~Zhou, ``Learning hierarchical cross-modal association for co-speech
  gesture generation,'' in \emph{Proc. IEEE/CVF-CVPR}, 2022, pp.
  10\,462--10\,472.

\bibitem{duchnowski1998automatic}
P.~Duchnowski, L.~D. Braida, D.~Lum, M.~Sexton, J.~Krause, and S.~Banthia,
  ``Automatic generation of cued speech for the deaf: status and outlook,'' in
  \emph{International Conference on Auditory-Visual Speech Processing (AVSP)},
  1998.

\bibitem{bailly2008retargeting}
G.~Bailly, Y.~Fang, F.~Elisei, and D.~Beautemps, ``Retargeting cued speech hand
  gestures for different talking heads and speakers,'' in \emph{Retargeting
  cued speech hand gestures for different talking heads and speakers},
  September 2008, p.~8.

\bibitem{prajwal2020lip}
P.~KR, M.~Rudrabha, P.~Namboodir, and C.~Jawahar, ``A lip sync expert is all
  you need for speech to lip generation in the wild,'' in \emph{Proc. ACM MM},
  2020.

\bibitem{wang2021audio2head}
S.~Wang, L.~Li, Y.~Ding, C.~Fan, and X.~Yu, ``Audio2head: Audio-driven one-shot
  talking-head generation with natural head motion,'' in \emph{Proc. IJCAI},
  2021.

\bibitem{guo2021ad}
Y.~Guo, K.~Chen, S.~Liang, Y.-J. Liu, H.~Bao, and J.~Zhang, ``Ad-nerf: Audio
  driven neural radiance fields for talking head synthesis,'' in \emph{Proc.
  IEEE/CVF-ICCV}, 2021, pp. 5784--5794.

\bibitem{shen2023difftalk}
S.~Shen, W.~Zhao, Z.~Meng, W.~Li, Z.~Zhu, J.~Zhou, and J.~Lu, ``Difftalk:
  Crafting diffusion models for generalized audio-driven portraits animation,''
  in \emph{Proc. IEEE/CVF-CVPR}, 2023, pp. 1982--1991.

\bibitem{lucey2008patch}
P.~Lucey, G.~Potamianos, and S.~Sridharan, ``Patch-based analysis of visual
  speech from multiple views,'' in \emph{International Conference on
  Auditory-Visual Speech Processing (AVSP)}.\hskip 1em plus 0.5em minus
  0.4em\relax AVISA, 2008, pp. 69--74.

\bibitem{zhou2011towards}
Z.~Zhou, G.~Zhao, and M.~Pietik{\"a}inen, ``Towards a practical lipreading
  system,'' in \emph{Proc. IEEE/CVF-CVPR}, 2011, pp. 137--144.

\bibitem{wu2016novel}
P.~Wu, H.~Liu, X.~Li, T.~Fan, and X.~Zhang, ``A novel lip descriptor for
  audio-visual keyword spotting based on adaptive decision fusion,'' \emph{IEEE
  Transactions on Multimedia}, vol.~18, no.~3, pp. 326--338, 2016.

\bibitem{ma2021end}
P.~Ma, S.~Petridis, and M.~Pantic, ``End-to-end audio-visual speech recognition
  with conformers,'' in \emph{Proc. IEEE-ICASSP}, 2021, pp. 7613--7617.

\bibitem{liu2019novel}
L.~Liu, G.~Feng, D.~Beautemps, and X.-P. Zhang, ``A novel resynchronization
  procedure for hand-lips fusion applied to continuous french cued speech
  recognition,'' in \emph{Proc. IEEE-EUSIPCO}, 2019, pp. 1--5.

\bibitem{papadimitriou2021multimodal}
K.~Papadimitriou, M.~Parelli, G.~Sapountzaki, G.~Pavlakos, P.~Maragos, and
  G.~Potamianos, ``Multimodal fusion and sequence learning for cued speech
  recognition from videos,'' in \emph{International Conference on
  Human-Computer Interaction}, 2021, pp. 277--290.

\bibitem{liu2020re}
L.~Liu, G.~Feng, B.~Denis, and X.-P. Zhang, ``Re-synchronization using the hand
  preceding model for multi-modal fusion in automatic continuous cued speech
  recognition,'' \emph{IEEE Transactions on Multimedia}, vol.~23, pp. 292--305,
  2020.

\bibitem{liu2023cross}
L.~Liu and L.~Liu, ``Cross-modal mutual learning for cued speech recognition,''
  in \emph{Proc. IEEE-ICASSP}, 2023, pp. 1--5.

\bibitem{zhang2014threshold}
J.~Zhang, W.~Zhou, and H.~Li, ``A threshold-based {HMM}-{DTW} approach for
  continuous sign language recognition,'' in \emph{Proceedings of International
  Conference on Internet Multimedia Computing and Service}.\hskip 1em plus
  0.5em minus 0.4em\relax Association for Computing Machinery, 2014, pp.
  237--240.

\bibitem{yang2016continuous}
W.~Yang, J.~Tao, and Z.~Ye, ``Continuous sign language recognition using level
  building based on fast hidden markov model,'' \emph{Pattern Recognition
  Letters}, vol.~78, pp. 28--35, 2016.

\bibitem{cheng2020fully}
K.~L. Cheng, Z.~Yang, Q.~Chen, and Y.-W. Tai, ``Fully convolutional networks
  for continuous sign language recognition,'' in \emph{Proc. ECCV}, 2020, pp.
  697--714.

\bibitem{wei2020semantic}
C.~Wei, J.~Zhao, W.~Zhou, and H.~Li, ``Semantic boundary detection with
  reinforcement learning for continuous sign language recognition,'' \emph{IEEE
  Transactions on Circuits and Systems for Video Technology}, vol.~31, no.~3,
  pp. 1138--1149, 2020.

\bibitem{liu2019pilot}
L.~Liu and G.~Feng, ``A pilot study on mandarin chinese cued speech,''
  \emph{American Annals of the Deaf}, vol. 164, no.~4, pp. 496--518, 2019.

\bibitem{unorgsign2022}
\BIBentryALTinterwordspacing
``Sign languages unite us!'' un.org, 2022. [Online]. Available:
  \url{https://www.un.org/en/observances/sign-languages-day}
\BIBentrySTDinterwordspacing

\bibitem{pan2022speaker}
Z.~Pan, X.~Qian, and H.~Li, ``Speaker extraction with co-speech gestures cue,''
  \emph{IEEE Signal Processing Letters}, vol.~29, pp. 1467--1471, 2022.

\bibitem{sondermann2023like}
C.~Sondermann and M.~Merkt, ``Like it or learn from it: Effects of talking
  heads in educational videos,'' \emph{Computers \& Education}, vol. 193, p.
  104675, 2023.

\bibitem{song2023virtual}
W.~Song, Q.~He, and G.~Chen, ``Virtual human talking-head generation,'' in
  \emph{Proceedings of the 2023 2nd Asia Conference on Algorithms, Computing
  and Machine Learning}, 2023, pp. 1--5.

\bibitem{kothadiya2022deepsign}
D.~Kothadiya, C.~Bhatt, K.~Sapariya, K.~Patel, A.-B. Gil-Gonz{\'a}lez, and
  J.~M. Corchado, ``Deepsign: Sign language detection and recognition using
  deep learning,'' \emph{Electronics}, vol.~11, no.~11, p. 1780, 2022.

\bibitem{de2023machine}
M.~De~Coster, D.~Shterionov, M.~Van~Herreweghe, and J.~Dambre, ``Machine
  translation from signed to spoken languages: State of the art and
  challenges,'' \emph{Universal Access in the Information Society}, pp. 1--27,
  2023.

\bibitem{kahlon2023machine}
N.~K. Kahlon and W.~Singh, ``Machine translation from text to sign language: a
  systematic review,'' \emph{Universal Access in the Information Society},
  vol.~22, no.~1, pp. 1--35, 2023.

\bibitem{liu2022objective}
L.~Liu, G.~Feng, X.~Ren, and X.~Ma, ``Objective hand complexity comparison
  between two mandarin chinese cued speech systems,'' in \emph{Proc.
  IEEE-ISCSLP}, 2022, p. 215–219.

\bibitem{csorg}
\BIBentryALTinterwordspacing
``Find your cued language,'' cuedspeech.org. [Online]. Available:
  \url{https://cuedspeech.org/learn/find-your-cued-language/}
\BIBentrySTDinterwordspacing

\bibitem{liu2018modeling}
L.~Liu, ``Modeling for continuous cued speech recognition in french using
  advanced machine learning methods,'' Ph.D. dissertation, Universite Grenoble
  Alpes, 2018.

\bibitem{liu2018automatic}
L.~Liu, G.~Feng, and D.~Beautemps, ``Automatic temporal segmentation of hand
  movements for hand positions recognition in french cued speech,'' in
  \emph{Proc. IEEE-ICASSP}, 2018, pp. 3061--3065.

\bibitem{wang2021cross}
J.~Wang, Z.~Tang, X.~Li, M.~Yu, Q.~Fang, and L.~Liu, ``Cross-modal knowledge
  distillation method for automatic cued speech recognition,'' in \emph{Proc.
  Interspeech}, 2021, p. 2986–2990.

\bibitem{liu2019automatic}
L.~Liu, J.~Li, G.~Feng, and X.-P.~S. Zhang, ``Automatic detection of the
  temporal segmentation of hand movements in british english cued speech.'' in
  \emph{Proc. Interspeech}, 2019, pp. 2285--2289.

\bibitem{park2022synctalkface}
S.~J. Park, M.~Kim, J.~Hong, J.~Choi, and Y.~M. Ro, ``Synctalkface: Talking
  face generation with precise lip-syncing via audio-lip memory,'' in
  \emph{Proc. Conf AAAI Artif. Intell.}, vol.~36, no.~2, 2022, pp. 2062--2070.

\bibitem{liu2017inneravsp}
L.~Liu, G.~Feng, and D.~Beautemps, ``Inner lips parameter estimation based on
  adaptive ellipse model,'' in \emph{International Conference on
  Auditory-Visual Speech Processing (AVSP)}, 2017.

\bibitem{liu2017automatic}
------, ``Automatic dynamic template tracking of inner lips based on clnf,'' in
  \emph{Proc. IEEE-ICASSP}, 2017, p. 5130–5134.

\bibitem{alibali1999illuminating}
M.~W. Alibali, M.~Bassok, K.~O. Solomon, S.~E. Syc, and S.~Goldin-Meadow,
  ``Illuminating mental representations through speech and gesture,''
  \emph{Psychological Science}, vol.~10, no.~4, pp. 327--333, 1999.

\bibitem{kang2016hands}
S.~Kang and B.~Tversky, ``From hands to minds: Gestures promote
  understanding,'' \emph{Cognitive Research: Principles and Implications},
  vol.~1, no.~1, pp. 1--15, 2016.

\bibitem{kendon1994gestures}
A.~Kendon, ``Do gestures communicate? a review,'' \emph{Research on language
  and social interaction}, vol.~27, no.~3, pp. 175--200, 1994.

\bibitem{kendon2004gesture}
K.~Adam, \emph{Gesture: Visible action as utterance}.\hskip 1em plus 0.5em
  minus 0.4em\relax Cambridge University Press, 2004.

\bibitem{ferstl2018investigating}
Y.~Ferstl and R.~McDonnell, ``Investigating the use of recurrent motion
  modelling for speech gesture generation,'' in \emph{Proc. ACM IVA}, 2018, pp.
  93--98.

\bibitem{yoon2019robots}
Y.~Yoon, W.-R. Ko, M.~Jang, J.~Lee, J.~Kim, and G.~Lee, ``Robots learn social
  skills: End-to-end learning of co-speech gesture generation for humanoid
  robots,'' in \emph{Proc. IEEE-International Conference in Robotics and
  Automation (ICRA)}, 2019, pp. 4303--4309.

\bibitem{yoon2022genea}
Y.~Yoon, P.~Wolfert, T.~Kucherenko, C.~Viegas, T.~Nikolov, M.~Tsakov, and G.~E.
  Henter, ``The genea challenge 2022: A large evaluation of data-driven
  co-speech gesture generation,'' in \emph{Proc. ACM-International Conference
  on Multimodal Interaction}, 2022, pp. 736--747.

\bibitem{ginosar2019learning}
S.~Ginosar, A.~Bar, G.~Kohavi, C.~Chan, A.~Owens, and J.~Malik, ``Learning
  individual styles of conversational gesture,'' in \emph{Proc. IEEE/CVF-CVPR},
  2019, pp. 3497--3506.

\bibitem{poppe2010survey}
R.~Poppe, ``A survey on vision-based human action recognition,'' \emph{Image
  and vision computing}, vol.~28, no.~6, pp. 976--990, 2010.

\bibitem{wu2016deep}
D.~Wu, L.~Pigou, P.-J. Kindermans, N.~D.-H. Le, L.~Shao, J.~Dambre, and J.-M.
  Odobez, ``Deep dynamic neural networks for multimodal gesture segmentation
  and recognition,'' \emph{IEEE transactions on pattern analysis and machine
  intelligence}, vol.~38, no.~8, pp. 1583--1597, 2016.

\bibitem{wan2016chalearn}
J.~Wan, Y.~Zhao, S.~Zhou, I.~Guyon, S.~Escalera, and S.~Z. Li, ``Chalearn
  looking at people rgb-d isolated and continuous datasets for gesture
  recognition,'' in \emph{Proceedings of the IEEE conference on computer vision
  and pattern recognition workshops}, 2016, pp. 56--64.

\bibitem{materzynska2019jester}
J.~Materzynska, G.~Berger, I.~Bax, and R.~Memisevic, ``The jester dataset: A
  large-scale video dataset of human gestures,'' in \emph{Proceedings of the
  IEEE/CVF international conference on computer vision workshops}, 2019, pp.
  0--0.

\bibitem{zeng2022gesturelens}
H.~Zeng, X.~Wang, Y.~Wang, A.~Wu, T.-C. Pong, and H.~Qu, ``Gesturelens: Visual
  analysis of gestures in presentation videos,'' \emph{IEEE Transactions on
  Visualization and Computer Graphics}, 2022.

\bibitem{efthimiou2010dicta}
E.~Efthimiou, S.-E. Fotinea, T.~Hanke, J.~Glauert, R.~Bowden, A.~Braffort,
  C.~Collet, P.~Maragos, and F.~Goudenove, ``Dicta-sign: sign language
  recognition, generation and modelling with application in deaf
  communication,'' in \emph{LREC}.\hskip 1em plus 0.5em minus 0.4em\relax
  European Language Resources Association (ELRA), 2010, pp. 80--83.

\bibitem{koller2015continuous}
O.~Koller, J.~Forster, and H.~Ney, ``Continuous sign language recognition:
  Towards large vocabulary statistical recognition systems handling multiple
  signers,'' \emph{Computer Vision and Image Understanding}, vol. 141, pp.
  108--125, 2015.

\bibitem{neidle2012challenges}
C.~Neidle, A.~Thangali, and S.~Sclaroff, ``Challenges in development of the
  american sign language lexicon video dataset (asllvd) corpus,'' in
  \emph{LREC}.\hskip 1em plus 0.5em minus 0.4em\relax Citeseer, 2012.

\bibitem{agris2010signum}
U.~v. Agris and K.-F. Kraiss, ``Signum database: Video corpus for
  signer-independent continuous sign language recognition,'' in
  \emph{LREC}.\hskip 1em plus 0.5em minus 0.4em\relax European Language
  Resources Association (ELRA), 2010, pp. 243--246.

\bibitem{lin2015curve}
Y.~Lin, X.~Chai, Y.~Zhou, and X.~Chen, ``Curve matching from the view of
  manifold for sign language recognition,'' in \emph{ACCV Workshops}, 2014.

\bibitem{caselli2017asl}
N.~K. Caselli, Z.~S. Sehyr, A.~M. Cohen-Goldberg, and K.~Emmorey, ``Asl-lex: A
  lexical database of american sign language,'' \emph{Behavior research
  methods}, vol.~49, pp. 784--801, 2017.

\bibitem{camgoz2018neural}
N.~C. Camgoz, S.~Hadfield, O.~Koller, H.~Ney, and R.~Bowden, ``Neural sign
  language translation,'' in \emph{Proc. IEEE-CVPR}, 2018, pp. 7784--7793.

\bibitem{mavi2022new}
A.~Mavi and Z.~Dikle, ``A new 27 class sign language dataset collected from 173
  individuals,'' \emph{arXiv preprint arXiv:2203.03859}, 2022.

\bibitem{ko2019neural}
S.-K. Ko, C.~J. Kim, H.~Jung, and C.~Cho, ``Neural sign language translation
  based on human keypoint estimation,'' \emph{Applied sciences}, vol.~9,
  no.~13, p. 2683, 2019.

\bibitem{adaloglou2021comprehensive}
N.~Adaloglou, T.~Chatzis, I.~Papastratis, A.~Stergioulas, G.~T. Papadopoulos,
  V.~Zacharopoulou, G.~J. Xydopoulos, K.~Atzakas, D.~Papazachariou, and
  P.~Daras, ``A comprehensive study on deep learning-based methods for sign
  language recognition,'' \emph{IEEE Transactions on Multimedia}, vol.~24, pp.
  1750--1762, 2021.

\bibitem{sehyr2021asl}
Z.~S. Sehyr, N.~Caselli, A.~M. Cohen-Goldberg, and K.~Emmorey, ``The asl-lex
  2.0 project: A database of lexical and phonological properties for 2,723
  signs in american sign language,'' \emph{The Journal of Deaf Studies and Deaf
  Education}, vol.~26, no.~2, pp. 263--277, 2021.

\bibitem{duarte2021how2sign}
A.~Duarte, S.~Palaskar, L.~Ventura, D.~Ghadiyaram, K.~DeHaan, F.~Metze,
  J.~Torres, and X.~Giro-i Nieto, ``How2sign: a large-scale multimodal dataset
  for continuous american sign language,'' in \emph{Proc. IEEE/CVF-CVPR}, 2021,
  pp. 2735--2744.

\bibitem{kapitanov2023slovo}
A.~Kapitanov, K.~Kvanchiani, A.~Nagaev, and E.~Petrova, ``Slovo: Russian sign
  language dataset,'' \emph{arXiv preprint arXiv:2305.14527}, 2023.

\bibitem{al2023rgb}
M.~Al-Barham, A.~Alsharkawi, M.~Al-Yaman, M.~Al-Fetyani, A.~Elnagar, A.~A.
  SaAleek, and M.~Al-Odat, ``Rgb arabic alphabets sign language dataset,''
  \emph{arXiv preprint arXiv:2301.11932}, 2023.

\bibitem{liu2018visual}
L.~Liu, H.~Thomas, G.~Feng, and B.~Denis, ``Visual recognition of continuous
  cued speech using a tandem cnn-hmm approach.'' in \emph{Proc. Interspeech},
  2018, pp. 2643--2647.

\bibitem{Trochymiuk2007VOT}
T.~A., ``{VOT and durational properties of selected segments in the speech of
  deaf and normally hearing children},'' \emph{Studia Phonetica Posnaniensia},
  vol.~8, pp. 111--142, 2007.

\bibitem{bigi2022clelfpc}
B.~Bigi, M.~Zimmermann, and C.~Andr{\'e}, ``Clelfpc: a large open multi-speaker
  corpus of french cued speech,'' in \emph{LREC}, 2022, pp. 987--994.

\bibitem{cooke2006Grid}
M.~Cooke, J.~Barker, S.~Cunningham, and X.~Shao, ``The grid audio-visual speech
  corpus (1.0) [data set],'' in \emph{Zenodo}.\hskip 1em plus 0.5em minus
  0.4em\relax Zenodo, 2006.

\bibitem{martin2006enterface}
O.~Martin, I.~Kotsia, B.~Macq, and I.~Pitas, ``The enterface'05 audio-visual
  emotion database,'' in \emph{Proc. IEEE-22nd international conference on data
  engineering workshops}, 2006, pp. 8--8.

\bibitem{rekik2016adaptive}
A.~Rekik, A.~Ben-Hamadou, and W.~Mahdi, ``An adaptive approach for lip-reading
  using image and depth data,'' \emph{Multimedia Tools and Applications},
  vol.~75, pp. 8609--8636, 2016.

\bibitem{cao2014crema}
H.~Cao, D.~G. Cooper, M.~K. Keutmann, R.~C. Gur, A.~Nenkova, and R.~Verma,
  ``Crema-d: Crowd-sourced emotional multimodal actors dataset,'' \emph{IEEE
  Transactions on Affective Computing}, vol.~5, no.~4, pp. 377--390, 2014.

\bibitem{czyzewski2017audio}
A.~Czyzewski, B.~Kostek, P.~Bratoszewski, J.~Kotus, and M.~Szykulski, ``An
  audio-visual corpus for multimodal automatic speech recognition,''
  \emph{Journal of Intelligent Information Systems}, vol.~49, pp. 167--192,
  2017.

\bibitem{chung2017lip}
J.~S. Chung and A.~Zisserman, ``Lip reading in the wild,'' \emph{Proc. ACCV},
  pp. 87--103, 2017.

\bibitem{busso2016msp}
C.~Busso, S.~Parthasarathy, A.~Burmania, M.~AbdelWahab, N.~Sadoughi, and E.~M.
  Provost, ``{MSP-IMPROV}: An acted corpus of dyadic interactions to study
  emotion perception,'' \emph{IEEE Transactions on Affective Computing},
  vol.~8, no.~1, pp. 67--80, 2016.

\bibitem{suwajanakorn2017synthesizing}
S.~Suwajanakorn, S.~M. Seitz, and I.~Kemelmacher-Shlizerman, ``Synthesizing
  obama: learning lip sync from audio,'' \emph{ACM Transactions on Graphics
  (ToG)}, vol.~36, no.~4, pp. 1--13, 2017.

\bibitem{nagrani2017voxceleb}
A.~Nagrani, J.~S. Chung, and A.~Zisserman, ``Voxceleb: a large-scale speaker
  identification dataset,'' \emph{Telephony}, vol.~3, pp. 33--039, 2017.

\bibitem{chung2018voxceleb2}
J.~S. Chung, A.~Nagrani, and A.~Zisserman, ``Voxceleb2: Deep speaker
  recognition,'' \emph{Proc. Interspeech}, 2018.

\bibitem{afouras2018deepAVSR}
T.~Afouras, J.~S. Chung, A.~Senior, O.~Vinyals, and A.~Zisserman, ``Deep
  audio-visual speech recognition,'' \emph{IEEE transactions on pattern
  analysis and machine intelligence}, vol.~44, no.~12, pp. 8717--8727, 2018.

\bibitem{afouras2018lrs3}
T.~Afouras, J.~S. Chung, and A.~Zisserman, ``{LRS3-TED}: a large-scale dataset
  for visual speech recognition,'' \emph{arXiv preprint arXiv:1809.00496},
  2018.

\bibitem{livingstone2018ryerson}
S.~R. Livingstone and F.~A. Russo, ``The ryerson audio-visual database of
  emotional speech and song (ravdess): A dynamic, multimodal set of facial and
  vocal expressions in north american english,'' \emph{PloS one}, vol.~13,
  no.~5, p. e0196391, 2018.

\bibitem{poria2018meld}
S.~Poria, D.~Hazarika, N.~Majumder, G.~Naik, E.~Cambria, and R.~Mihalcea,
  ``Meld: A multimodal multi-party dataset for emotion recognition in
  conversations,'' \emph{arXiv preprint arXiv:1810.02508}, 2018.

\bibitem{ephrat2018looking}
A.~Ephrat, I.~Mosseri, O.~Lang, T.~Dekel, K.~Wilson, A.~Hassidim, W.~T.
  Freeman, and M.~Rubinstein, ``Looking to listen at the cocktail party: a
  speaker-independent audio-visual model for speech separation,'' \emph{ACM
  Transactions on Graphics (TOG)}, vol.~37, no.~4, pp. 1--11, 2018.

\bibitem{cudeiro2019capture}
D.~Cudeiro, T.~Bolkart, C.~Laidlaw, A.~Ranjan, and M.~J. Black, ``Capture,
  learning, and synthesis of 3d speaking styles,'' in \emph{Proc.
  IEEE/CVF-CVPR}, 2019, pp. 10\,101--10\,111.

\bibitem{yang2019lrw}
S.~Yang, Y.~Zhang, D.~Feng, M.~Yang, C.~Wang, J.~Xiao, K.~Long, S.~Shan, and
  X.~Chen, ``{LRW}-1000: A naturally-distributed large-scale benchmark for lip
  reading in the wild,'' in \emph{Proceedings of 14th IEEE international
  conference on automatic face \& gesture recognition}, 2019, pp. 1--8.

\bibitem{rossler2019faceforensics++}
A.~Rossler, D.~Cozzolino, L.~Verdoliva, C.~Riess, J.~Thies, and M.~Nie{\ss}ner,
  ``Faceforensics++: Learning to detect manipulated facial images,'' in
  \emph{Proc. IEEE/CVF-ICCV}, 2019, pp. 1--11.

\bibitem{wang2020mead}
K.~Wang, Q.~Wu, L.~Song, Z.~Yang, W.~Wu, C.~Qian, R.~He, Y.~Qiao, and C.~C.
  Loy, ``Mead: A large-scale audio-visual dataset for emotional talking-face
  generation,'' in \emph{Proc. ECCV}, 2020, pp. 700--717.

\bibitem{zhang2021flow}
Z.~Zhang, L.~Li, Y.~Ding, and C.~Fan, ``Flow-guided one-shot talking face
  generation with a high-resolution audio-visual dataset,'' in \emph{Proc.
  IEEE/CVF-CVPR}, 2021, pp. 3661--3670.

\bibitem{kim2022animeceleb}
K.~Kim, S.~Park, J.~Lee, S.~Chung, J.~Lee, and J.~Choo, ``{AnimeCeleb}:
  Large-scale animation celebheads dataset for head reenactment,'' in
  \emph{Proc. ECCV}, 2022, pp. 414--430.

\bibitem{berkol2023visual}
A.~Berkol, T.~T{\"u}mer-Sivri, N.~Pervan-Akman, M.~{\c{C}}olak, and H.~Erdem,
  ``Visual lip reading dataset in turkish,'' \emph{Data}, vol.~8, no.~1, p.~15,
  2023.

\bibitem{hwang2023discohead}
G.~Hwang, S.~Hong, S.~Lee, S.~Park, and G.~Chae, ``{DisCoHead}:
  Audio-and-video-driven talking head generation by disentangled control of
  head pose and facial expressions,'' in \emph{Proc. IEEE-ICASSP}, 2023, pp.
  1--5.

\bibitem{matthews2002extraction}
I.~Matthews, T.~F. Cootes, J.~A. Bangham, S.~Cox, and R.~Harvey, ``Extraction
  of visual features for lipreading,'' \emph{IEEE Transactions on Pattern
  Analysis and Machine Intelligence}, vol.~24, no.~2, pp. 198--213, 2002.

\bibitem{petridis2018visual}
S.~Petridis, J.~Shen, D.~Cetin, and M.~Pantic, ``Visual-only recognition of
  normal, whispered and silent speech,'' in \emph{Proc. IEEE-ICASSP}, 2018, pp.
  6219--6223.

\bibitem{takeuchi2017creating}
K.~Takeuchi, S.~Kubota, K.~Suzuki, D.~Hasegawa, and H.~Sakuta, ``Creating a
  gesture-speech dataset for speech-based automatic gesture generation,''
  \emph{Communications in Computer and Information Science}, pp. 198--202,
  2017.

\bibitem{singh2018p2pstory}
N.~Singh, J.~J. Lee, I.~Grover, and C.~Breazeal, ``P2pstory: dataset of
  children as storytellers and listeners in peer-to-peer interactions,'' in
  \emph{Proceedings of the CHI Conference on Human Factors in Computing
  Systems}, 2018, pp. 1--11.

\bibitem{mahmood2019amass}
N.~Mahmood, N.~Ghorbani, N.~F. Troje, G.~Pons-Moll, and M.~J. Black, ``{AMASS}:
  Archive of motion capture as surface shapes,'' in \emph{Proc. IEEE/CVF-ICCV},
  2019, pp. 5442--5451.

\bibitem{luo2020arbee}
Y.~Luo, J.~Ye, R.~B. Adams, J.~Li, M.~G. Newman, and J.~Z. Wang, ``{ARBEE}:
  Towards automated recognition of bodily expression of emotion in the wild,''
  \emph{International journal of computer vision}, vol. 128, pp. 1--25, 2020.

\bibitem{ahuja2020no}
C.~Ahuja, D.~W. Lee, R.~Ishii, and L.-P. Morency, ``No gestures left behind:
  Learning relationships between spoken language and freeform gestures,'' in
  \emph{Findings of the Association for Computational Linguistics: EMNLP},
  2020, pp. 1884--1895.

\bibitem{punnakkal2021babel}
A.~R. Punnakkal, A.~Chandrasekaran, N.~Athanasiou, A.~Quiros-Ramirez, and M.~J.
  Black, ``Babel: Bodies, action and behavior with english labels,'' in
  \emph{Proc. IEEE/CVF-CVPR}, 2021, pp. 722--731.

\bibitem{guo2022generating}
C.~Guo, S.~Zou, X.~Zuo, S.~Wang, W.~Ji, X.~Li, and L.~Cheng, ``Generating
  diverse and natural 3d human motions from text,'' in \emph{Proc.
  IEEE/CVF-CVPR}, 2022, pp. 5152--5161.

\bibitem{liu2022beat}
H.~Liu, Z.~Zhu, N.~Iwamoto, Y.~Peng, Z.~Li, Y.~Zhou, E.~Bozkurt, and B.~Zheng,
  ``Beat: A large-scale semantic and emotional multi-modal dataset for
  conversational gestures synthesis,'' in \emph{Proc. ECCV}, 2022, pp.
  612--630.

\bibitem{wang2023emotional}
J.~Wang, Y.~Zhao, L.~Liu, T.~Xu, Q.~Li, and S.~Li, ``Emotional talking head
  generation based on memory-sharing and attention-augmented networks,''
  \emph{arXiv preprint arXiv:2306.03594}, 2023.

\bibitem{wang2023memory}
J.~Wang, Y.~Zhao, H.~Fan, T.~Xu, Q.~Li, S.~Li, and L.~Liu, ``Memory-augmented
  contrastive learning for talking head generation,'' in \emph{Proc.
  IEEE-ICASSP}, 2023, p. 1–5.

\bibitem{cosatto2003lifelike}
E.~Cosatto, J.~Ostermann, H.~P. Graf, and J.~Schroeter, ``Lifelike talking
  faces for interactive services,'' \emph{Proceedings of the IEEE}, vol.~91,
  no.~9, pp. 1406--1429, 2003.

\bibitem{gambino2008virtual}
O.~Gambino, A.~Augello, A.~Caronia, G.~Pilato, R.~Pirrone, and S.~Gaglio,
  ``Virtual conversation with a real talking head,'' in \emph{Proc.
  IEEE-Conference on Human System Interactions}, 2008, pp. 263--268.

\bibitem{yang2012articulated}
Y.~Yang and D.~Ramanan, ``Articulated human detection with flexible mixtures of
  parts,'' \emph{IEEE Transactions on Pattern Analysis and Machine
  Intelligence}, vol.~35, no.~12, pp. 2878--2890, 2012.

\bibitem{yoon2020speech}
Y.~Yoon, B.~Cha, J.-H. Lee, M.~Jang, J.~Lee, J.~Kim, and G.~Lee, ``Speech
  gesture generation from the trimodal context of text, audio, and speaker
  identity,'' \emph{ACM Transactions on Graphics (TOG)}, vol.~39, no.~6, pp.
  1--16, 2020.

\bibitem{asakawa2022evaluation}
E.~Asakawa, N.~Kaneko, D.~Hasegawa, and S.~Shirakawa, ``Evaluation of
  text-to-gesture generation model using convolutional neural network,''
  \emph{Neural Networks}, vol. 151, pp. 365--375, 2022.

\bibitem{buehler2009learning}
P.~Buehler, A.~Zisserman, and M.~Everingham, ``Learning sign language by
  watching tv (using weakly aligned subtitles),'' in \emph{Proc. IEEE-CVPR},
  2009, pp. 2961--2968.

\bibitem{wang2019novel}
H.~Wang, X.~Chai, and X.~Chen, ``A novel sign language recognition framework
  using hierarchical grassmann covariance matrix,'' \emph{IEEE Transactions on
  Multimedia}, vol.~21, no.~11, pp. 2806--2814, 2019.

\bibitem{pfister2013large}
T.~Pfister, J.~Charles, and A.~Zisserman, ``Large-scale learning of sign
  language by watching tv (using co-occurrences).'' in \emph{Proc. BMVC}.\hskip
  1em plus 0.5em minus 0.4em\relax British Machine Vision Association, 2013.

\bibitem{he2016deep}
K.~He, X.~Zhang, S.~Ren, and J.~Sun, ``Deep residual learning for image
  recognition,'' in \emph{Proc. IEEE-CVPR}, 2016, pp. 770--778.

\bibitem{carreira2017quo}
J.~Carreira and A.~Zisserman, ``Quo vadis, action recognition? a new model and
  the kinetics dataset,'' in \emph{Proc. IEEE-CVPR}, 2017, pp. 6299--6308.

\bibitem{qiu2017learning}
Z.~Qiu, T.~Yao, and T.~Mei, ``Learning spatio-temporal representation with
  pseudo-3d residual networks,'' in \emph{Proc. IEEE-ICCV}, 2017, pp.
  5533--5541.

\bibitem{qiu2019learning}
Z.~Qiu, T.~Yao, C.-W. Ngo, X.~Tian, and T.~Mei, ``Learning spatio-temporal
  representation with local and global diffusion,'' in \emph{Proc.
  IEEE/CVF-CVPR}, 2019, pp. 12\,056--12\,065.

\bibitem{hu2022collaborative}
H.~Hu, J.~Pu, W.~Zhou, and H.~Li, ``Collaborative multilingual continuous sign
  language recognition: A unified framework,'' \emph{IEEE Transactions on
  Multimedia}, 2022.

\bibitem{cui2019deep}
R.~Cui, H.~Liu, and C.~Zhang, ``A deep neural framework for continuous sign
  language recognition by iterative training,'' \emph{IEEE Transactions on
  Multimedia}, vol.~21, no.~7, pp. 1880--1891, 2019.

\bibitem{pu2020boosting}
J.~Pu, W.~Zhou, H.~Hu, and H.~Li, ``Boosting continuous sign language
  recognition via cross modality augmentation,'' in \emph{Proc. ACM MM}, 2020,
  pp. 1497--1505.

\bibitem{pu2019iterative}
J.~Pu, W.~Zhou, and H.~Li, ``Iterative alignment network for continuous sign
  language recognition,'' in \emph{Proc. IEEE/CVF-CVPR}, 2019, pp. 4165--4174.

\bibitem{huang2018video}
J.~Huang, W.~Zhou, Q.~Zhang, H.~Li, and W.~Li, ``Video-based sign language
  recognition without temporal segmentation,'' in \emph{Proc. Conf AAAI Artif.
  Intell.}, vol.~32, no.~1, 2018.

\bibitem{hu2021global}
H.~Hu, W.~Zhou, J.~Pu, and H.~Li, ``Global-local enhancement network for
  nmf-aware sign language recognition,'' \emph{ACM transactions on multimedia
  computing, communications, and applications (TOMM)}, vol.~17, no.~3, pp.
  1--19, 2021.

\bibitem{wei2019deep}
C.~Wei, W.~Zhou, J.~Pu, and H.~Li, ``Deep grammatical multi-classifier for
  continuous sign language recognition,'' in \emph{International Conference on
  Multimedia Big Data (BigMM)}, 2019, pp. 435--442.

\bibitem{guo2019dense}
D.~Guo, S.~Wang, Q.~Tian, and M.~Wang, ``Dense temporal convolution network for
  sign language translation.'' in \emph{Proc.IJCAI}, 2019, pp. 744--750.

\bibitem{zhou2019dynamic}
H.~Zhou, W.~Zhou, and H.~Li, ``Dynamic pseudo label decoding for continuous
  sign language recognition,'' in \emph{International conference on multimedia
  and expo (ICME)}, 2019, pp. 1282--1287.

\bibitem{nadeemhashmi2018lip}
S.~NadeemHashmi, H.~Gupta, D.~Mittal, K.~Kumar, A.~Nanda, and S.~Gupta, ``A lip
  reading model using cnn with batch normalization,'' in \emph{Proc. IEEE-11th
  international conference on contemporary computing (IC3)}, 2018, pp. 1--6.

\bibitem{slimane2021context}
F.~B. Slimane and M.~Bouguessa, ``Context matters: Self-attention for sign
  language recognition,'' in \emph{International Conference on Pattern
  Recognition (ICPR)}, 2021, pp. 7884--7891.

\bibitem{zhou2020self}
M.~Zhou, M.~Ng, Z.~Cai, and K.~C. Cheung, ``Self-attention-based
  fully-inception networks for continuous sign language recognition,'' in
  \emph{24th European Conference on Artificial Intelligence}, 2020, pp.
  2832--2839.

\bibitem{koller2019weakly}
O.~Koller, N.~C. Camgoz, H.~Ney, and R.~Bowden, ``Weakly supervised learning
  with multi-stream cnn-lstm-hmms to discover sequential parallelism in sign
  language videos,'' \emph{IEEE transactions on pattern analysis and machine
  intelligence}, vol.~42, no.~9, pp. 2306--2320, 2019.

\bibitem{koller2017re}
O.~Koller, S.~Zargaran, and H.~Ney, ``{Re-sign: Re-aligned end-to-end sequence
  modelling with deep recurrent CNN-HMMs},'' in \emph{Proc. IEEE-CVPR}, 2017,
  pp. 4297--4305.

\bibitem{koller2018deep}
O.~Koller, S.~Zargaran, H.~Ney, and R.~Bowden, ``Deep sign: Enabling robust
  statistical continuous sign language recognition via hybrid cnn-hmms,''
  \emph{International Journal of Computer Vision}, vol. 126, pp. 1311--1325,
  2018.

\bibitem{cui2017recurrent}
R.~Cui, H.~Liu, and C.~Zhang, ``Recurrent convolutional neural networks for
  continuous sign language recognition by staged optimization,'' in \emph{Proc.
  IEEE-CVPR}, 2017, pp. 7361--7369.

\bibitem{min2021visual}
Y.~Min, A.~Hao, X.~Chai, and X.~Chen, ``Visual alignment constraint for
  continuous sign language recognition,'' in \emph{Proc. IEEE/CVF-ICCV}, 2021,
  pp. 11\,542--11\,551.

\bibitem{cihan2017subunets}
N.~Cihan~Camgoz, S.~Hadfield, O.~Koller, and R.~Bowden, ``Subunets: End-to-end
  hand shape and continuous sign language recognition,'' in \emph{Proc.
  IEEE-ICCV}, 2017, pp. 3056--3065.

\bibitem{guo2019hierarchical}
D.~Guo, W.~Zhou, A.~Li, H.~Li, and M.~Wang, ``Hierarchical recurrent deep
  fusion using adaptive clip summarization for sign language translation,''
  \emph{IEEE Transactions on Image Processing}, vol.~29, pp. 1575--1590, 2019.

\bibitem{li2020key}
H.~Li, L.~Gao, R.~Han, L.~Wan, and W.~Feng, ``Key action and joint
  ctc-attention based sign language recognition,'' in \emph{IEEE International
  Conference on Acoustics, Speech and Signal Processing (ICASSP)}, 2020, pp.
  2348--2352.

\bibitem{hao2021self}
A.~Hao, Y.~Min, and X.~Chen, ``Self-mutual distillation learning for continuous
  sign language recognition,'' in \emph{Proc. IEEE/CVF-ICCV}, 2021, pp.
  11\,303--11\,312.

\bibitem{zhou2021spatial}
H.~Zhou, W.~Zhou, Y.~Zhou, and H.~Li, ``Spatial-temporal multi-cue network for
  sign language recognition and translation,'' \emph{IEEE Transactions on
  Multimedia}, vol.~24, pp. 768--779, 2021.

\bibitem{pei2019continuous}
X.~Pei, D.~Guo, and Y.~Zhao, ``Continuous sign language recognition based on
  pseudo-supervised learning,'' in \emph{Proceedings of the 2nd Workshop on
  Multimedia for Accessible Human Computer Interfaces}, 2019, pp. 33--39.

\bibitem{zhang2019continuous}
Z.~Zhang, J.~Pu, L.~Zhuang, W.~Zhou, and H.~Li, ``Continuous sign language
  recognition via reinforcement learning,'' in \emph{Proc. IEEE-ICIP}.\hskip
  1em plus 0.5em minus 0.4em\relax IEEE, 2019, pp. 285--289.

\bibitem{niu2020stochastic}
Z.~Niu and B.~Mak, ``Stochastic fine-grained labeling of multi-state sign
  glosses for continuous sign language recognition,'' in \emph{Proc. ECCV},
  2020, pp. 172--186.

\bibitem{koishybay2021continuous}
K.~Koishybay, M.~Mukushev, and A.~Sandygulova, ``Continuous sign language
  recognition with iterative spatiotemporal fine-tuning,'' in \emph{Proc.
  IEEE-ICPR}, 2021, pp. 10\,211--10\,218.

\bibitem{Papastratis2021Continuous}
I.~Papastratis, K.~Dimitropoulos, and P.~Daras, ``Continuous sign language
  recognition through a context-aware generative adversarial network,''
  \emph{Sensors}, vol.~21, no.~7, 2021.

\bibitem{chen2022two}
Y.~Chen, R.~Zuo, F.~Wei, Y.~Wu, S.~Liu, and B.~Mak, ``Two-stream network for
  sign language recognition and translation,'' \emph{Advances in Neural
  Information Processing Systems (NIPS)}, vol.~35, pp. 17\,043--17\,056, 2022.

\bibitem{hu2023self}
L.~Hu, L.~Gao, Z.~Liu, and W.~Feng, ``Self-emphasizing network for continuous
  sign language recognition,'' in \emph{Proc. Conf AAAI Artif. Intell.},
  vol.~37, no.~1, 2023, pp. 854--862.

\bibitem{zheng2023cvt}
J.~Zheng, Y.~Wang, C.~Tan, S.~Li, G.~Wang, J.~Xia, Y.~Chen, and S.~Z. Li,
  ``Cvt-slr: Contrastive visual-textual transformation for sign language
  recognition with variational alignment,'' in \emph{Proc. IEEE/CVF-CVPR},
  2023, pp. 23\,141--23\,150.

\bibitem{papadimitriou2021fully}
K.~Papadimitriou and G.~Potamianos, ``A fully convolutional sequence learning
  approach for cued speech recognition from videos,'' in \emph{Proc.
  IEEE-EUSIPCO}, 2021, pp. 326--330.

\bibitem{sankar2022multistream}
S.~Sankar, D.~Beautemps, and T.~Hueber, ``Multistream neural architectures for
  cued speech recognition using a pre-trained visual feature extractor and
  constrained ctc decoding,'' in \emph{Proc. IEEE-ICASSP}, 2022, pp.
  8477--8481.

\bibitem{kipp2011sign}
M.~Kipp, A.~Heloir, and Q.~Nguyen, ``Sign language avatars: Animation and
  comprehensibility,'' in \emph{Intelligent Virtual Agents}.\hskip 1em plus
  0.5em minus 0.4em\relax Springer Berlin Heidelberg, 2011, pp. 113--126.

\bibitem{mcdonald2016automated}
J.~McDonald, R.~Wolfe, J.~Schnepp, J.~Hochgesang, D.~G. Jamrozik, M.~Stumbo,
  L.~Berke, M.~Bialek, and F.~Thomas, ``An automated technique for real-time
  production of lifelike animations of american sign language,''
  \emph{Universal Access in the Information Society}, vol.~15, pp. 551--566,
  2016.

\bibitem{gibet2016interactive}
S.~Gibet, F.~Lefebvre-Albaret, L.~Hamon, R.~Brun, and A.~Turki, ``Interactive
  editing in french sign language dedicated to virtual signers: Requirements
  and challenges,'' \emph{Universal Access in the Information Society},
  vol.~15, pp. 525--539, 2016.

\bibitem{Zelinka2020Neural}
J.~Zelinka and J.~Kanis, ``Neural sign language synthesis: Words are our
  glosses,'' in \emph{Proc. IEEE/CVF-WACV}, March 2020.

\bibitem{camgoz2020multichannel}
N.~C. Camgoz, O.~Koller, S.~Hadfield, and R.~Bowden, ``Multi-channel
  transformers for multi-articulatory sign language translation,'' 2020.

\bibitem{Saunders2020AdversarialTF}
B.~Saunders, N.~C. Camg{\"o}z, and R.~Bowden, ``Adversarial training for
  multi-channel sign language production,'' in \emph{The 31st British Machine
  Vision Virtual Conference}.\hskip 1em plus 0.5em minus 0.4em\relax British
  Machine Vision Association, 2020.

\bibitem{i̇nan2022modeling}
M.~Inan, Y.~Zhong, S.~Hassan, L.~Quandt, and M.~Alikhani, ``Modeling
  intensification for sign language generation: A computational approach,'' in
  \emph{Findings of the Association for Computational Linguistics}, 2022, pp.
  2897--2911.

\bibitem{Saunders2022signgan}
B.~Saunders, N.~C. Camgoz, and R.~Bowden, ``Signing at scale: Learning to
  co-articulate signs for large-scale photo-realistic sign language
  production,'' in \emph{Proc. IEEE/CVF-CVPR}, 2022, pp. 5141--5151.

\bibitem{xie2022vector}
P.~Xie, Q.~Zhang, Z.~Li, H.~Tang, Y.~Du, and X.~Hu, ``Vector quantized
  diffusion model with codeunet for text-to-sign pose sequences generation,''
  \emph{arXiv preprint arXiv:2208.09141}, 2022.

\bibitem{chiu2015predicting}
C.-C. Chiu, L.-P. Morency, and S.~Marsella, ``Predicting co-verbal gestures: A
  deep and temporal modeling approach,'' in \emph{Proc. Intelligent Virtual
  Agents}.\hskip 1em plus 0.5em minus 0.4em\relax Springer International
  Publishing, 2015, pp. 152--166.

\bibitem{alexanderson2020style}
S.~Alexanderson, G.~E. Henter, T.~Kucherenko, and J.~Beskow,
  ``Style-controllable speech-driven gesture synthesis using normalising
  flows,'' in \emph{Computer Graphics Forum}, vol.~39, no.~2.\hskip 1em plus
  0.5em minus 0.4em\relax Wiley Online Library, 2020, pp. 487--496.

\bibitem{li2021audio2gestures}
J.~Li, D.~Kang, W.~Pei, X.~Zhe, Y.~Zhang, Z.~He, and L.~Bao, ``Audio2gestures:
  Generating diverse gestures from speech audio with conditional variational
  autoencoders,'' in \emph{Proc. IEEE/CVF-ICCV}, 2021, pp. 11\,293--11\,302.

\bibitem{bhattacharya2021text2gestures}
U.~Bhattacharya, N.~Rewkowski, A.~Banerjee, P.~Guhan, A.~Bera, and D.~Manocha,
  ``Text2gestures: A transformer-based network for generating emotive body
  gestures for virtual agents,'' in \emph{IEEE virtual reality and 3D user
  interfaces (VR)}, 2021, pp. 1--10.

\bibitem{ghorbani2022exemplar}
S.~Ghorbani, Y.~Ferstl, and M.-A. Carbonneau, ``Exemplar-based stylized gesture
  generation from speech: An entry to the {GENEA} challenge 2022,'' in
  \emph{Proc. ACM-International Conference on Multimodal Interaction}, 2022,
  pp. 778--783.

\bibitem{li2008novel}
M.~Li and Y.-m. Cheung, ``A novel motion based lip feature extraction for
  lip-reading,'' in \emph{Proc. IEEE-International Conference on Computational
  Intelligence and Security}, vol.~1, 2008, pp. 361--365.

\bibitem{alizadeh2008lip}
S.~Alizadeh, R.~Boostani, and V.~Asadpour, ``Lip feature extraction and
  reduction for hmm-based visual speech recognition systems,'' in \emph{Pro.
  IEEE-9th International Conference on Signal Processing}, 2008, pp. 561--564.

\bibitem{ma2016lip}
X.~Ma, L.~Yan, and Q.~Zhong, ``Lip feature extraction based on improved
  jumping-snake model,'' in \emph{Proc. IEEE-35th Chinese Control Conference
  (CCC)}, 2016, pp. 6928--6933.

\bibitem{lan2012view}
Y.~Lan, B.-J. Theobald, and R.~Harvey, ``View independent computer
  lip-reading,'' in \emph{Proc. IEEE-International Conference on Multimedia and
  Expo}, 2012, pp. 432--437.

\bibitem{watanabe2017lip}
T.~Watanabe, K.~Katsurada, and Y.~Kanazawa, ``Lip reading from multi view
  facial images using {3D-AAM},'' in \emph{Proc. ACCV}.\hskip 1em plus 0.5em
  minus 0.4em\relax Springer Verlag, 2017, pp. 303--316.

\bibitem{liu2017inner}
L.~Liu, G.~Feng, and D.~Beautemps, ``Inner lips feature extraction based on
  clnf with hybrid dynamic template for cued speech,'' \emph{EURASIP Journal on
  Image and Video Processing}, vol. 2017, p. 1–15, 2017.

\bibitem{liu2016extraction}
------, ``Extraction automatique de contour de levre a partir du modele clnf,''
  in \emph{JEP-TALN-RECITAL 2016-conference conjointe 31e Journees d’Etudes
  sur la Parole, 23e Traitement Automatique des Langues Naturelles, 18e
  Rencontre des Etudiants Chercheurs en Informatique pour le Traitement
  Automatique des Langues}, 2016.

\bibitem{wang2021three}
J.~Wang, T.~Wu, S.~Wang, M.~Yu, Q.~Fang, J.~Zhang, and L.~Liu,
  ``Three-dimensional lip motion network for text-independent speaker
  recognition,'' in \emph{Proc. IEEE-ICPR}, 2021, p. 3380–3387.

\bibitem{garg2016lip}
A.~Garg, J.~Noyola, and S.~Bagadia, ``Lip reading using cnn and lstm,''
  \emph{Technical report, Stanford University, CS231 n project report}, 2016.

\bibitem{lee2016multi}
D.~Lee, J.~Lee, and K.-E. Kim, ``Multi-view automatic lip-reading using neural
  network,'' in \emph{Proc. ACCV Workshop on Multi-view Lip-reading
  Challenges}, 2016.

\bibitem{fung2018end}
I.~Fung and B.~Mak, ``End-to-end low-resource lip-reading with maxout {CNN} and
  {LSTM},'' in \emph{Proc. IEEE-ICASSP}, 2018, pp. 2511--2515.

\bibitem{xu2018lcanet}
K.~Xu, D.~Li, N.~Cassimatis, and X.~Wang, ``{LCANet}: End-to-end lipreading
  with cascaded attention-{CTC},'' in \emph{Proc. IEEE-FG}, 2018, pp. 548--555.

\bibitem{wiriyathammabhum2020spotfast}
P.~Wiriyathammabhum, ``Spotfast networks with memory augmented lateral
  transformers for lipreading,'' in \emph{International Conference on Neural
  Information Processing}, 2020, pp. 554--561.

\bibitem{weng2019learning}
X.~Weng and K.~Kitani, ``Learning spatio-temporal features with two-stream deep
  3d cnns for lipreading,'' \emph{arXiv preprint arXiv:1905.02540}, 2019.

\bibitem{feng2021efficient}
D.~Feng, S.~Yang, and S.~Shan, ``An efficient software for building lip reading
  models without pains,'' in \emph{Proc. IEEE-ICMEW}, 2021, pp. 1--2.

\bibitem{xu2020discriminative}
B.~Xu, C.~Lu, Y.~Guo, and J.~Wang, ``Discriminative multi-modality speech
  recognition,'' in \emph{Proc. IEEE/CVF-CVPR}, 2020, pp. 14\,433--14\,442.

\bibitem{luo2020pseudo}
M.~Luo, S.~Yang, S.~Shan, and X.~Chen, ``Pseudo-convolutional policy gradient
  for sequence-to-sequence lip-reading,'' in \emph{Proc. IEEE-FG}, 2020, pp.
  273--280.

\bibitem{gehring2013extracting}
J.~Gehring, Y.~Miao, F.~Metze, and A.~Waibel, ``Extracting deep bottleneck
  features using stacked auto-encoders,'' in \emph{Proc. IEEE-ICASSP}, 2013,
  pp. 3377--3381.

\bibitem{noda2015audio}
K.~Noda, Y.~Yamaguchi, K.~Nakadai, H.~G. Okuno, and T.~Ogata, ``Audio-visual
  speech recognition using deep learning,'' \emph{Applied intelligence},
  vol.~42, pp. 722--737, 2015.

\bibitem{petridis2016deep}
S.~Petridis and M.~Pantic, ``Deep complementary bottleneck features for visual
  speech recognition,'' in \emph{Proc. IEEE-ICASSP}, 2016, pp. 2304--2308.

\bibitem{wang2021attention}
J.~Wang, N.~Gu, M.~Yu, X.~Li, Q.~Fang, and L.~Liu, ``An attention
  self-supervised contrastive learning based three-stage model for hand shape
  feature representation in cued speech,'' \emph{arXiv preprint
  arXiv:2106.14016}, 2021.

\bibitem{wand2018investigations}
M.~Wand, J.~Schmidhuber, and N.~T. Vu, ``Investigations on end-to-end
  audiovisual fusion,'' in \emph{Proc. IEEE-ICASSP}, 2018, pp. 3041--3045.

\bibitem{zhang2020can}
Y.~Zhang, S.~Yang, J.~Xiao, S.~Shan, and X.~Chen, ``Can we read speech beyond
  the lips? rethinking roi selection for deep visual speech recognition,'' in
  \emph{Proc. IEEE-FG}, 2020, pp. 356--363.

\bibitem{xiao2020deformation}
J.~Xiao, S.~Yang, Y.~Zhang, S.~Shan, and X.~Chen, ``Deformation flow based
  two-stream network for lip reading,'' in \emph{Proc. IEEE-FG}, 2020, pp.
  364--370.

\bibitem{zhao2020mutual}
X.~Zhao, S.~Yang, S.~Shan, and X.~Chen, ``Mutual information maximization for
  effective lip reading,'' in \emph{Proc. IEEE-FG}, 2020, pp. 420--427.

\bibitem{bai2018empirical}
S.~Bai, J.~Z. Kolter, and V.~Koltun, ``An empirical evaluation of generic
  convolutional and recurrent networks for sequence modeling,'' \emph{arXiv
  preprint arXiv:1803.01271}, 2018.

\bibitem{martinez2020lipreading}
B.~Martinez, P.~Ma, S.~Petridis, and M.~Pantic, ``Lipreading using temporal
  convolutional networks,'' in \emph{Proc. IEEE-ICASSP}, 2020, pp. 6319--6323.

\bibitem{vaswani2017attention}
V.~Ashish, S.~Noam, P.~Niki, U.~Jakob, J.~Llion, G.~A. N, K.~{\L}ukasz, and
  P.~Illia, ``Attention is all you need,'' in \emph{Advances in Neural
  Information Processing Systems (NIPS)}, 2017.

\bibitem{afouras2018deepLR}
T.~Afouras, J.~S. Chung, and A.~Zisserman, ``Deep lip reading: a comparison of
  models and an online application,'' \emph{arXiv preprint arXiv:1806.06053},
  2018.

\bibitem{son2017lip}
J.~Son~Chung, A.~Senior, O.~Vinyals, and A.~Zisserman, ``Lip reading sentences
  in the wild,'' in \emph{Proc. IEEE-CVPR}, 2017, pp. 6447--6456.

\bibitem{lu2019automatic}
Y.~Lu and H.~Li, ``Automatic lip-reading system based on deep convolutional
  neural network and attention-based long short-term memory,'' \emph{Applied
  Sciences}, vol.~9, no.~8, p. 1599, 2019.

\bibitem{zhou2019modality}
P.~Zhou, W.~Yang, W.~Chen, Y.~Wang, and J.~Jia, ``Modality attention for
  end-to-end audio-visual speech recognition,'' in \emph{Proc. IEEE-ICASSP},
  2019, pp. 6565--6569.

\bibitem{zhang2019understanding}
X.~Zhang, H.~Gong, X.~Dai, F.~Yang, N.~Liu, and M.~Liu, ``Understanding
  pictograph with facial features: End-to-end sentence-level lip reading of
  chinese,'' in \emph{Proc. Conf AAAI Artif. Intell.}, vol.~33, no.~01, 2019,
  pp. 9211--9218.

\bibitem{torfi20173d}
A.~Torfi, S.~M. Iranmanesh, N.~Nasrabadi, and J.~Dawson, ``3d convolutional
  neural networks for cross audio-visual matching recognition,'' \emph{IEEE
  Access}, vol.~5, pp. 22\,081--22\,091, 2017.

\bibitem{heracleous2010cued}
P.~Heracleous, D.~Beautemps, and N.~Aboutabit, ``Cued speech automatic
  recognition in normal-hearing and deaf subjects,'' \emph{Speech
  Communication}, vol.~52, no.~6, pp. 504--512, 2010.

\bibitem{heracleous2012continuous}
P.~Heracleous, D.~Beautemps, and N.~Hagita, ``Continuous phoneme recognition in
  cued speech for french,'' in \emph{Proc. IEEE-EUSIPCO}, 2012, pp. 2090--2093.

\bibitem{burger2005cued}
T.~Burger, A.~Caplier, and S.~Mancini, ``Cued speech hand gestures recognition
  tool,'' in \emph{Proc. IEEE-EUSIPCO}, 2005, pp. 1--4.

\bibitem{gao2023novel}
L.~Gao, S.~Huang, and L.~Liu, ``A novel interpretable and generalizable
  re-synchronization model for cued speech based on a multi-cuer corpus,''
  \emph{arXiv preprint arXiv:2306.02596}, 2023.

\bibitem{zhang2023cuing}
Y.~Zhang, L.~Liu, and L.~Liu, ``Cuing without sharing: A federated cued speech
  recognition framework via mutual knowledge distillation,'' \emph{arXiv
  preprint arXiv:2308.03432}, 2023.

\bibitem{graves2006connectionist}
A.~Graves, S.~Fern{\'a}ndez, F.~Gomez, and J.~Schmidhuber, ``Connectionist
  temporal classification: labelling unsegmented sequence data with recurrent
  neural networks,'' in \emph{Proc. ICML}, 2006, pp. 369--376.

\bibitem{boeck2014disposition}
R.~Boeck, K.~Bergmann, and P.~Jaecks, ``Disposition recognition from
  spontaneous speech towards a combination with co-speech gestures,'' in
  \emph{Proceedings of the 2nd International Workshop on Multimodal Analyses
  enabling Artificial Agents in Human-Machine Interaction}, 2014.

\bibitem{bhattacharya2021speech2affectivegestures}
U.~Bhattacharya, E.~Childs, N.~Rewkowski, and D.~Manocha,
  ``Speech2affectivegestures: Synthesizing co-speech gestures with generative
  adversarial affective expression learning,'' in \emph{Proc. ACM MM}, 2021,
  pp. 2027--2036.

\bibitem{wen2019face}
Y.~Wen, B.~Raj, and R.~Singh, ``Face reconstruction from voice using generative
  adversarial networks,'' \emph{Advances in Neural Information Processing
  Systems (NIPS)}, vol.~32, 2019.

\bibitem{schroff2015facenet}
F.~Schroff, D.~Kalenichenko, and J.~Philbin, ``Facenet: A unified embedding for
  face recognition and clustering,'' in \emph{Proc. IEEE-CVPR}, 2015, pp.
  815--823.

\bibitem{deng2019arcface}
J.~Deng, J.~Guo, N.~Xue, and S.~Zafeiriou, ``Arcface: Additive angular margin
  loss for deep face recognition,'' in \emph{Proc. IEEE/CVF-CVPR}, 2019, pp.
  4690--4699.

\bibitem{saunders2020progressive}
B.~Saunders, N.~C. Camgoz, and R.~Bowden, ``Progressive transformers for
  end-to-end sign language production,'' in \emph{Proc. ECCV}, 2020, pp.
  687--705.

\bibitem{stoll2018sign}
S.~Stoll, N.~C. Camg{\"o}z, S.~Hadfield, and R.~Bowden, ``Sign language
  production using neural machine translation and generative adversarial
  networks,'' in \emph{Proc. BMVC}.\hskip 1em plus 0.5em minus 0.4em\relax
  British Machine Vision Association, 2018.

\bibitem{vasani2020generation}
N.~Vasani, P.~Autee, S.~Kalyani, and R.~Karani, ``Generation of indian sign
  language by sentence processing and generative adversarial networks,'' in
  \emph{Proc. IEEE-ICISS}, 2020, pp. 1250--1255.

\bibitem{ventura2020can}
L.~Ventura, A.~Duarte, and X.~Gir{\'o}-i Nieto, ``Can everybody sign now?
  exploring sign language video generation from 2d poses,'' \emph{arXiv
  preprint arXiv:2012.10941}, 2020.

\bibitem{xiao2020skeleton}
Q.~Xiao, M.~Qin, and Y.~Yin, ``Skeleton-based chinese sign language recognition
  and generation for bidirectional communication between deaf and hearing
  people,'' \emph{Neural networks}, vol. 125, pp. 41--55, 2020.

\bibitem{rothauser1969ieee}
E.~Rothauser, ``Ieee recommended practice for speech quality measurements,''
  \emph{IEEE Transactions on Audio and Electroacoustics}, vol.~17, no.~3, pp.
  225--246, 1969.

\bibitem{kim2023flame}
J.~Kim, J.~Kim, and S.~Choi, ``Flame: Free-form language-based motion synthesis
  \& editing,'' in \emph{Proc. Conf AAAI Artif. Intell.}, vol.~37, no.~7, 2023,
  pp. 8255--8263.

\bibitem{cassell2001beat}
J.~Cassell, H.~H. Vilhj{\'a}lmsson, and T.~Bickmore, ``Beat: the behavior
  expression animation toolkit,'' in \emph{Proc. ACM SIGGRAPH}, 2001, pp.
  477--486.

\bibitem{cassell1994animated}
J.~Cassell, C.~Pelachaud, N.~Badler, M.~Steedman, B.~Achorn, T.~Becket,
  B.~Douville, S.~Prevost, and M.~Stone, ``Animated conversation: rule-based
  generation of facial expression, gesture \& spoken intonation for multiple
  conversational agents,'' in \emph{Proc. ACM SIGGRAPH}, 1994, pp. 413--420.

\bibitem{wagner2014gesture}
P.~Wagner, Z.~Malisz, and S.~Kopp, ``Gesture and speech in interaction: An
  overview,'' \emph{Speech Communication}, vol.~57, pp. 209--232, 2014.

\bibitem{levine2010gesture}
S.~Levine, P.~Kr{\"a}henb{\"u}hl, S.~Thrun, and V.~Koltun, ``Gesture
  controllers,'' \emph{ACM Transactions on Graphics (TOG)}, vol.~29, no.~4, pp.
  1--11, 2010.

\bibitem{qian2021speech}
S.~Qian, Z.~Tu, Y.~Zhi, W.~Liu, and S.~Gao, ``Speech drives templates:
  Co-speech gesture synthesis with learned templates,'' in \emph{Proc.
  IEEE/CVF-ICCV}, 2021, pp. 11\,077--11\,086.

\bibitem{Dhariwal2021diffusion}
P.~Dhariwal and A.~Q. Nichol, ``Diffusion models beat {GAN}s on image
  synthesis,'' in \emph{Advances in Neural Information Processing Systems
  (NIPS)}, 2021.

\bibitem{zhu2023taming}
L.~Zhu, X.~Liu, X.~Liu, R.~Qian, Z.~Liu, and L.~Yu, ``Taming diffusion models
  for audio-driven co-speech gesture generation,'' in \emph{Proc.
  IEEE/CVF-CVPR}, 2023, pp. 10\,544--10\,553.

\bibitem{teager1990evidence}
H.~Teager and S.~Teager, ``Evidence for nonlinear sound production mechanisms
  in the vocal tract,'' \emph{Speech production and speech modelling}, pp.
  241--261, 1990.

\bibitem{oh2019speech2face}
T.-H. Oh, T.~Dekel, C.~Kim, I.~Mosseri, W.~T. Freeman, M.~Rubinstein, and
  W.~Matusik, ``Speech2face: Learning the face behind a voice,'' in \emph{Proc.
  IEEE/CVF-CVPR}, 2019, pp. 7539--7548.

\bibitem{parkhi2015deep}
O.~Parkhi, A.~Vedaldi, and A.~Zisserman, ``Deep face recognition,'' in
  \emph{Proc. BMVC}.\hskip 1em plus 0.5em minus 0.4em\relax British Machine
  Vision Association, 2015.

\bibitem{duarte2019wav2pix}
A.~C. Duarte, F.~Roldan, M.~Tubau, J.~Escur, S.~Pascual, A.~Salvador,
  E.~Mohedano, K.~McGuinness, J.~Torres, and X.~Giro-i Nieto, ``Wav2pix:
  Speech-conditioned face generation using generative adversarial networks.''
  in \emph{Proc. IEEE-ICASSP}, 2019, pp. 8633--8637.

\bibitem{pascual2017segan}
S.~Pascual, A.~Bonafonte, and J.~Serr{\`a}, ``{SEGAN}: Speech enhancement
  generative adversarial network,'' \emph{Proc. Interspeech}, pp. 3642--3646,
  2017.

\bibitem{fang2022facial}
Z.~Fang, Z.~Liu, T.~Liu, C.-C. Hung, J.~Xiao, and G.~Feng, ``Facial expression
  gan for voice-driven face generation,'' \emph{The Visual Computer}, pp.
  1--14, 2022.

\bibitem{cole2017synthesizing}
F.~Cole, D.~Belanger, D.~Krishnan, A.~Sarna, I.~Mosseri, and W.~T. Freeman,
  ``Synthesizing normalized faces from facial identity features,'' in
  \emph{Proc. IEEE-CVPR}, 2017, pp. 3703--3712.

\bibitem{goodfellow2020generative}
I.~Goodfellow, J.~Pouget-Abadie, M.~Mirza, B.~Xu, D.~Warde-Farley, S.~Ozair,
  A.~Courville, and Y.~Bengio, ``Generative adversarial networks,''
  \emph{Communications of the ACM}, vol.~63, no.~11, pp. 139--144, 2020.

\bibitem{wang2022residual}
J.~Wang, Z.~Wang, X.~Hu, X.~Li, Q.~Fang, and L.~Liu, ``Residual-guided
  personalized speech synthesis based on face image,'' in \emph{Proc.
  IEEE-ICASSP}, 2022, p. 4743–4747.

\bibitem{king2009dlib}
D.~E. King, ``Dlib-ml: A machine learning toolkit,'' \emph{The Journal of
  Machine Learning Research}, vol.~10, pp. 1755--1758, 2009.

\bibitem{chen2018lip}
L.~Chen, Z.~Li, R.~K. Maddox, Z.~Duan, and C.~Xu, ``Lip movements generation at
  a glance,'' in \emph{Proc. ECCV}, 2018, pp. 520--535.

\bibitem{song2019talking}
Y.~Song, J.~Zhu, D.~Li, A.~Wang, and H.~Qi, ``Talking face generation by
  conditional recurrent adversarial network,'' in \emph{Proc. IJCAI}, 2019, pp.
  919--925.

\bibitem{zhou2019talking}
H.~Zhou, Y.~Liu, Z.~Liu, P.~Luo, and X.~Wang, ``Talking face generation by
  adversarially disentangled audio-visual representation,'' in \emph{Proc. Conf
  AAAI Artif. Intell.}, vol.~33, no.~01, 2019, pp. 9299--9306.

\bibitem{chen2019hierarchical}
L.~Chen, R.~K. Maddox, Z.~Duan, and C.~Xu, ``Hierarchical cross-modal talking
  face generation with dynamic pixel-wise loss,'' in \emph{Proc.
  IEEE/CVF-CVPR}, 2019, pp. 7832--7841.

\bibitem{vougioukas2019end}
K.~Vougioukas, S.~Petridis, and M.~Pantic, ``End-to-end speech-driven realistic
  facial animation with temporal {GAN}s.'' in \emph{CVPR Workshops}, 2019, pp.
  37--40.

\bibitem{kefalas2020speech}
T.~Kefalas, K.~Vougioukas, Y.~Panagakis, S.~Petridis, J.~Kossaifi, and
  M.~Pantic, ``Speech-driven facial animation using polynomial fusion of
  features,'' in \emph{Proc. IEEE-ICASSP}, 2020, pp. 3487--3491.

\bibitem{sinha2020identity}
S.~Sinha, S.~Biswas, and B.~Bhowmick, ``Identity-preserving realistic talking
  face generation,'' in \emph{Proc. IEEE-IJCNN}, 2020, pp. 1--10.

\bibitem{wang2020speech}
W.~Wang, Y.~Wang, J.~Sun, Q.~Liu, J.~Liang, and T.~Li, ``Speech driven talking
  head generation via attentional landmarks based representation.'' in
  \emph{Proc. Interspeech}, 2020, pp. 1326--1330.

\bibitem{eskimez2020end}
S.~E. Eskimez, R.~K. Maddox, C.~Xu, and Z.~Duan, ``End-to-end generation of
  talking faces from noisy speech,'' in \emph{Proc. IEEE-ICASSP}, 2020, pp.
  1948--1952.

\bibitem{yi2020audio}
R.~Yi, Z.~Ye, J.~Zhang, H.~Bao, and Y.-J. Liu, ``Audio-driven talking face
  video generation with learning-based personalized head pose,'' \emph{arXiv
  preprint arXiv:2002.10137}, 2020.

\bibitem{chen2020talking}
L.~Chen, G.~Cui, C.~Liu, Z.~Li, Z.~Kou, Y.~Xu, and C.~Xu, ``Talking-head
  generation with rhythmic head motion,'' in \emph{Proc. ECCV}, 2020.

\bibitem{mittal2020animating}
G.~Mittal and B.~Wang, ``Animating face using disentangled audio
  representations,'' in \emph{Proc. IEEE/CVF-WACV}, 2020, pp. 3290--3298.

\bibitem{zhu2021arbitrary}
H.~Zhu, H.~Huang, Y.~Li, A.~Zheng, and R.~He, ``Arbitrary talking face
  generation via attentional audio-visual coherence learning,'' in \emph{Proc.
  IJCAI}, 2021, pp. 2362--2368.

\bibitem{zhang2021facial}
C.~Zhang, Y.~Zhao, Y.~Huang, M.~Zeng, S.~Ni, M.~Budagavi, and X.~Guo, ``Facial:
  Synthesizing dynamic talking face with implicit attribute learning,'' in
  \emph{Proc. IEEE/CVF-ICCV}, 2021, pp. 3867--3876.

\bibitem{si2021speech2video}
S.~Si, J.~Wang, X.~Qu, N.~Cheng, W.~Wei, X.~Zhu, and J.~Xiao, ``Speech2video:
  Cross-modal distillation for speech to video generation,'' \emph{arXiv
  preprint arXiv:2107.04806}, 2021.

\bibitem{chen2021talking}
S.~Chen, Z.~Liu, J.~Liu, Z.~Yan, and L.~Wang, ``Talking head generation with
  audio and speech related facial action units,'' \emph{arXiv preprint
  arXiv:2110.09951}, 2021.

\bibitem{zhou2021pose}
H.~Zhou, Y.~Sun, W.~Wu, C.~C. Loy, X.~Wang, and Z.~Liu, ``Pose-controllable
  talking face generation by implicitly modularized audio-visual
  representation,'' in \emph{Proc. IEEE/CVF-CVPR}, 2021, pp. 4176--4186.

\bibitem{liang2022expressive}
B.~Liang, Y.~Pan, Z.~Guo, H.~Zhou, Z.~Hong, X.~Han, J.~Han, J.~Liu, E.~Ding,
  and J.~Wang, ``Expressive talking head generation with granular audio-visual
  control,'' in \emph{Proc. IEEE/CVF-CVPR}, 2022, pp. 3387--3396.

\bibitem{wang2022one}
S.~Wang, L.~Li, Y.~Ding, and X.~Yu, ``One-shot talking face generation from
  single-speaker audio-visual correlation learning,'' in \emph{Proc. Conf AAAI
  Artif. Intell.}, vol.~36, no.~3, 2022, pp. 2531--2539.

\bibitem{ji2022eamm}
X.~Ji, H.~Zhou, K.~Wang, Q.~Wu, W.~Wu, F.~Xu, and X.~Cao, ``Eamm: One-shot
  emotional talking face via audio-based emotion-aware motion model,'' in
  \emph{Proc. ACM SIGGRAPH}, 2022, pp. 1--10.

\bibitem{gururani2022spacex}
S.~Gururani, A.~Mallya, T.-C. Wang, R.~Valle, and M.-Y. Liu, ``Spacex:
  Speech-driven portrait animation with controllable expression,'' \emph{arXiv
  preprint arXiv:2211.09809}, 2022.

\bibitem{wu2023audio}
R.~Wu, Y.~Yu, F.~Zhan, J.~Zhang, X.~Zhang, and S.~Lu, ``Audio-driven talking
  face generation with diverse yet realistic facial animations,'' \emph{arXiv
  preprint arXiv:2304.08945}, 2023.

\bibitem{liu2023opt}
J.~Liu, X.~Wang, X.~Fu, Y.~Chai, C.~Yu, J.~Dai, and J.~Han, ``Opt: One-shot
  pose-controllable talking head generation,'' in \emph{Proc. IEEE-ICASSP},
  2023, pp. 1--5.

\bibitem{wang2023progressive}
D.~Wang, Y.~Deng, Z.~Yin, H.-Y. Shum, and B.~Wang, ``Progressive disentangled
  representation learning for fine-grained controllable talking head
  synthesis,'' in \emph{Proc. IEEE/CVF-CVPR}, 2023, pp. 17\,979--17\,989.

\bibitem{zhang2023talking}
L.~Zhang, Q.~Chen, and Z.~Liu, ``Talking head generation for media interaction
  system with feature disentanglement,'' in \emph{Proc. IEEE-ICPADS}.\hskip 1em
  plus 0.5em minus 0.4em\relax IEEE, 2023, pp. 403--410.

\bibitem{wiles2018x2face}
O.~Wiles, A.~Koepke, and A.~Zisserman, ``X2face: A network for controlling face
  generation using images, audio, and pose codes,'' in \emph{Proc. ECCV}, 2018,
  pp. 670--686.

\bibitem{jamaludin2019you}
A.~Jamaludin, J.~S. Chung, and A.~Zisserman, ``You said that?: Synthesising
  talking faces from audio,'' \emph{International Journal of Computer Vision},
  vol. 127, pp. 1767--1779, 2019.

\bibitem{wen2020photorealistic}
X.~Wen, M.~Wang, C.~Richardt, Z.-Y. Chen, and S.-M. Hu, ``Photorealistic
  audio-driven video portraits,'' \emph{IEEE Transactions on Visualization and
  Computer Graphics}, vol.~26, no.~12, pp. 3457--3466, 2020.

\bibitem{lahiri2021lipsync3d}
A.~Lahiri, V.~Kwatra, C.~Frueh, J.~Lewis, and C.~Bregler, ``Lipsync3d:
  Data-efficient learning of personalized 3d talking faces from video using
  pose and lighting normalization,'' in \emph{Proc. IEEE/CVF-CVPR}, 2021, pp.
  2755--2764.

\bibitem{lu2021live}
Y.~Lu, J.~Chai, and X.~Cao, ``Live speech portraits: real-time photorealistic
  talking-head animation,'' \emph{ACM Transactions on Graphics (TOG)}, vol.~40,
  no.~6, pp. 1--17, 2021.

\bibitem{bigioi2022pose}
D.~Bigioi, H.~Jordan, R.~Jain, R.~McDonnell, and P.~Corcoran, ``Pose-aware
  speech driven facial landmark animation pipeline for automated dubbing,''
  \emph{IEEE Access}, vol.~10, pp. 133\,357--133\,369, 2022.

\bibitem{zhang2023sadtalker}
W.~Zhang, X.~Cun, X.~Wang, Y.~Zhang, X.~Shen, Y.~Guo, Y.~Shan, and F.~Wang,
  ``Sadtalker: Learning realistic 3d motion coefficients for stylized
  audio-driven single image talking face animation,'' in \emph{Proc.
  IEEE/CVF-CVPR}, 2023, pp. 8652--8661.

\bibitem{yao2022dfa}
S.~Yao, R.~Zhong, Y.~Yan, G.~Zhai, and X.~Yang, ``Dfa-nerf: personalized
  talking head generation via disentangled face attributes neural rendering,''
  \emph{arXiv preprint arXiv:2201.00791}, 2022.

\bibitem{shen2022learning}
S.~Shen, W.~Li, Z.~Zhu, Y.~Duan, J.~Zhou, and J.~Lu, ``Learning dynamic facial
  radiance fields for few-shot talking head synthesis,'' in \emph{Proc. ECCV},
  2022, pp. 666--682.

\bibitem{liu2022semantic}
X.~Liu, Y.~Xu, Q.~Wu, H.~Zhou, W.~Wu, and B.~Zhou, ``Semantic-aware implicit
  neural audio-driven video portrait generation,'' in \emph{Proc. ECCV}, 2022,
  pp. 106--125.

\bibitem{yu2022talking}
Z.~Yu, Z.~Yin, D.~Zhou, D.~Wang, F.~Wong, and B.~Wang, ``Talking head
  generation with probabilistic audio-to-visual diffusion priors,'' \emph{arXiv
  preprint arXiv:2212.04248}, 2022.

\bibitem{zhua2023audio}
Y.~Zhua, C.~Zhanga, Q.~Liub, and X.~Zhoub, ``Audio-driven talking head video
  generation with diffusion model,'' in \emph{Proc. IEEE-ICASSP}.\hskip 1em
  plus 0.5em minus 0.4em\relax IEEE, 2023, pp. 1--5.

\bibitem{xu2023multimodal}
C.~Xu, S.~Zhu, J.~Zhu, T.~Huang, J.~Zhang, Y.~Tai, and Y.~Liu,
  ``Multimodal-driven talking face generation via a unified diffusion-based
  generator.'' \emph{CoRR}, 2023.

\bibitem{chatfield2014return}
K.~Chatfield, K.~Simonyan, A.~Vedaldi, and A.~Zisserman, ``Return of the devil
  in the details: delving deep into convolutional nets,'' in \emph{Proc.
  BMVC}.\hskip 1em plus 0.5em minus 0.4em\relax British Machine Vision
  Association, 2014.

\bibitem{martin2021nerf}
R.~Martin-Brualla, N.~Radwan, M.~S. Sajjadi, J.~T. Barron, A.~Dosovitskiy, and
  D.~Duckworth, ``Nerf in the wild: Neural radiance fields for unconstrained
  photo collections,'' in \emph{Proc. IEEE/CVF-CVPR}, 2021, pp. 7210--7219.

\bibitem{huang2022prodiff}
R.~Huang, Z.~Zhao, H.~Liu, J.~Liu, C.~Cui, and Y.~Ren, ``Prodiff: Progressive
  fast diffusion model for high-quality text-to-speech,'' in \emph{Proc. ACM
  MM}, 2022, pp. 2595--2605.

\bibitem{saharia2022palette}
C.~Saharia, W.~Chan, H.~Chang, C.~Lee, J.~Ho, T.~Salimans, D.~Fleet, and
  M.~Norouzi, ``Palette: Image-to-image diffusion models,'' in \emph{Proc. ACM
  SIGGRAPH}, 2022, pp. 1--10.

\bibitem{wang2004image}
Z.~Wang, A.~C. Bovik, H.~R. Sheikh, and E.~P. Simoncelli, ``Image quality
  assessment: from error visibility to structural similarity,'' \emph{IEEE
  transactions on image processing}, vol.~13, no.~4, pp. 600--612, 2004.

\bibitem{zhang2018unreasonable}
R.~Zhang, P.~Isola, A.~A. Efros, E.~Shechtman, and O.~Wang, ``The unreasonable
  effectiveness of deep features as a perceptual metric,'' in \emph{Proc.
  IEEE-CVPR}, 2018, pp. 586--595.

\bibitem{tulyakov2018mocogan}
S.~Tulyakov, M.-Y. Liu, X.~Yang, and J.~Kautz, ``Mocogan: Decomposing motion
  and content for video generation,'' in \emph{Proc. IEEE-CVPR}, 2018, pp.
  1526--1535.

\bibitem{chung2017out}
J.~Chung and A.~Zisserman, ``Out of time: automated lip sync in the wild,'' in
  \emph{Proc. ACCV}, 2017.

\bibitem{narvekar2009no}
N.~D. Narvekar and L.~J. Karam, ``A no-reference perceptual image sharpness
  metric based on a cumulative probability of blur detection,'' in \emph{Proc.
  IEEE-QoMEX}, 2009, pp. 87--91.

\bibitem{de2013image}
K.~De and V.~Masilamani, ``Image sharpness measure for blurred images in
  frequency domain,'' \emph{Procedia Engineering}, vol.~64, pp. 149--158, 2013.

\bibitem{zeng2022expression}
D.~Zeng, S.~Zhao, J.~Zhang, H.~Liu, and K.~Li, ``Expression-tailored talking
  face generation with adaptive cross-modal weighting,'' \emph{Neurocomputing},
  vol. 511, pp. 117--130, 2022.

\bibitem{vougioukas2020realistic}
K.~Vougioukas, S.~Petridis, and M.~Pantic, ``Realistic speech-driven facial
  animation with {GAN}s,'' \emph{International Journal of Computer Vision},
  vol. 128, pp. 1398--1413, 2020.

\bibitem{wang2022acoustic}
J.~Wang, J.~Liu, L.~Zhao, S.~Wang, R.~Yu, and L.~Liu,
  ``Acoustic-to-articulatory inversion based on speech decomposition and
  auxiliary feature,'' in \emph{Proc. IEEE-ICASSP}, 2022, p. 4808–4812.

\bibitem{mori2012valley}
M.~Mori, K.~MacDorman, and N.~Kageki, ``The uncanny valley [from the field],''
  \emph{IEEE Robotics and Automation Magazine}, vol.~19, pp. 98--100, 06 2012.

\bibitem{sheng2021cross}
C.~Sheng, M.~Pietik{\"a}inen, Q.~Tian, and L.~Liu, ``Cross-modal
  self-supervised learning for lip reading: When contrastive learning meets
  adversarial training,'' in \emph{Proc. ACM MM}, 2021, pp. 2456--2464.

\bibitem{Mroueh2015deep}
Y.~Mroueh, E.~Marcheret, and V.~Goel, ``Deep multimodal learning for
  audio-visual speech recognition,'' in \emph{Proc. IEEE-ICASSP}, 2015, pp.
  2130--2134.

\bibitem{shillingford2019large}
B.~Shillingford, Y.~Assael, M.~W. Hoffman, T.~Paine, C.~Hughes, U.~Prabhu,
  H.~Liao, H.~Sak, K.~Rao, L.~Bennett \emph{et~al.}, ``Large-scale visual
  speech recognition,'' \emph{Proc. Interspeech}, pp. 4135--4139, 2019.

\bibitem{radford2021learning}
A.~Radford, J.~W. Kim, C.~Hallacy, A.~Ramesh, G.~Goh, S.~Agarwal, G.~Sastry,
  A.~Askell, P.~Mishkin, J.~Clark \emph{et~al.}, ``Learning transferable visual
  models from natural language supervision,'' in \emph{Proc. ICML}, 2021, pp.
  8748--8763.

\bibitem{Ramesh2021DALL-E}
A.~Ramesh, M.~Pavlov, G.~Goh, S.~Gray, C.~Voss, A.~Radford, M.~Chen, and
  I.~Sutskever, ``Zero-shot text-to-image generation,'' in \emph{Proc. ICML},
  2021, pp. 8821--8831.

\bibitem{Heusel2017FID}
M.~Heusel, H.~Ramsauer, T.~Unterthiner, B.~Nessler, and S.~Hochreiter, ``{GAN}s
  trained by a two time-scale update rule converge to a local nash
  equilibrium,'' \emph{Proc. Advances in neural information processing systems
  (NIPS)}, vol.~30, 2017.

\bibitem{mu2023generating}
E.~Mu, K.~M. Lewis, A.~V. Dalca, and J.~Guttag, ``Generating image-specific
  text improves fine-grained image classification,'' \emph{arXiv preprint
  arXiv:2307.11315}, 2023.

\end{thebibliography}
\end{document}